\documentclass{article}

\usepackage{arxiv}

\usepackage[utf8]{inputenc} 
\usepackage[T1]{fontenc}    
\usepackage{hyperref}       
\usepackage{url}            
\usepackage{booktabs}       
\usepackage{amsfonts}       
\usepackage{nicefrac}       
\usepackage{microtype}      
\usepackage{lipsum}
\usepackage{amsfonts,amssymb,amsbsy,textcomp,marvosym,amsmath,threeparttable,amsthm,subfigure,float,lastpage,lscape}
\usepackage{eurosym,mathrsfs,fancyhdr,CJK,multicol,graphics,indentfirst,color,bm,upgreek,booktabs,graphicx,multirow,warpcol}
\graphicspath{ {./images/} }

\title{AcceRL: Policy Acceleration Framework for Deep Reinforcement Learning}

\author{
 Hongjie Zhang \\
  School of Computer Science \\
  Sichuan Normal University\\
  Chengdu, Sichuan, China \\
  \texttt{zhanghongjie@sicnu.edu.cn} \\
}

\begin{document}
\maketitle
\begin{abstract}
Deep reinforcement learning has achieved great success in various fields with its super decision-making ability. However, the policy learning process requires a large amount of training time, causing energy consumption. Inspired by the redundancy of neural networks, we propose a lightweight parallel training framework based on neural network compression, \textit{AcceRL}, to accelerate the policy learning while ensuring policy quality. Specifically, \textit{AcceRL} speeds up the experience collection by flexibly combining various neural network compression methods. Overall, the \textit{AcceRL} consists of five components, namely \textit{Actor}, \textit{Learner}, \textit{Compressor}, \textit{Corrector}, and \textit{Monitor}. The \textit{Actor} uses the \textit{Compressor} to compress the \textit{Learner}'s policy network to interact with the environment. And the generated experiences are transformed by the \textit{Corrector} with Off-Policy methods, such as V-trace, Retrace and so on. Then the corrected experiences are feed to the \textit{Learner} for policy learning. We believe this is the first general reinforcement learning framework that incorporates multiple neural network compression techniques. Extensive experiments conducted in gym show that the \textit{AcceRL} reduces the time cost of the actor by about 2.0 $\times$ to 4.13 $\times$ compared to the traditional methods. Furthermore, the \textit{AcceRL} reduces the whole training time by about 29.8\% to 40.3\% compared to the traditional methods while keeps the same policy quality.
\end{abstract}


\section{Introduction}
Deep Reinforcement Learning (DRL) has achieved milestones in various decision-making tasks, such as chess \cite{mcgrath2022acquisition}, video games \cite{badia2020agent57}, financial investment \cite{yang2020qlib}, mathematics \cite{fawzi2022discovering}, and so on. The success of DRL is due to the increased amount of computation, giving it more opportunities to explore the environment and improve its policy quality. Different from the traditional supervised learning, its data are continuously generated by the agent interacting with the environment, which causes data collection to become the bottleneck of DRL training. In detail, the agent performs neural network inference on each state, and the environment updates the state according to the specific action. The states, actions and rewards constitute the experience data. To speed up the data collection, massively distributed parallel DRL frameworks are proposed. Such frameworks usually separate Actor and Learner, and use multiple environments to collect data simultaneously, which is to exchange space for time. Specifically, the parallel and distributed training framework includes GPU-based A3C (GA3C) \cite{babaeizadeh2016ga3c}, Sample Factory \cite{petrenko2020sample}, Importance Weighted Actor-Learner Architecture (IMPALA) \cite{espeholt2018impala}, SEED RL \cite{espeholt2019seed}, Ape-X (Distributed PER) \cite{horgan2018distributed}, RLlib \cite{liang2018rllib}, and so on. These parallel frameworks reduce the training time from a few days to a few hours.

\begin{figure}
\centering
\centerline{\includegraphics[width=6in]{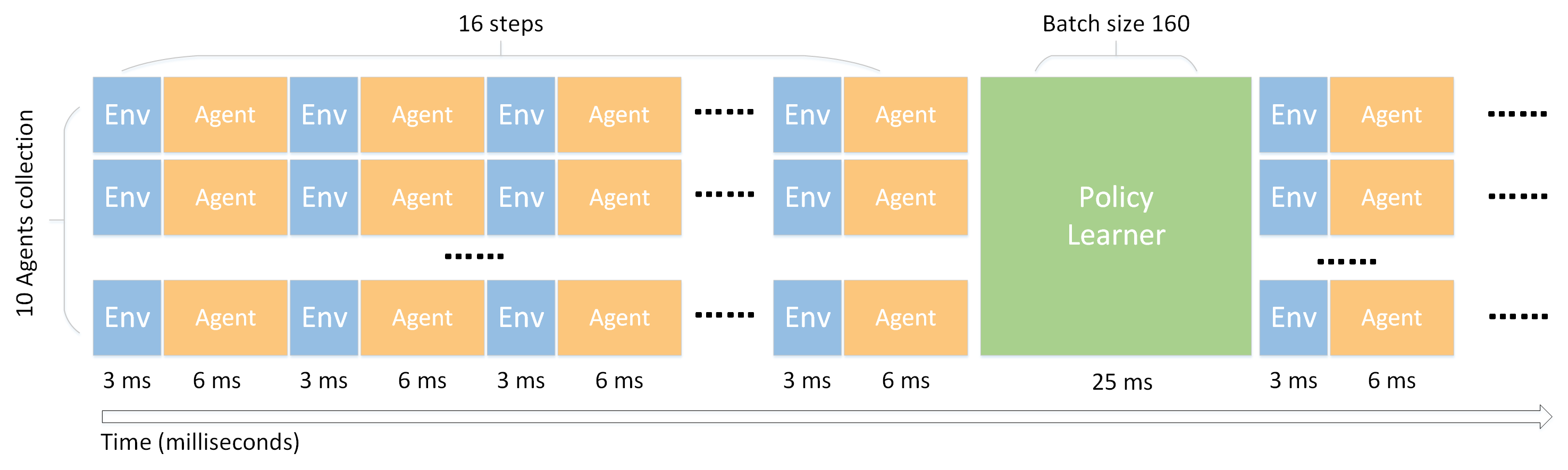}}
\caption{Quantitative analysis of data collection process and time spent}
\label{fig:data_collection}
\end{figure}

However, the above works can not directly solve the bottleneck of data collection. They just use distributed computing resources, which leads to increased energy consumption. In this work, we decomposed the data collection process and quantitatively analyzed the time spent by each module, as shown in Figure \ref{fig:data_collection}. In detail, the Agent executes policy inference to predict the action $a$ based on current state $s$, and the Env updates its state to $s'$ according to the action $a$. At each step, the Env returns a immediate reward $r$. After $n$-steps, the data $(s,a,r,s')$ are feed to Policy Learner to calculate gradient and update the parameters of the policy network. As shown in Figure \ref{fig:data_collection}, the policy inference of Agent is the bottleneck. We conducted experiments with the IMPALA framework on the Atari games, where the Env update time is 3 milliseconds, the Agent execution time is 6 milliseconds, and the policy learning time is 25 milliseconds. We set the batch size to 160, the number of agents to 10, and the steps $n$ to 16. Then the data collection time is 144 milliseconds, accounting for 85.2\% of the entire training process, and the Agent execution time accounts for 56.8\%. Based on the experimental results, we can greatly improve the training efficiency by compressing the execution time and resource consumption of the Agent.

Inspired by neural network compression, we propose a lightweight parallel reinforcement learning framework, \textit{AcceRL}, which flexibly combines multiple compression algorithms to accelerate the data collection process of Agents. Our previous work NNC-DRL \cite{zhang2019accelerating} has verified that knowledge distillation can effectively compress policy network and maintain similar convergence. Recently, CMU proposed ALD framework \cite{parisotto2021efficient}, which is similar as NNC-DRL, to compress policy network. However, the compressed policy networks of NNC-DRL and ALD are not enough to handle complex tasks, which sacrificed the final policy quality compared to original training framework, such as GA3C and IMPALA. Recently, DeepMind proposed QuaRL \cite{51674}, which leverages 8-bit quantized actors to speed up data collection. But the QuaRL only speedups the agent by about 2 $\times$ because of 8-bit quantized, which is slower than NNC-DRL. Different from the above algorithms, AcceRL combines multiple compression algorithms in a flexible way and dynamically adjusts the compression strategy to maximize the data collection of the agent while ensuring the quality of the policy. Specifically, the NNC-DRL, ALD and QuaRL are special cases of AcceRL with only one compression methods.

\textit{AcceRL} is designed to be more flexible, and we have decoupled five main modules, namely \textit{Actor}, \textit{Learner}, \textit{Compressor}, \textit{Corrector}, and \textit{Monitor}. The \textit{Actor} and \textit{Learner} are basic ideas proposed by each distributed DRL framework, which decouples policy training and data collection process. The \textit{Compressor} is responsible for the compression of the policy network, and flexibly combines various compression algorithms based on the chain of responsibility of design pattern. The \textit{Compressor} encapsulates compression algorithms, such as structured pruning \cite{cai2022structured}, knowledge distillation \cite{czarnecki2019distilling}, parameter quantization \cite{kryzhanovskiy2021qpp}, graph fusion \cite{wang2022qgtc}, matrix decomposition \cite{zhou2019tensor}, etc. If you want to combine quantization and distillation, you can easily do it by a single line of code with our \textit{AcceRL}. After compression, the Agent's policy network is inconsistent with the Learner's policy network, which leads to policy convergence problems. We designed the \textit{Corrector} module to fix the data collected by Agents. The \textit{Corrector} implements a variety of off-policy data correction algorithms, such as Importance Sampling \cite{precup2001off}, V-trace \cite{espeholt2018impala}, Retrace \cite{munos2016safe}, Tree-backup \cite{precup2001off} and so on. Even with the off-policy algorithm, the quality of the policy cannot be guaranteed, especially when the degree of compression is high. To guarantee the policy quality, we designed the \textit{Monitor} module to evaluate some key indicators in real time and dynamically adjust the compression strategy based on these indicators. We have implemented a number of commonly used indicators, such as the mean of Kullback-Leibler (KL) divergence \cite{kim2021comparing}, KL variance, KL maximum value, cumulative reward, policy entropy and so on. We implement the \textit{AcceRL} based on parallel DRL framework IMPALA, and conduct plenty of experiments on Atari games, which shows the time cost reduction of the actor by about 2.0 $\times$ to 4.13 $\times$ compared to the traditional methods while keeps the same policy quality.

To summarize, we make the following contributions:

1. We design a lightweight, extensive and flexible parallel DRL framework, \textit{AcceRL}, which reduces the cost of data collection significantly while keeping the same policy quality. As far as we know, \textit{AcceRL} is the first work that combines different compression methods in DRL training process.

2. The \textit{Compressor} provides users with an interface to flexibly combine various compression algorithms through the chain of responsibility pattern to speed up the data collection of Agent. And we design \textit{Corrector} to provide users with an interface to use its built-in data correction algorithm to ensure policy convergence.

3. We have conducted the experiments on GTX 1080Ti server to prove the efficiency of our \textit{AcceRL}, which reduces the whole training time by about 29.8\% to 40.3\% compared to the traditional methods while keeps the same policy quality.

\section{Methodology}
\subsection{Overview}
\begin{figure}[!ht]
\centering
\centerline{\includegraphics[width=5.5in]{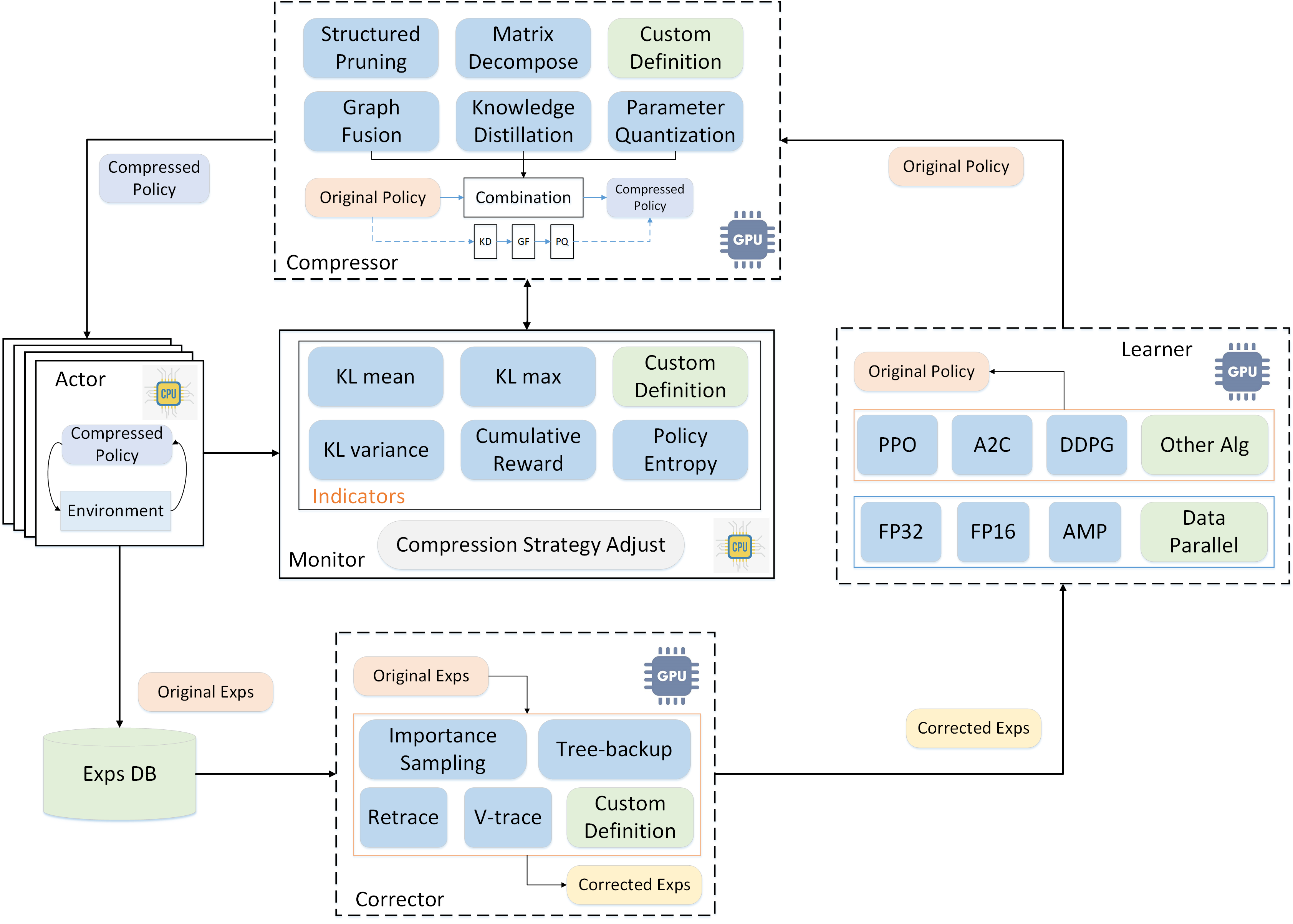}}
\caption{The high-level overview of the \textit{AcceRL} framework.}
\label{fig:accerl}
\end{figure}

Figure \ref{fig:accerl} shows a high-level overview of the \textit{AcceRL} framework. Different from the traditional Actor-Learner framework, our \textit{AcceRL} adds three additional modules (\textit{Compressor}, \textit{Corrector}, \textit{Monitor}) for policy network compression and policy quality assurance. In detail, the \textit{Compressor} fetches the original policy network from Learner, and outputs the compressed policy network by combining many built-in or custom defined compression methods. The Actors interact with each environment based on the compressed policy, which reduces the cost of data collection. To ensure the convergence of the policy, the \textit{Corrector} uses the off-policy method to post-process each experience data. Finlay, the \textit{Learner} computes the gradient of the original policy network and updates the parameters. To speedup the \textit{Learner}, we integrated the Automatic Mixed Precision (AMP) and data parallel training paradigms. To guarantee the policy quality, the \textit{Monitor} collects training information and calculates many metrics. Based on the metrics, the \textit{Monitor} dynamically adjusts the compression strategy. The whole training process is shown in Figure \ref{fig:accerl}.

In the next sections, we will introduce the design patterns and algorithms of each module in detail.

\subsection{Compressor}
As shown in Figure \ref{fig:accerl}, the \textit{Compressor} is the core in our framework. In the current version of \textit{AcceRL}, we implemented the knowledge distillation, structured pruning, parameter quantization, graph fusion and static computational graph. And the matrix low-rank factorization is ongoing. We experimented with a large number of state-of-the-art compression algorithms and found that few are compatible with commercial CPUs and GPUs. In this section, we describe in detail how each compression algorithm is implemented in the reinforcement learning and how to combine each method.

\paragraph{Knowledge Distillation}
Our previous work NNC-DRL \cite{zhang2019accelerating} already verified the effectiveness of the algorithm. Firstly, you need to define a smaller policy network before training. Secondly, the \textit{Compressor} will transfer the policy from original policy network to the smaller network. Specifically, we use the KL divergence loss to transfer the knowledge of action distribution. And we adopt the Mean Squared Error (MSE) loss to transfer the knowledge of state value function. For example, the formal definition of knowledge distillation based on Actor-Critic algorithm is shown in equation \ref{equ:KD}. In equation \ref{equ:KD}, the $\pi_{original}$ and $\pi_{smaller}$ represent the action distribution of the original policy and smaller policy network, respectively. And the $v_{original}$ and $v_{smaller}$ represent the value function of the original policy and smaller policy network, respectively. In addition, we use entropy regularization to prevent policy overfitting and the $\beta$ controls the strength of the regularization.
\begin{equation}
Loss_{KD}=KL(\pi_{original}(\cdot|s)||\pi_{smaller}(\cdot|s))+\frac{1}{2}(v_{original}(s)-v_{smaller}(s))^2-\beta H(\pi_{smaller}(\cdot|s))\label{equ:KD}
\end{equation}

\begin{figure}[!ht]
\centering
\centerline{\includegraphics[width=3.5in]{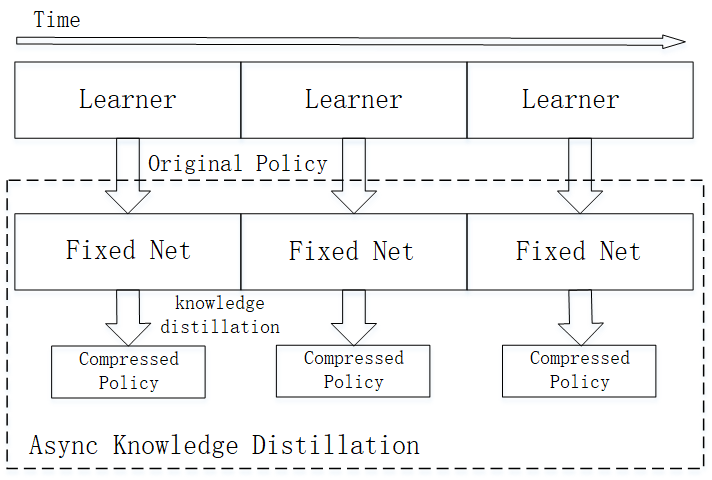}}
\caption{The asynchronous knowledge distillation framework}
\label{fig:nnc}
\end{figure}

To reduce the influence of knowledge distillation, we execute the \textit{Learner} and \textit{Compressor} asynchronously, and transfer experience data through shared memory. However, the parameters of the original policy are updated all the time, which influences the stable of knowledge distillation. To tackle this problem, we set a fixed policy network, which synchronizes parameters from the Learner at some intervals. Figure \ref{fig:nnc} shows the asynchronous knowledge distillation framework.

\paragraph{Structured Pruning}
Neural network pruning is the earliest network compression algorithm. Its core idea is to delete a batch of parameters in the neural network that have the least impact on the prediction results, thereby making the network sparse and reducing storage and execution overhead. However, the pruned neural networks use sparse matrix multiplication operations, and they usually need to be executed on specific devices, such as FPGAs \cite{vestias2021efficient}, but are not accelerated on commodity CPUs and GPUs. In our framework \textit{AcceRL}, we have adopted the structured neural network pruning technique, which has a significant acceleration effect on the server. The structured pruning try to delete feature maps in Convolutional Neural Network (CNN) and neurons in Fully Connected (FC) layers. It is equivalent to directly deleting the rows and columns of the parameter-matrix, so it can effectively speed up the calculation.

To reduce the influence of structured pruning, we still use asynchronous process to compress the policy network, which is the same as Figure \ref{fig:nnc}. Because of the prediction error of pruned network, we apply the knowledge distillation method to improve the quality of compressed policy. In our framework, the structured pruning and knowledge distillation are always used together. The challenge is how to set sparsity parameter such that it keeps policy quality while minimizing network execution time. In \textit{AcceRL}, we give the interface to the custom to set the sparsity of their own policy networks. The sparsity of pruning is sensitive to the original policy network and the environment. In order to maximize the compression effect, we provide users with an interface in \textit{Monitor} to customize the sparsity adjustment strategy.

\paragraph{Parameter Quantization}
Parameter quantization try to use low-bit value (int8 or fp16) to represent the original parameters (fp32). \textit{QuaRL} applied Post-Training Quantization (PTQ) to compress the policy in Actors. We still use PTQ in our framework. In PTQ, we design asynchronous process to quantize the policy network, which is the same as Figure \ref{fig:nnc}. Usually, PTQ and graph fusion are used together to reduce the neural network prediction overhead. Specifically, the graph fusion methods will merge two adjacent layers (eg. FC+Relu, Conv+Relu) into a single-layer, reducing the writing and reading costs of the intermediate data. In most cases, the PTQ could reduce the inference time about 2 $\times$ compared to original policy network.

\paragraph{Static Computational Graph}
The static computational graph has higher execution efficiency, especially in deep learning frameworks based on dynamic computing graphs, such as PyTorch. In our framework, the original policy network is transferred to static graph, such as Open Neural Network Exchange (ONNX) \cite{chang2021lightweight}, TorchScript \cite{devito2022torchscript}, TFLite \cite{verma2021performance} and so on. In most time, the cost of convert process cannot be ignored. We still apply the asynchronous process to convert policy network.

\begin{figure}[!ht]
\centering
\centerline{\includegraphics[width=5in]{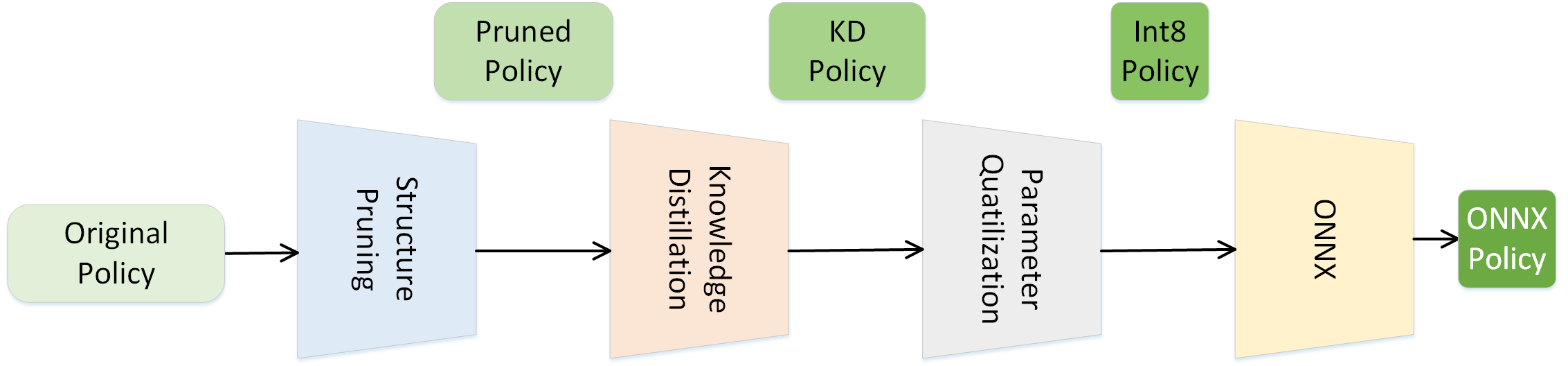}}
\caption{The chain of responsibility of different compression methods in \textit{AcceRL}}
\label{fig:compressor}
\end{figure}

More and more compression methods are on the fly, such as the Dynamic Policy Network \cite{zhang2022apinf}. There is no conflict between different compression methods. Usually, multiple compression algorithms are used together to maximize the compression ratio of the policy network. Han et at. proposed Deep Compression method to reduce the neural network in supervised learning \cite{han2015deep}. In our framework, we also apply different methods together, and design interface to customs to organize their compression strategy. Different methods are organized by the pattern of chain of responsibility, which is shown in Figure \ref{fig:compressor}. The corresponding python source code is shown in Code \ref{tab:compressor}.

\begin{table}[!ht]
\caption{The source code of the definition of compression methods.}
\label{tab:compressor}
\centering
\begin{tabular}{ll}
\toprule
1. & \textcolor{green}{\# other codes.} \\
2. & \textcolor{green}{\# The model of original policy network.} \\
3. & policyNet=AtariCNN(action\_space) \\
4. & \textcolor{green}{\# The chain of responsibility of different compression methods.} \\
5. & compreNet=CompressorChain() \\
6. & compreNet.add(Pruning(policyNet, 0.5)).add(Distiller(policyNet)).add(Quantilizer('INT8')).add(ONNX()) \\
7. & \textcolor{green}{\# Create N Actors to collect data by using compressed policy network.} \\
8. & Agents=[Actor(env, compreNet.policy, rank) \textcolor{blue}{for} rank \textcolor{blue}{in} range(num\_agents)] \\
9. & \textcolor{green}{\# learner processes.} \\
10. & \textcolor{green}{\# monitor processes.} \\
\bottomrule
\end{tabular}
\end{table}

\subsection{Corrector}
As we mentioned before, the compressed policy will sacrifice the quality of the final policy. We have carefully studied the effect of policy quality in compressed network. In our framework, the \textit{Corrector} integrates a variety of different off-policy algorithms to correct data bias. In the current version, we implemented V-trace and Importance Sampling methods. The V-trace corrects the state value function of each data collected by the \textit{Actor}. And the Importance Sampling corrects the policy gradient of each data. In most cases, the two methods are applied together.

\paragraph{V-trace Corrector}
The V-trace is proposed by DeepMind in IMPALA framework. The formal definition of V-trace \cite{espeholt2018impala} is shown in equation \ref{equ:V-trace}. Specifically, the constant $\hat{\rho}$ and $\hat{c}$ usually equals to 1. The value function $V(s_t)$ is estimated by the compressed policy network, which usually has bias. In addition, the $x_t$ and $r_t$ is the state and reward of the environment at time step $t$. The \textit{Learner} apply the corrected state value function to calculate the policy gradient, which keeps a good convergence.
\begin{equation}
\begin{split}
& v_s=V(x_s)+\sum_{t=s}^{s+n-1}{\gamma^{t-s}\left(\prod_{i=s}^{t-1}{c_i}\right)\delta_tV} \\
& \delta_t V=\rho_t(r_t+\gamma V(x_{t+1})-V(x_t)) \\
& \rho_t=min(\hat{\rho},\frac{\pi_{original}}{\pi_{smaller}}) \\
& c_i=min(\hat{c},\frac{\pi_{original}}{\pi_{smaller}}) \label{equ:V-trace}
\end{split}
\end{equation}

\paragraph{Importance Sampling Corrector}
Importance sampling uses importance ratio $\pi_{original}/\pi_{smaller}$ to correct the policy gradient of the \textit{Learner}. The formal definition of importance sampling policy gradient \cite{espeholt2018impala} is shown in equation \ref{equ:IS}, where $ADV$ is the advantage function of state-action pair $(x_t,a)$. To reduce the variance of importance sampling, we still limit the maximum value of importance ratio to 1.
\begin{equation}
E\left[\frac{\pi_{original}}{\pi_{smaller}}\triangledown log\pi_{original}(a|x_t)ADV(s_t,a)\right] \label{equ:IS}
\end{equation}

For Instance, we apply the V-trace and Importance Sampling Corrector on the Asynchronous Advantage Actor-Critic (A3C) \cite{babaeizadeh2017reinforcement} algorithm. The formal definitions are shown in equation \ref{equ:A3C-V}, which is same as our previous work NNC-DRL \cite{zhang2019accelerating}.
\begin{equation}
\begin{split}
& \rho_t \triangledown_\theta log \pi_\theta(a_t|x_t)(r_t+\gamma v_{t+1}-V_\theta(x_t)) \\
& (v_t-V_\theta(x_t))\triangledown_\theta(x_t) \\
& -\triangledown_\theta\sum_a{\pi_\theta(a_t|x_t)log \pi_\theta(a_t|x_t)} \label{equ:A3C-V}
\end{split}
\end{equation}

We also leave the interface to custom defined corrector in our framework. The usage of \textit{AcceRL} with corrector is shown in Code \ref{tab:corrector}, which is the successor of Code \ref{tab:compressor}. Specifically, the \textit{AcceRL} also uses the design pattern with chain of responsibility. The experience replay memory collect data from each Actor and correct data before feed to Learner. As shown in line 7, the learner apply Actor-Critic algorithm to train the policy network. Furthermore, we use data parallel paradigm (dpp) and Automatic Mixed Precision to accelerate policy learning.
\begin{table}[!ht]
\caption{The source code of the corrector usage.}
\label{tab:corrector}
\centering
\begin{tabular}{ll}
\toprule
1. & \textcolor{green}{\# After the Code \ref{tab:compressor}.} \\
2. & \textcolor{green}{\# The experience replay memory, which save the data from each Actor.} \\
3. & memory=QManager(Agents) \\
4. & \textcolor{green}{\# The chain of responsibility of different correction methods.} \\
5. & memory.add\_corrector(VTraceCorrector(tho,c)).add\_corrector(ISCorrector(tho)) \\
6. & \textcolor{green}{\# Create the learner with policy network and memory.} \\
7. & learner=Learner(policyNet,memory,alg='AC',dpp=True,AMP=True) \\
8. & \textcolor{green}{\# monitor processes.} \\
\bottomrule
\end{tabular}
\end{table}

\subsection{Monitor}
\begin{figure}
	\centering
	\subfigure[The training stages of Atari pong.]{\label{fig:stage:a}
		\includegraphics[width=3in]{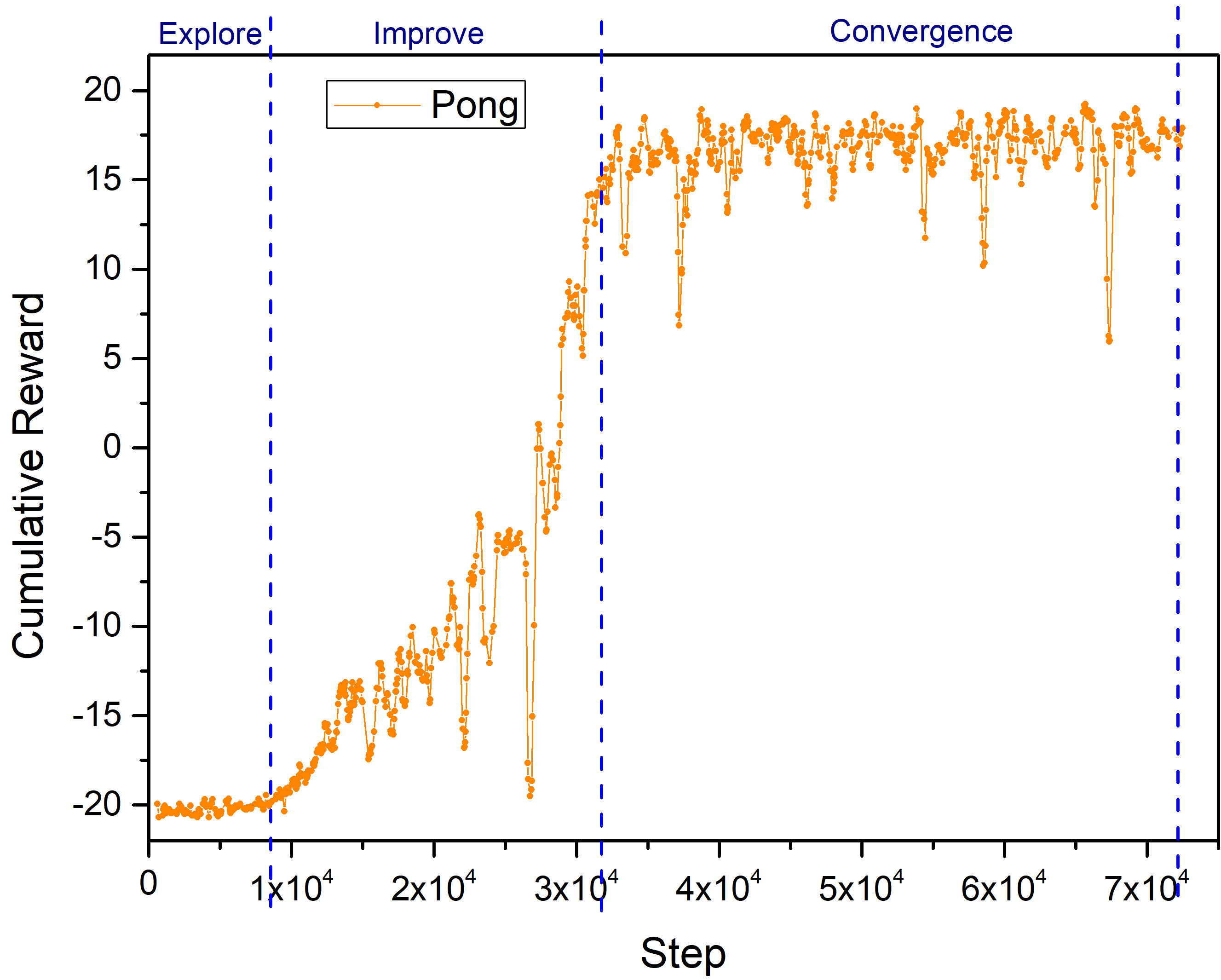}}
	\subfigure[The training stages of Atari road\_runner.]{\label{fig:stage:b}
		\includegraphics[width=3in]{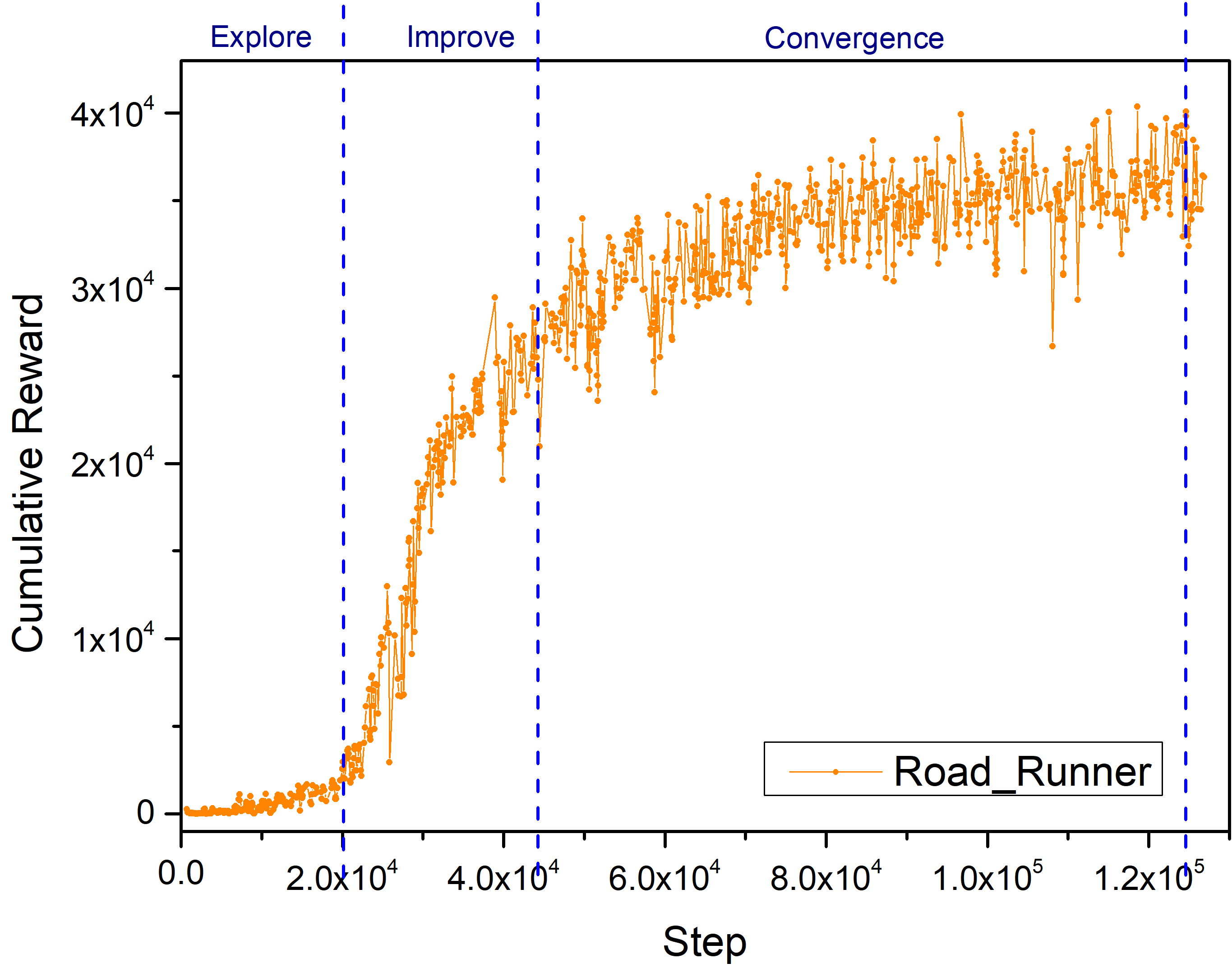}}
	\caption{We split the policy learning into multi-stages based on the cumulative rewards.}
	\label{fig:stages}
\end{figure}

Even though we improve the policy convergence with the off-policy corrector, the policy quality is still difficult to guarantee when the policy network is compressed too small. We notice that the training of DRL could split into multi-stages, which is shown in Figure 1. For example, in Figure \ref{fig:stages}, we trained the policy to complete the Atari games pong and road\_runner, and split the training steps into three stages, named exploration stage, improving stage and convergence stage. In the exploration stage, the policy randomly explores the environment, and there is no significant improvement in the policy quality at this stage.In this stage, we could use smallest policy network to collect data in each Actor. When the policy improves fast, we increase the policy network by setting bigger sparsity value in structured pruning or more bits in parameters quantization. In the convergence stage, we apply larger sparsity or lower bits to compress the policy network. The key problem is how to judge which stage the training is in. The \textit{AcceRL} provides the custom to define their own monitor indicators and compression strategies. In the current version of our framework, we implemented many indicators, such as KL mean, KL max, cumulative rewards, entropy, and so on. Based on the indicators, we designed some strategies to adjust the compression pipeline.

\paragraph{KL Max Indicator}
When we use knowledge distillation, we will monitor the KL divergence between the original policy and compressed policy with a certain interval. The definition of the KL divergence is shown in equation \ref{equ:KL}. The KL max is the maximum of the KL divergence, which means the compressed policy can not learn the action distribution of the original one. We set a predefined threshold $\epsilon$, and the $Compressor$ will adopt a conservative strategy for network compression if the KL max greater than $\epsilon$.
\begin{equation}
KL(\pi_{original}||\pi_{smaller})=\sum_a{\pi_{original}(a|x)log\frac{\pi_{original}(a|x)}{\pi_{smaller}(a|x)}} \label{equ:KL}
\end{equation}

\begin{figure}[!ht]
\centering
\centerline{\includegraphics[width=3.5in]{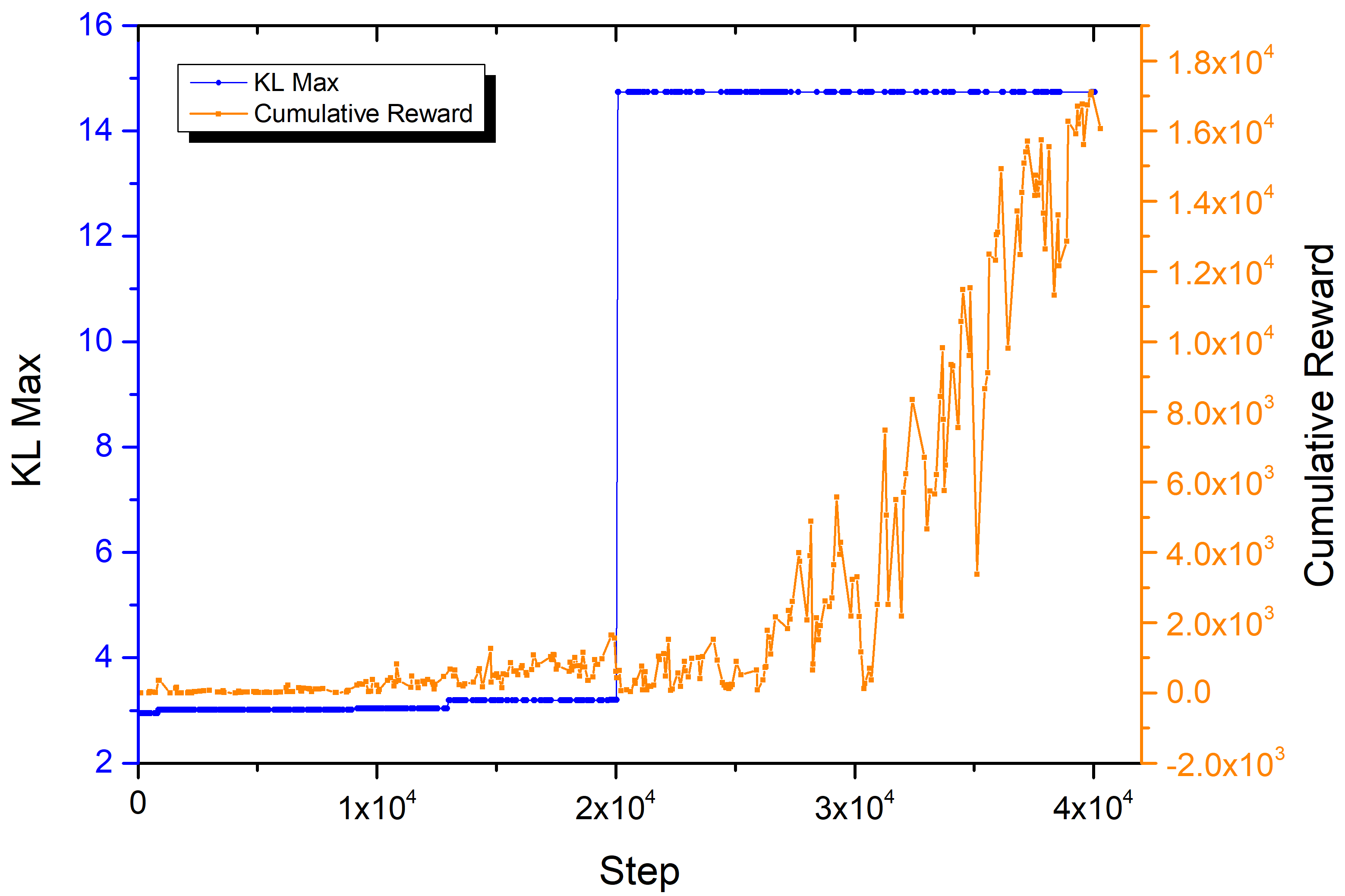}}
\caption{The KL max indicator and cumulative reward of Atari game Road Runner.}
\label{fig:kl_max}
\end{figure}

We experimented on Atari game road\_runner with KL max, which is shown in Figure \ref{fig:kl_max}. Specifically, we set the threshold $\epsilon=14.0$. We notice that the trends of KL max indicator and cumulative reward correspond to each other. We define a compression strategy with quantization, which switches the bit from 8 to 16 when $KL_{max}>14.0$. Based on the strategy, the final quality of our policy is guaranteed.

\begin{figure}[!ht]
\centering
\centerline{\includegraphics[width=3.8in]{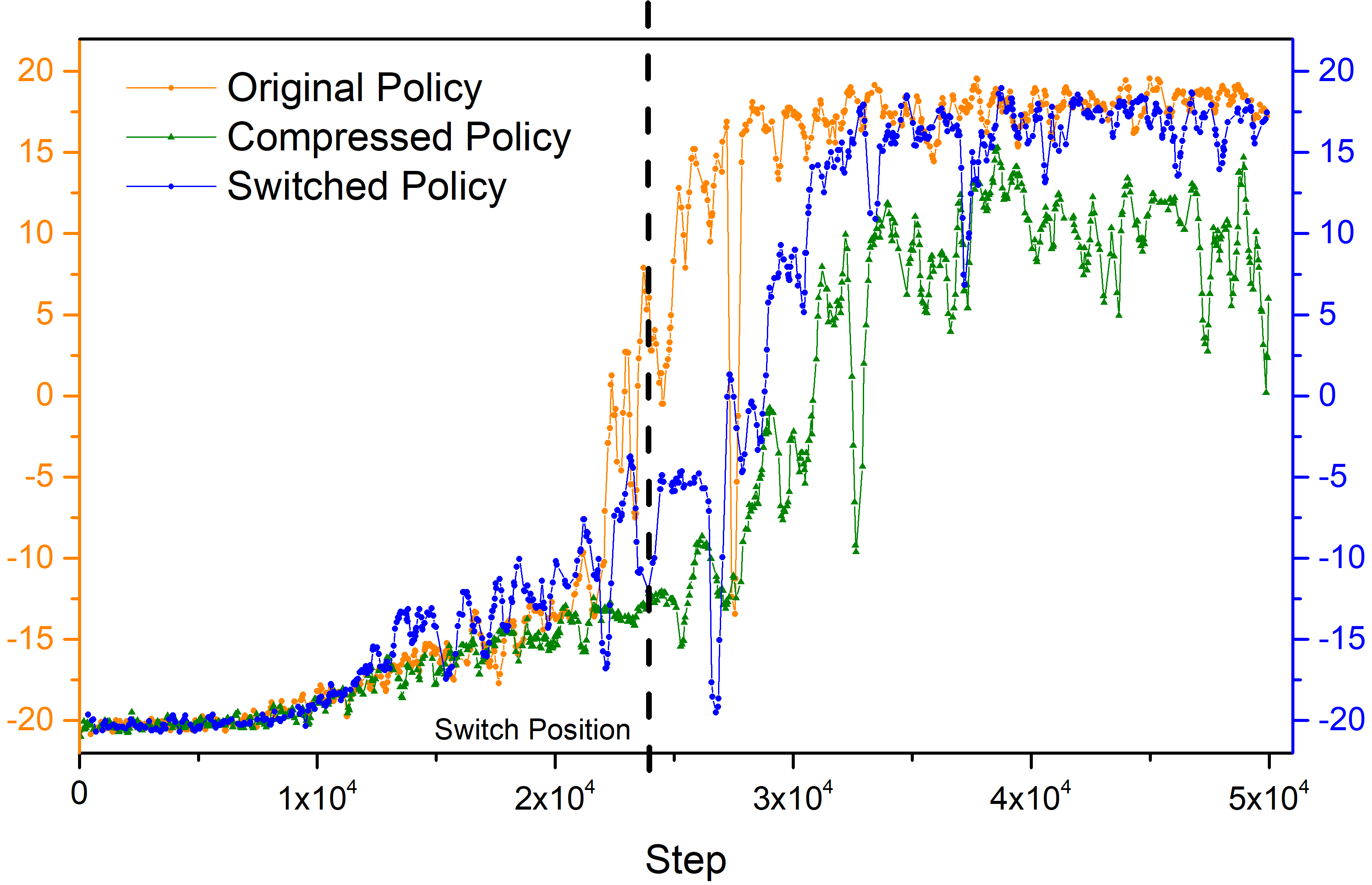}}
\caption{The cumulative reward indicator of Atari game Pong.}
\label{fig:cumu}
\end{figure}

\paragraph{Cumulative Reward Indicator}
To directly compare the quality of the original policy and compressed policy, we design the cumulative rewards indicator. When the reward gap greater than a predefined threshold $\delta$, the $Compressor$ will adopt a conservative strategy for network compression. We experimented the Cumulative Reward Indicator on Atari game Pong, which uses the Structured Pruning to compress the policy network. The result is shown in Figure \ref{fig:cumu}. In detail, the threshold $\delta=10.0$ and the switch point is $2.4\times 10^4$. At the switch point, we increase the sparsity from 0.5 to 0.2, which guarantees the final quality of the policy.

\paragraph{Advanced Indicator}
The purpose of our framework \textit{AcceRL} is minimize the training time while keeps the same policy quality. In other words, we want to maximize the growth of the cumulative reward per unit of time. To do that, we could design advanced compression strategies, which formalize it as optimization problem (e.g. equation \ref{equ:advance}). Specifically, $CR$ represents the Cumulative Reward, and $\Delta$ represents the time duration. The $\pi$ is the policy that controls the compression strategy, such as sparsity and bit-width. The compress policy $\pi$ outputs strategy based on monitor indicators. To solve the optimization problem \ref{equ:advance}, we could use policy gradient or other DRL algorithms.
\begin{equation}
\begin{split}
& \mathop{max}_p \frac{CR_{t+\Delta}-CR_{t}}{\Delta} \\
& where.\quad p\sim \pi(\cdot|KL,CR,entropy,t,others) \label{equ:advance}
\end{split}
\end{equation}

It is easy for users to apply those indicators in our \textit{AcceRL}, and the example source code is shown in Code \ref{tab:monitor}.
\begin{table}[!ht]
\caption{The source code of the monitor usage.}
\label{tab:monitor}
\centering
\begin{tabular}{ll}
\toprule
1. & \textcolor{green}{\# After the Code \ref{tab:corrector}} \\
2. & \textcolor{green}{\# Define a KLMonitor with threshold 14.0.} \\
3. & klMonitor=KLMonitor(policyNet, compreNet.policy, thre=14.0) \\
4. & \textcolor{green}{\# Create a compression strategy object.} \\
5. & compressStratege=Strategy(compreNet, klMonitor) \\
6. & \textcolor{green}{\# Processes start loop.} \\
7. & \textcolor{blue}{for} actor \textcolor{blue}{in} Agents: \\
8. & $\quad\quad$\textcolor{green}{\# Start actor process.} \\
9. & $\quad\quad$actor.start() \\
10. & \textcolor{green}{\# Start learner process.} \\
11. & learner.start() \\
12. & \textcolor{green}{\# Start monitor process.} \\
13. & compressStratege.start() \\
14. & \textcolor{green}{\# Rest Code.} \\
\bottomrule
\end{tabular}
\end{table}

\subsection{Extensible and Flexible}
Our framework \textit{AcceRL} is extensible and flexible. We provide users with custom interfaces that support extra \textit{Compressor}, \textit{Corrector}, and \textit{Monitor}. Users only need to inherit the corresponding abstract class and implement the performing interface. In terms of flexibility, users can combine different compression algorithms into a chain pattern through a single line of code, and dynamically adjust based on the strategy of \textit{Monitor}.

\section{Experiments}
We have conducted extensive experiments on multiple game tasks based on framework \textit{AcceRL} in this section. We implement our framework in PyTorch 1.10.1 version. To prove the efficiency of our algorithm, we used a GPU server with GTX 1080ti and 48-core Intel Xeon Gold 5118 processor. Each result is conducted three times with different random seed.

\subsection{Experiments Setup}
\paragraph{Training setups}
In this work, we implement \textit{AcceRL} based on parallel framework IMPALA which is the baseline algorithm. The 12 Atari games used are shown in Table \ref{tab:atari}. The DRL algorithm is Actor-Critic and the optimizer is Adam. We use FP32 to train the policy network because that the GTX 1080ti does not support FP16 and AMP training.
\begin{table}[!ht]
\caption{The 12 Atari games in the experiment.}
\label{tab:atari}
\centering
\begin{tabular}{c|c|c|c|c|c}
\toprule
alien & amidar & beam\_rider & name\_this\_game & riverraid & hero \\
\midrule
space\_invaders & breakout & road\_runner & bank\_heist & pong & seaquest \\
\bottomrule
\end{tabular}
\end{table}

For a fair comparison of each algorithm, we set the framework parameters as shown in Table \ref{tab:param_set}, where NA, ND and NT represent the number of Agent, Distiller and Learner respectively. the n-steps means the number of steps to calculate the state value function. $lr$ is the learning rate of Adam. The entropy and value represent the strength of entropy regularization and value respectively.
\begin{table}[!ht]
\caption{The parameters setting in each training process.}
\label{tab:param_set}
\centering
\begin{tabular}{l|ccccccccc}
\toprule
\textbf{Parameter} & NA & ND & NT & n-steps & $lr$ & $\gamma$ & entropy & value & batch size \\
\midrule
\textbf{Value} & 12/8 & 4/1 & 1 & 5 & $3\times10^{-4}$ & 0.99 & 0.02 & 0.25 & 160 \\
\bottomrule
\end{tabular}
\end{table}

\paragraph{Networks setups}
The policy networks in our experiment are CNN based network. The network architecture and inference time on server are shown in Table \ref{tab:param_net}. Specifically, the Net1, Net2 and Net3 are compressed networks with different size. The Input format is $C\times H\times W$. The Conv means the filer shape of the convolution layer, and F.C means the fully connected layer. The Output varies from task to task.
\begin{table}[!ht]
\caption{The network architecture and inference time.}
\label{tab:param_net}
\centering
\begin{tabular}{llllllll}
\toprule
Network & Input & Conv.1 & Conv.2 & Conv.3 & F.C & Output & Time(ms) \\
\midrule
Original & $4\times84\times84$ & 32(8) & 64(4) & 64(3) & 2048 & [1-18] & 6.0 \\
Net1 & $4\times84\times84$ & 16(8) & 32(4) & - & 256 & [1-18] & 2.5 \\
Net2 & $4\times84\times84$ & 8(8) & 16(4) & - & 128 & [1-18] & 1.9 \\
Net3 & $4\times84\times84$ & 4(8) & 4(4) & - & 32 & [1-18] & 1.2 \\
\bottomrule
\end{tabular}
\end{table}

\subsection{Inference Speedups and Convergence}
\begin{figure}[!ht]
	\centering
	\subfigure[The knowledge distillation results on training steps.]{\label{fig:kd:a}
		\includegraphics[width=3.08in]{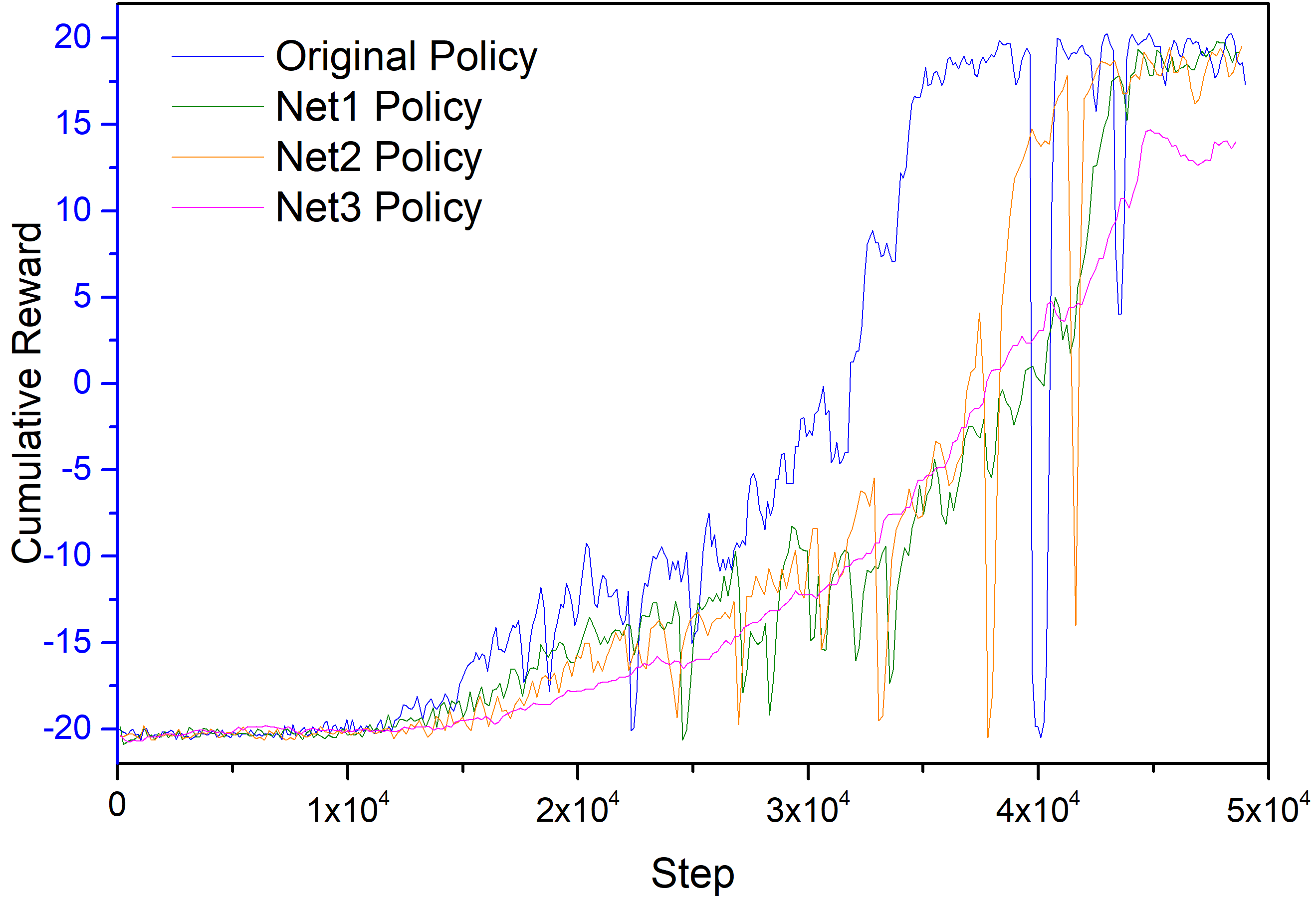}}
	\subfigure[The knowledge distillation results on training time.]{\label{fig:kd:b}
		\includegraphics[width=3in]{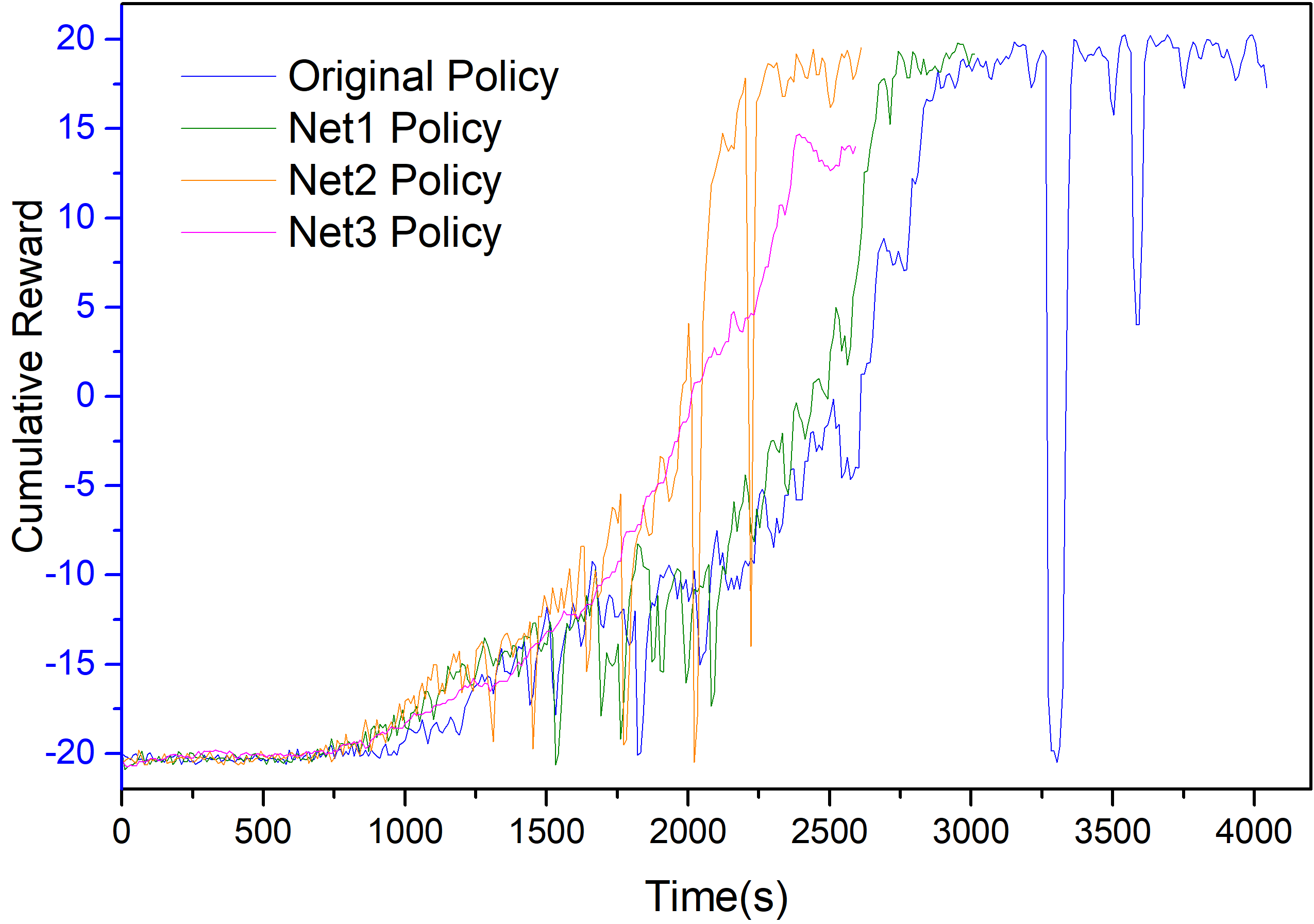}}
	\caption{The knowledge distillation results of Atari pong with $NA=12$ and $ND=4$.}
	\label{fig:kd}
\end{figure}

\paragraph{Knowledge Distillation Results}
Our previous work NNC-DRL \cite{zhang2019accelerating} has done relevant experimental comparisons under the framework of GA3C. In this work, we show the results on IMPALA framework. The comparison results on the Pong game are shown in Figure \ref{fig:kd}. The result shows that knowledge distillation could speedup the training process but decrease the final quality when the policy network is too small. Intuitively, the smaller the policy network is, the harder it is to learn the knowledge of the original policy network.

\begin{figure}[!ht]
	\centering
	\subfigure[The linear strategy results on training steps.]{\label{fig:alpha:a}
		\includegraphics[width=3.4in]{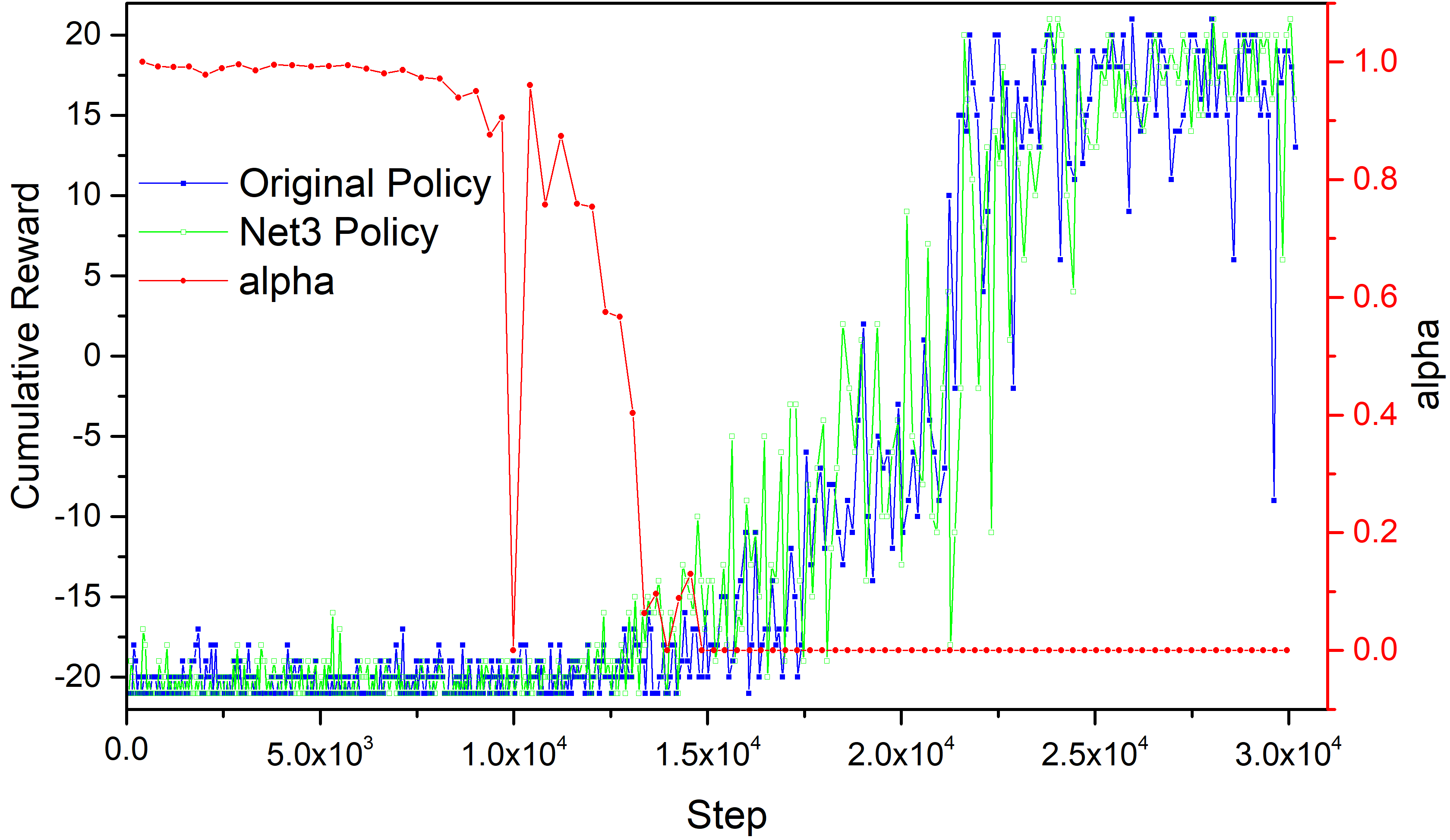}}
	\subfigure[The linear strategy results on training time.]{\label{fig:alpha:b}
		\includegraphics[width=2.6in]{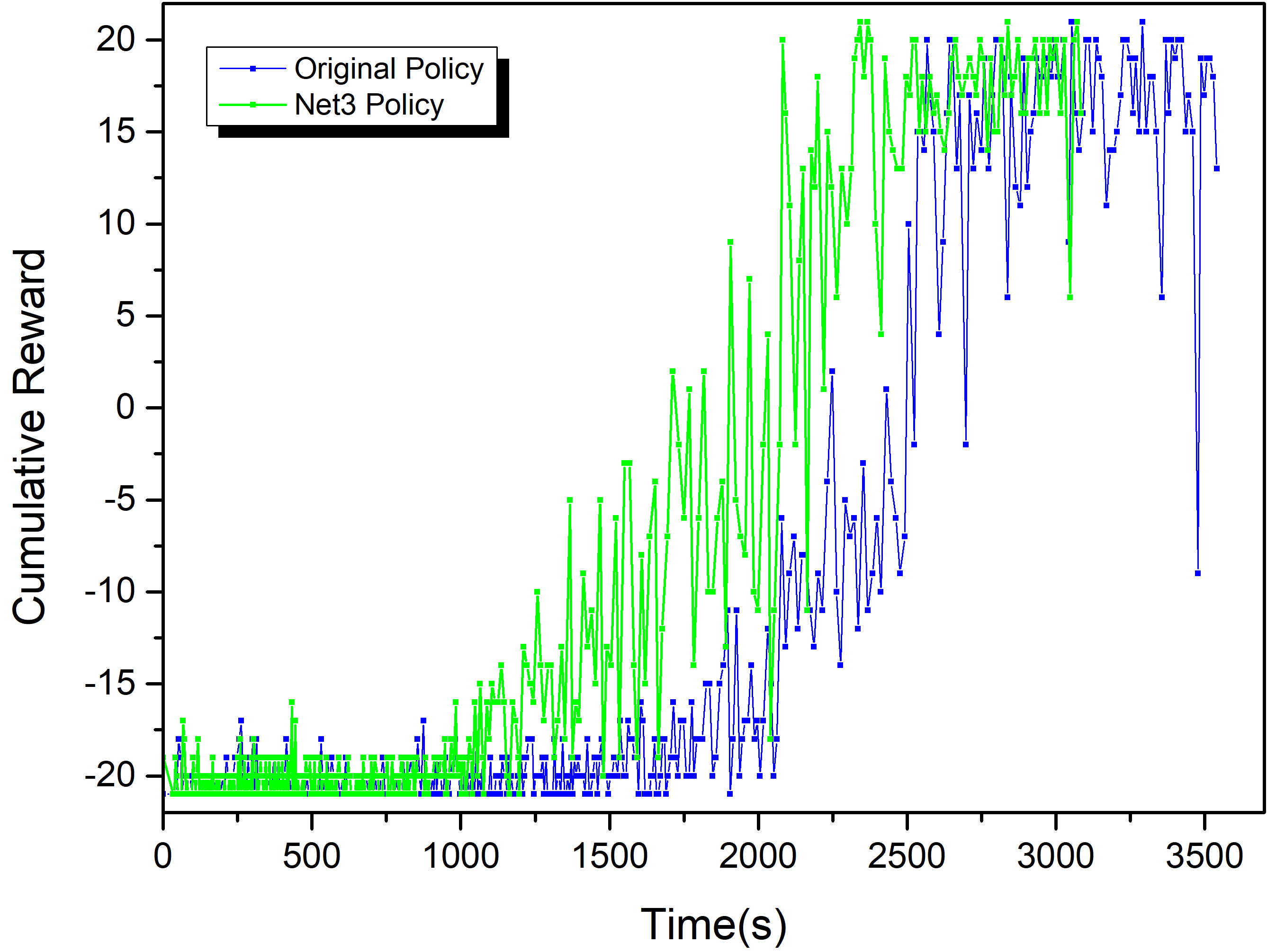}}
	\caption{The knowledge distillation results of Atari pong based on linear strategy with $NA=8$ and $ND=4$.}
	\label{fig:alpha}
\end{figure}

Specifically, the final score of Net3 is 17.3 while the final score of Original is 20.5. To guarantee the final score, we use the KL mean as indicator to dynamic adjust the compress strategy, which is shown in Figure \ref{fig:alpha}. In detail, the Actor has $\alpha$ probability of using Net3 and $(1-\alpha)$ the probability of choosing Original network, and the $\alpha$ decreases linearly with the increase of KL mean indicator. It means the Actor chooses Net3 when $\alpha=1$, which chooses Original network when $\alpha=0$. The result shows that both Net3 and Original network reach 20.4 on cumulative reward.

\paragraph{Parameter Quantization Results}
We tested INT8 quantization on 12 Atari games. The results are shown in Figure \ref{fig:quant}. The \textit{Corrector} used V-trace to fix the off-policy problem. The inference time cost of Original policy is 6 millisecond, and the 8-bits policy spends 3 millisecond per state. The results show that the \textit{AcceRL} reduces the whole training time by about 29.8\% to 40.3\% compared to the traditional methods while keeps the same policy quality. In Parameter Quantization, we use the 8-bits policy from start to end, which means the compression strategy of \textit{Monitor} is always use 8-bits network. In Figure \ref{fig:quant}, the 1-st and 3-rd columns show the policy convergence along with training step, and the 2-ed and 4-th columns show the policy convergence along with training time. We notice that the 8-bits compressed policy and Original (FP32) policy have the same convergence rate. But our 8-bits compressed policy is faster than Original policy. 

\begin{figure}[h]
	\centering
	\subfigure[The structured pruning results on training steps.]{\label{fig:prune:a}
		\includegraphics[width=3in]{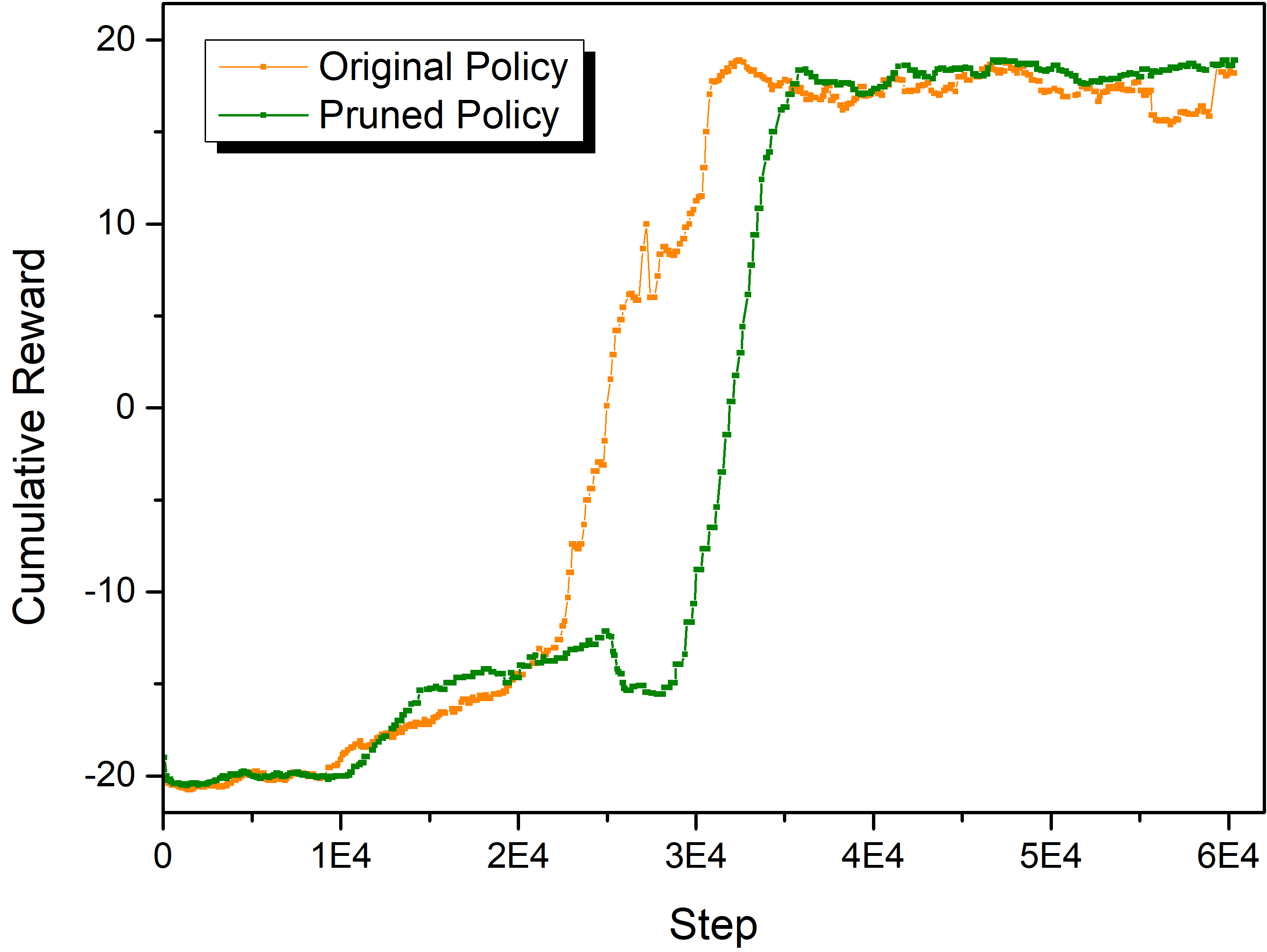}}
	\subfigure[The structured pruning results on training time.]{\label{fig:prune:b}
		\includegraphics[width=3in]{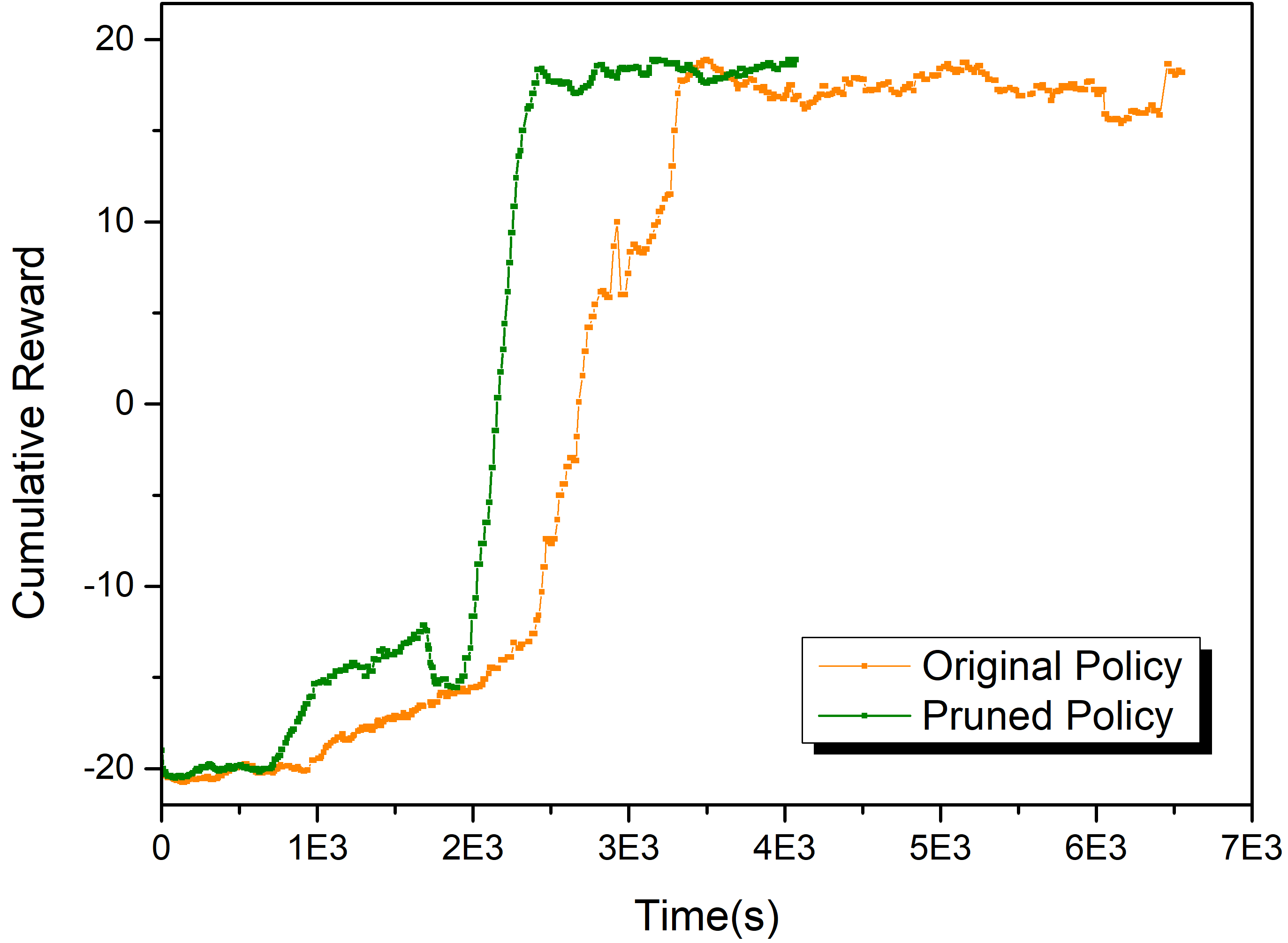}}
	\caption{The structured pruning results of Atari pong with $NA=12$ and $ND=1$.}
	\label{fig:prune}
\end{figure}

\paragraph{Structured Pruning Results}
In Structured Pruning, the users must provide a sparsity value. The larger the sparsity, the smaller the policy network and the faster the calculation speed. We experimented the structured pruning on game Pong, which is shown in Figure \ref{fig:prune}. Specifically, we set the sparsity as 0.5 and utilize V-trace to correct off-policy problem. The results show that structured pruning significantly speeds up DRL training and maintains similar convergence. The problem is how to choose the best sparsity value to minimize the training time while keep the same convergence rate. In our framework, we provide the interface to dynamic adjust the sparsity value in module \textit{Monitor}.

\begin{figure}[H]
	\centering
	\subfigbottomskip=2pt
	\subfigcapskip=-5pt
	\subfigure[alien-step]{\label{fig:comb:1}
		\includegraphics[width=0.25\linewidth]{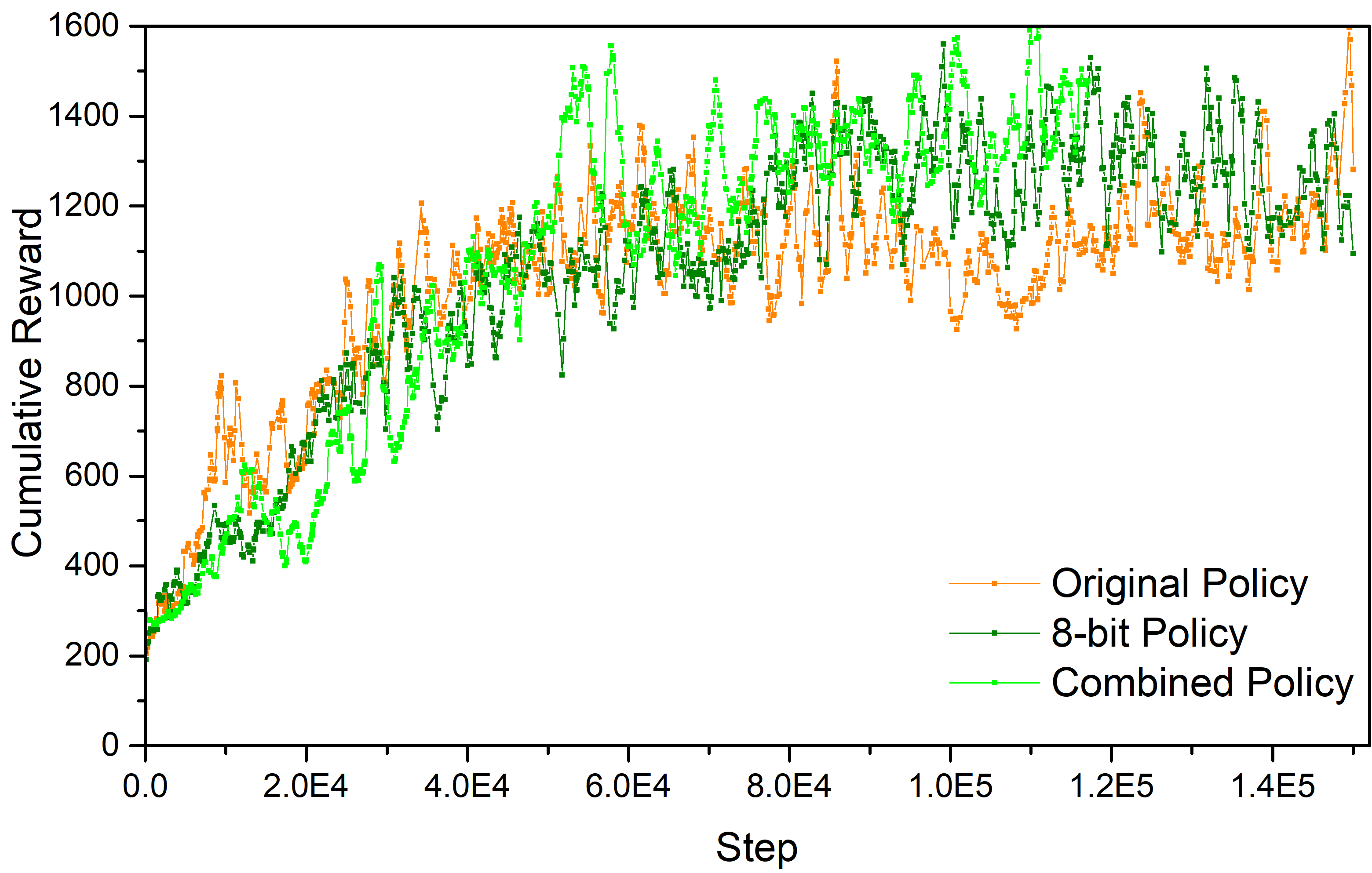}}
	\subfigure[alien-time]{\label{fig:comb:2}
		\includegraphics[width=0.25\linewidth]{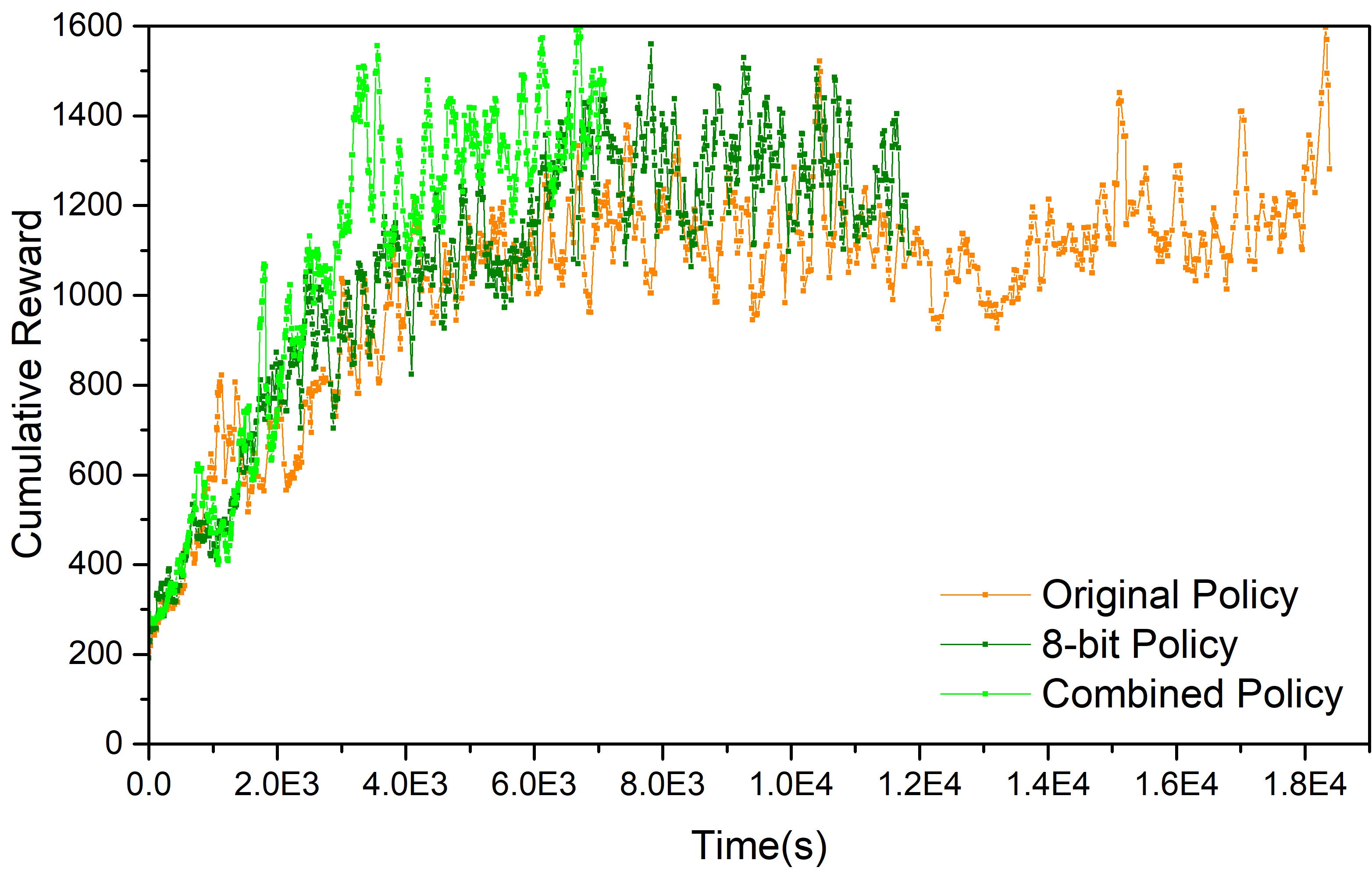}}
	\subfigure[amidar-step]{\label{fig:comb:3}
		\includegraphics[width=0.24\linewidth]{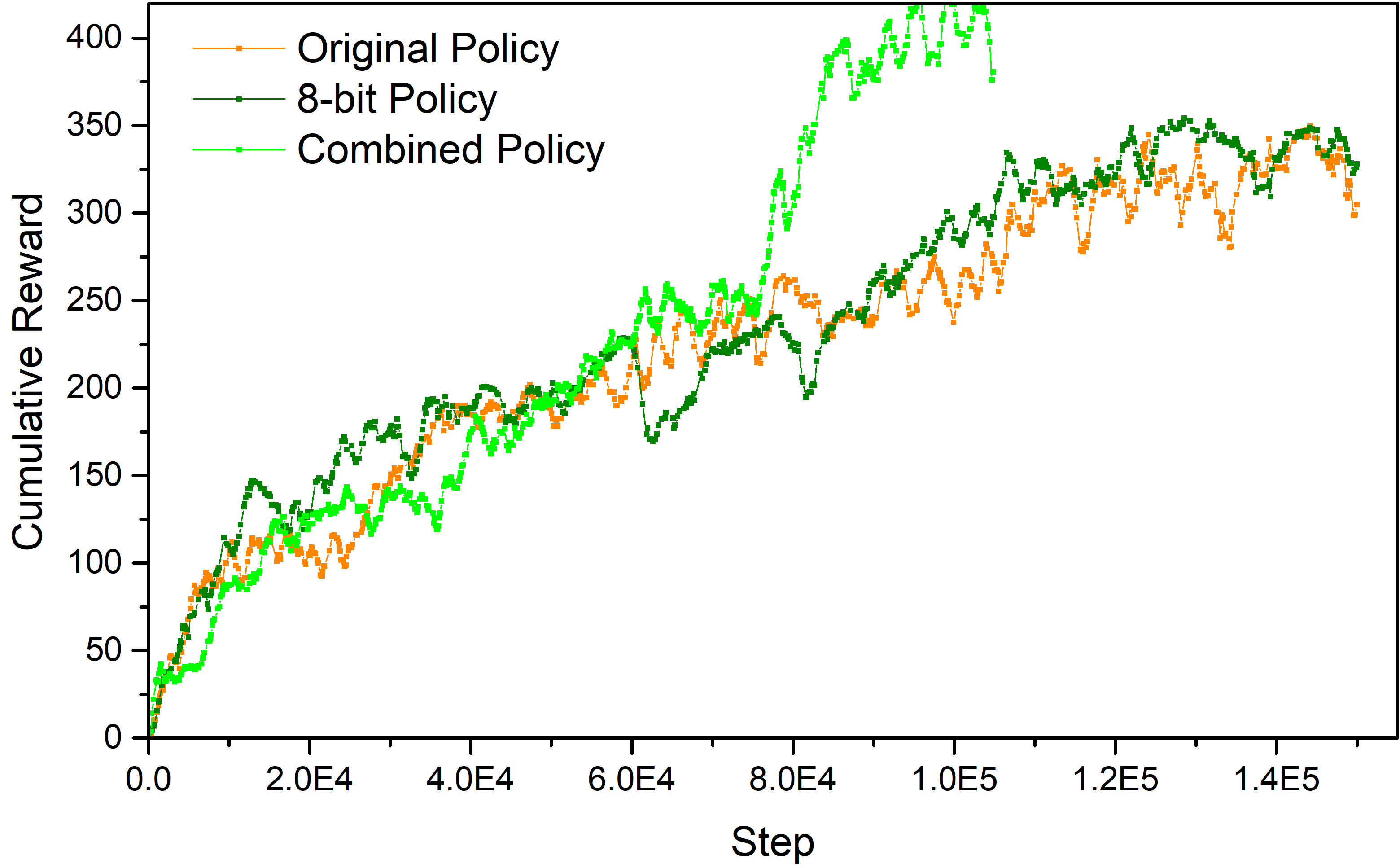}}
	\subfigure[amidar-time]{\label{fig:comb:4}
		\includegraphics[width=0.24\linewidth]{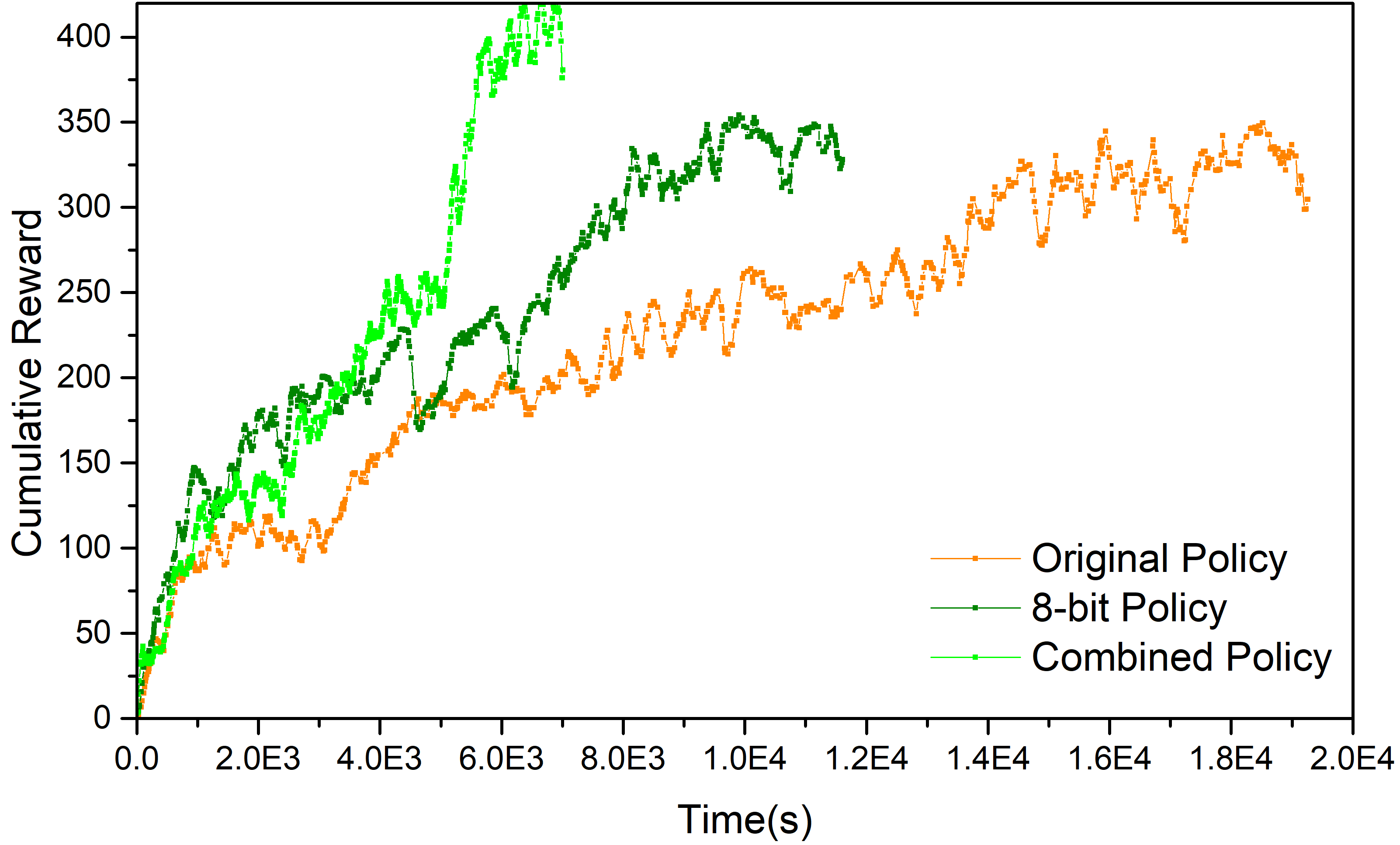}}
		
	\subfigure[beam-step]{\label{fig:comb:5}
		\includegraphics[width=0.24\linewidth]{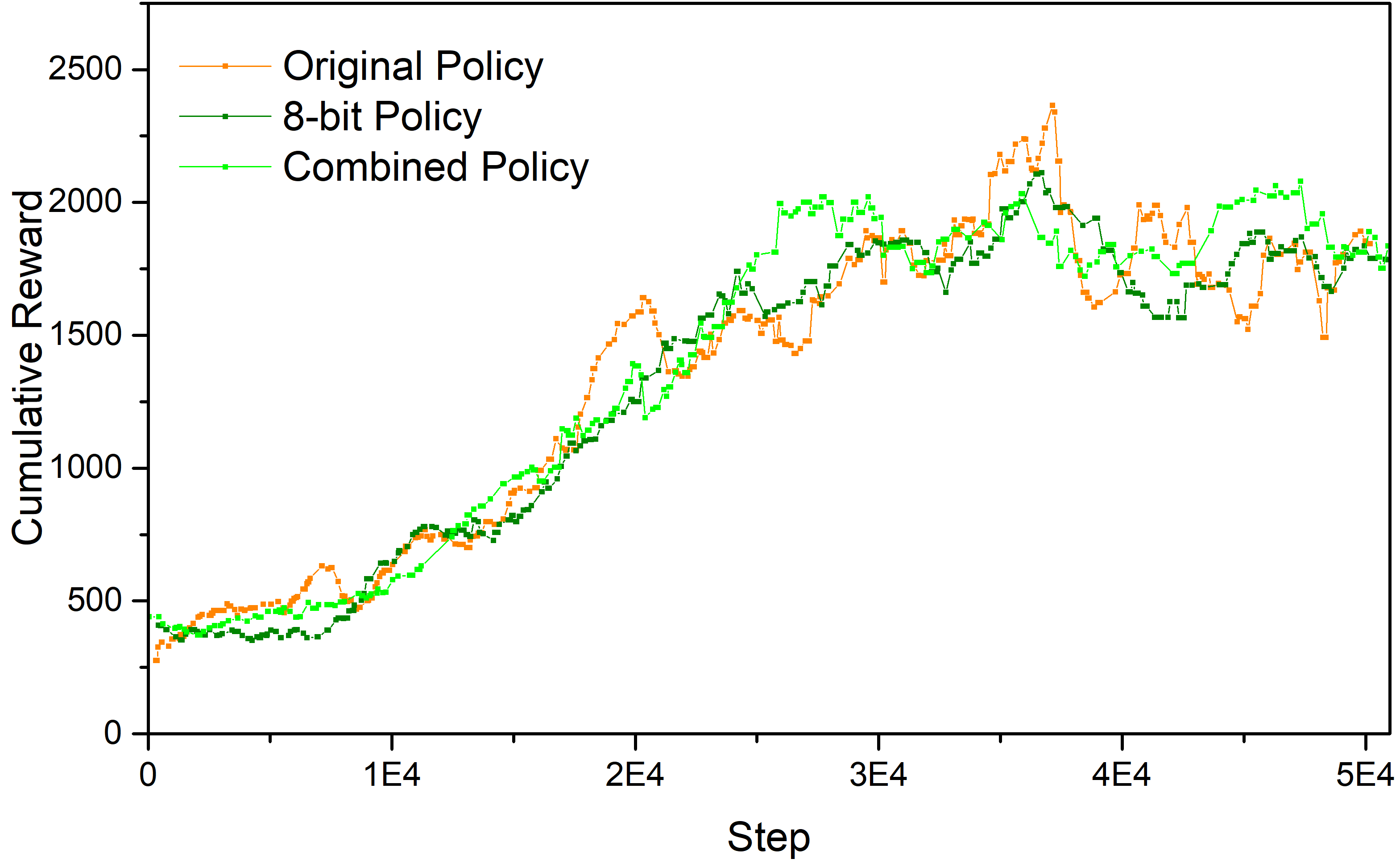}}
	\subfigure[beam-time]{\label{fig:comb:6}
		\includegraphics[width=0.25\linewidth]{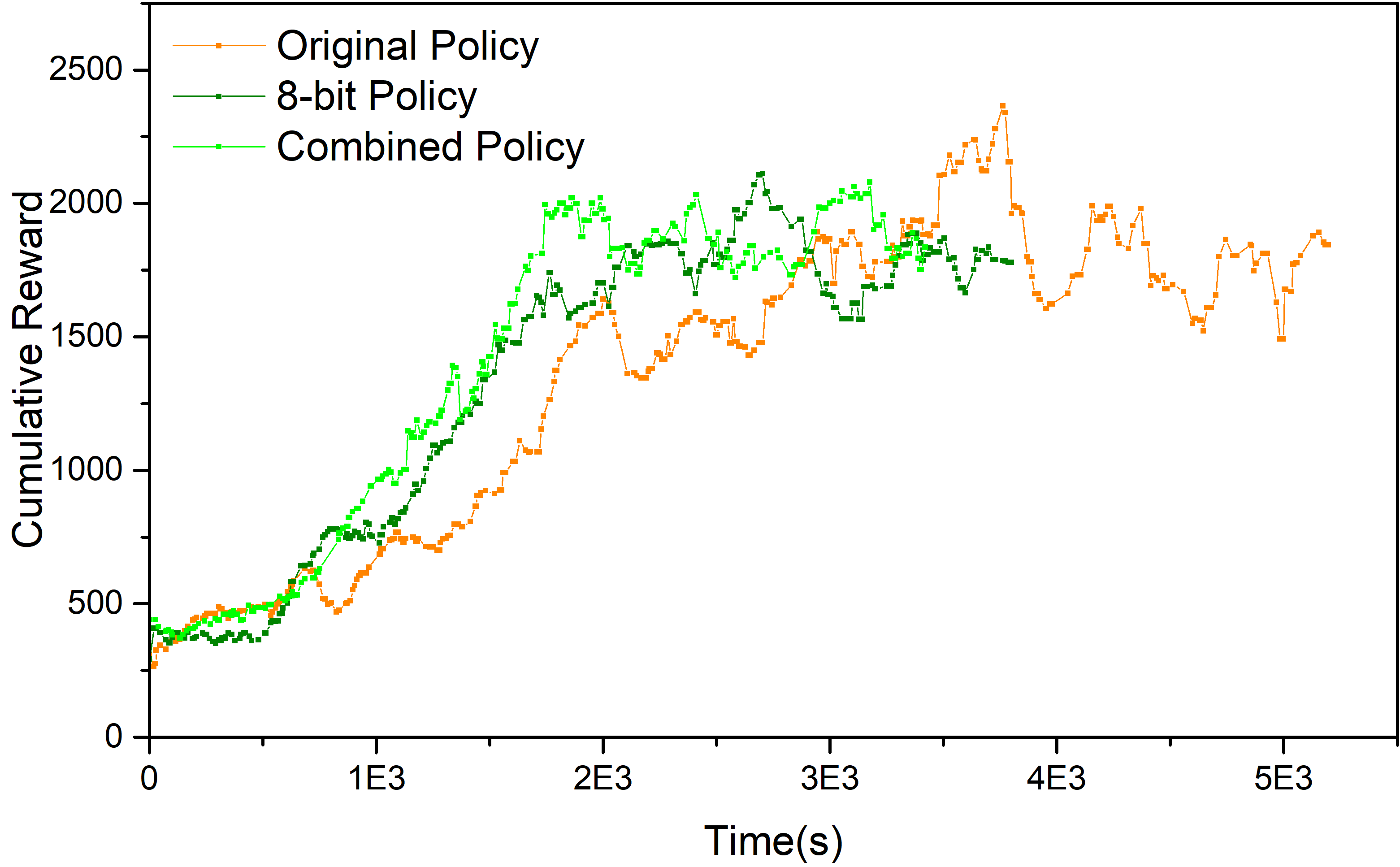}}
	\subfigure[hero-step]{\label{fig:comb:7}
		\includegraphics[width=0.245\linewidth]{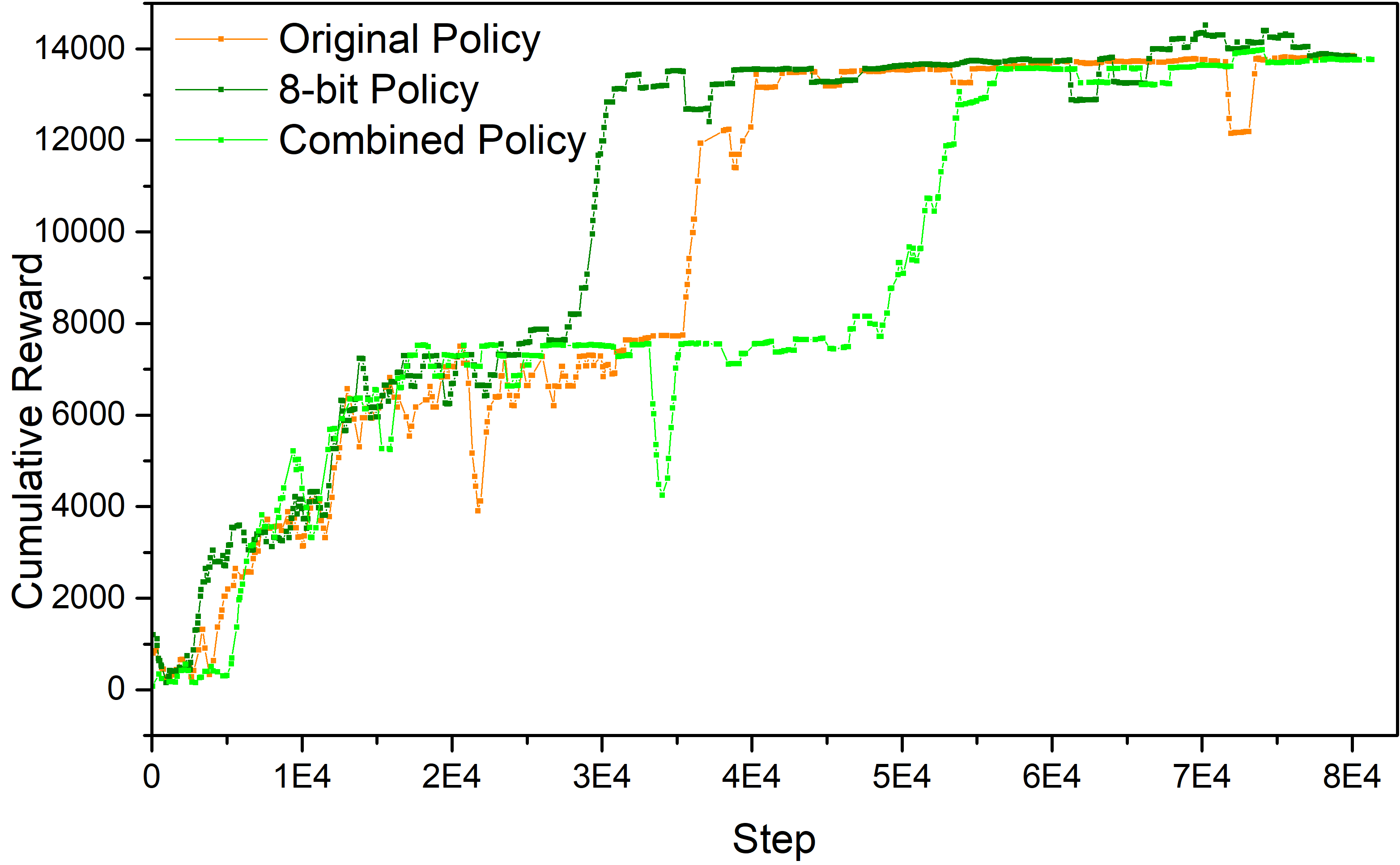}}
	\subfigure[hero-time]{\label{fig:comb:8}
		\includegraphics[width=0.245\linewidth]{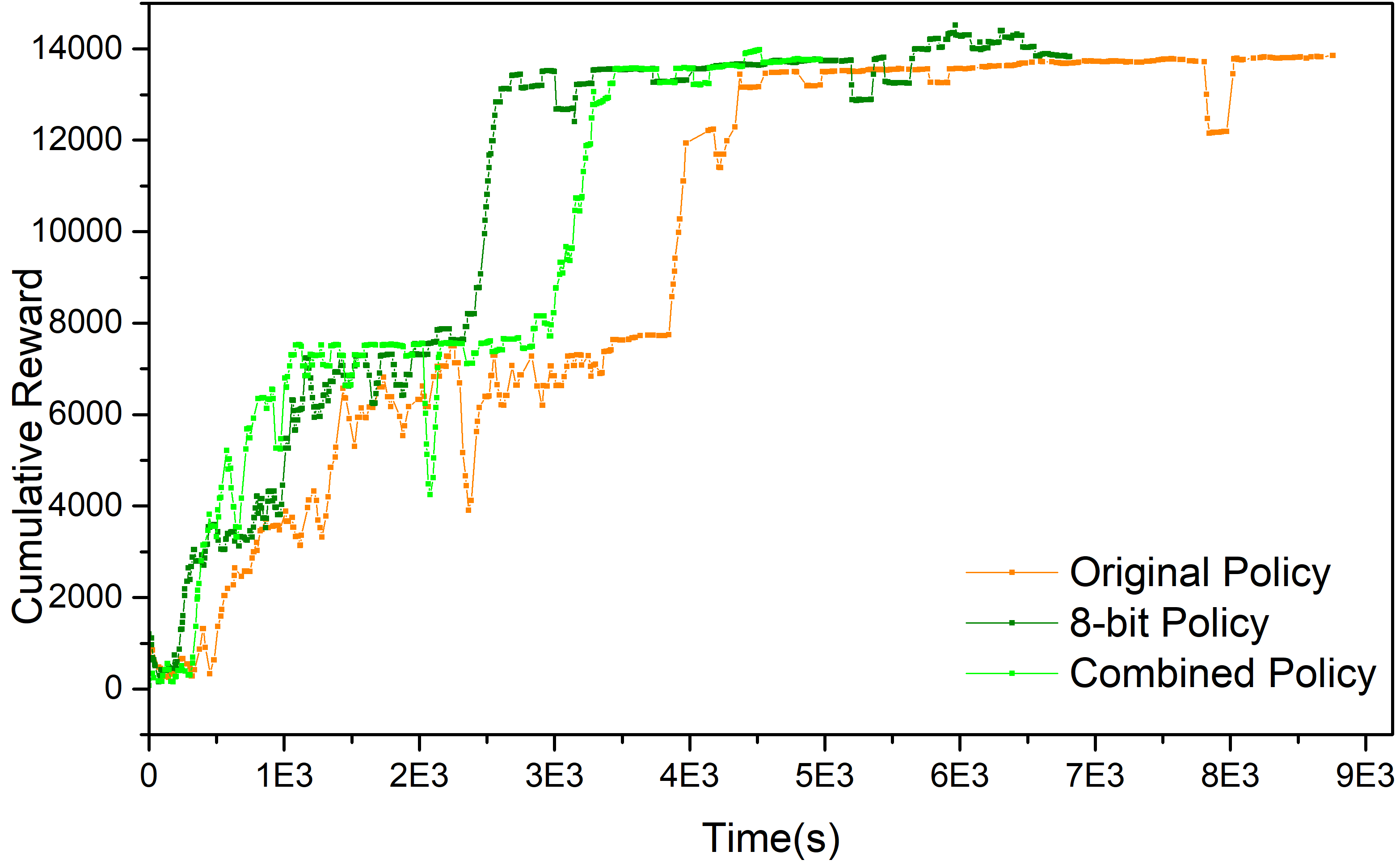}}
		
	\subfigure[name-step]{\label{fig:comb:9}
		\includegraphics[width=0.245\linewidth]{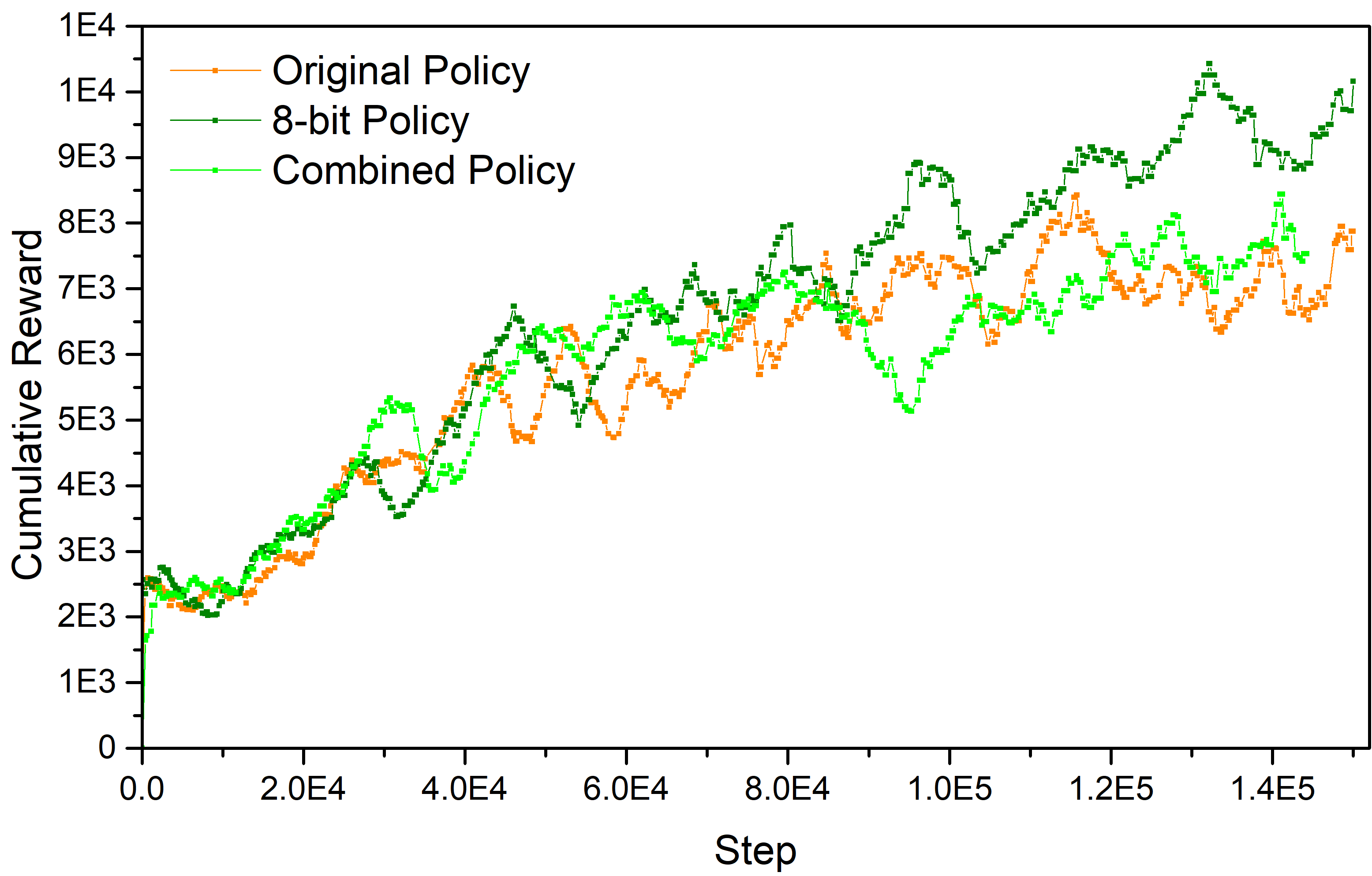}}
	\subfigure[name-time]{\label{fig:comb:10}
		\includegraphics[width=0.24\linewidth]{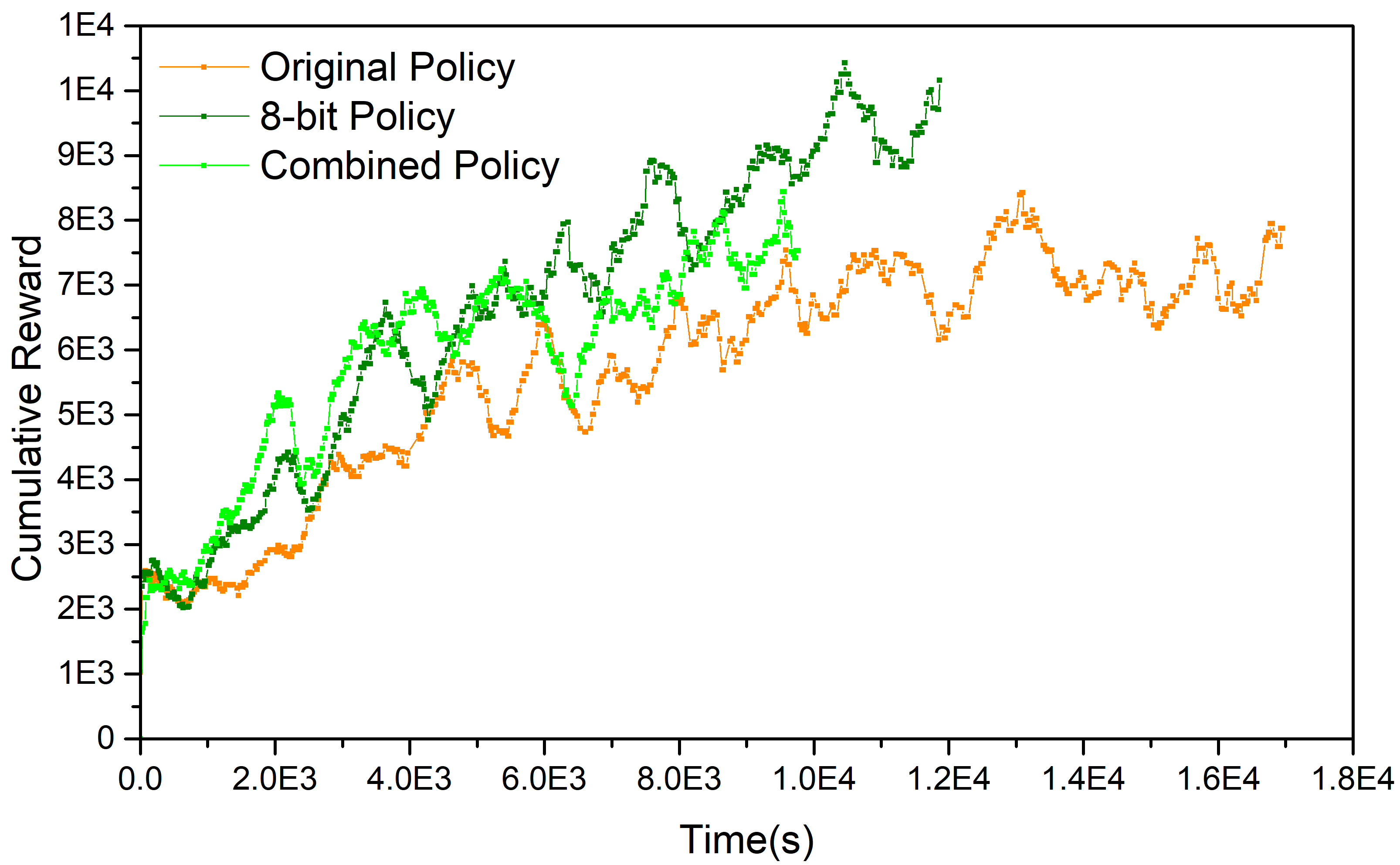}}
	\subfigure[sea-step]{\label{fig:comb:11}
		\includegraphics[width=0.245\linewidth]{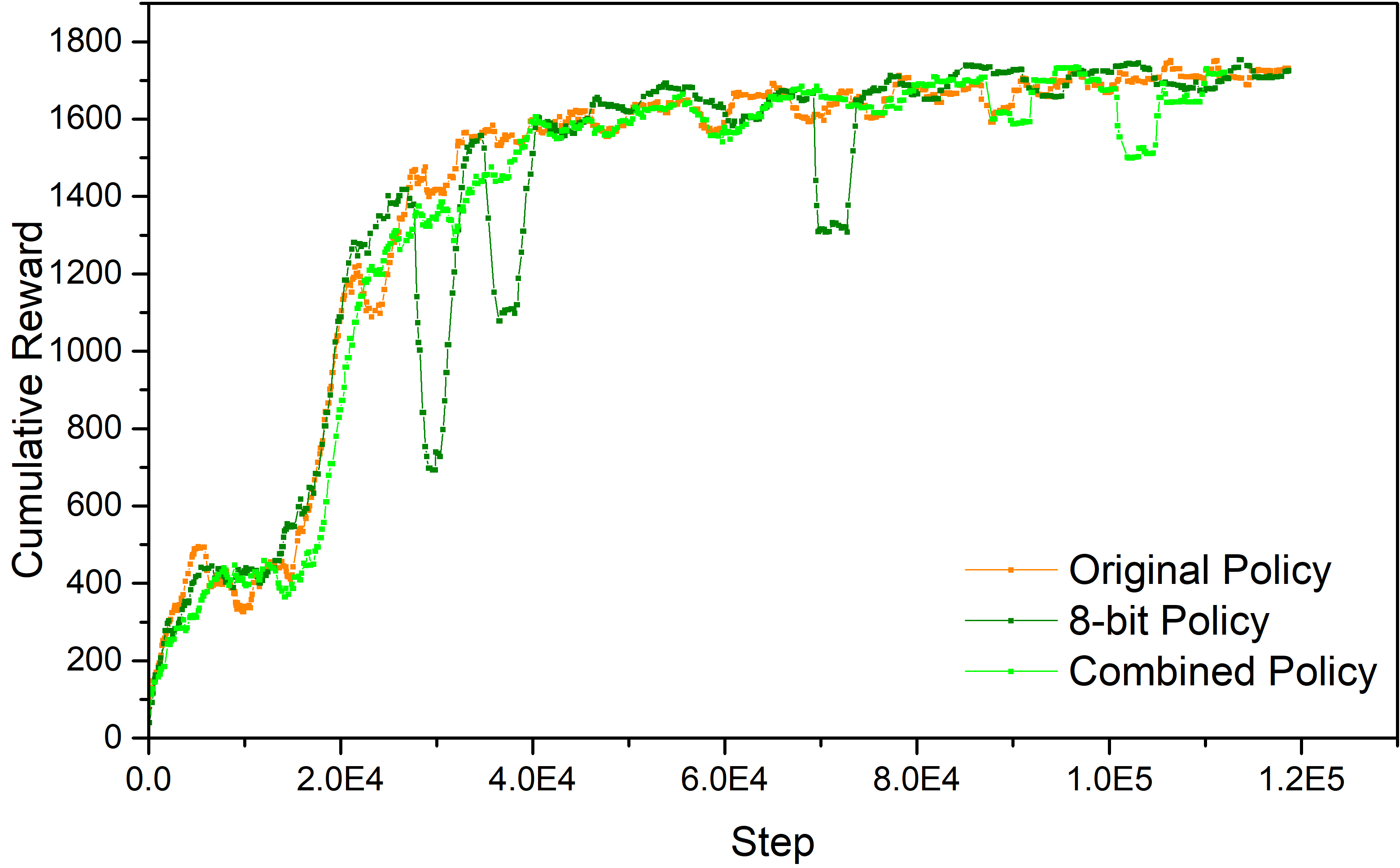}}
	\subfigure[sea-time]{\label{fig:comb:12}
		\includegraphics[width=0.245\linewidth]{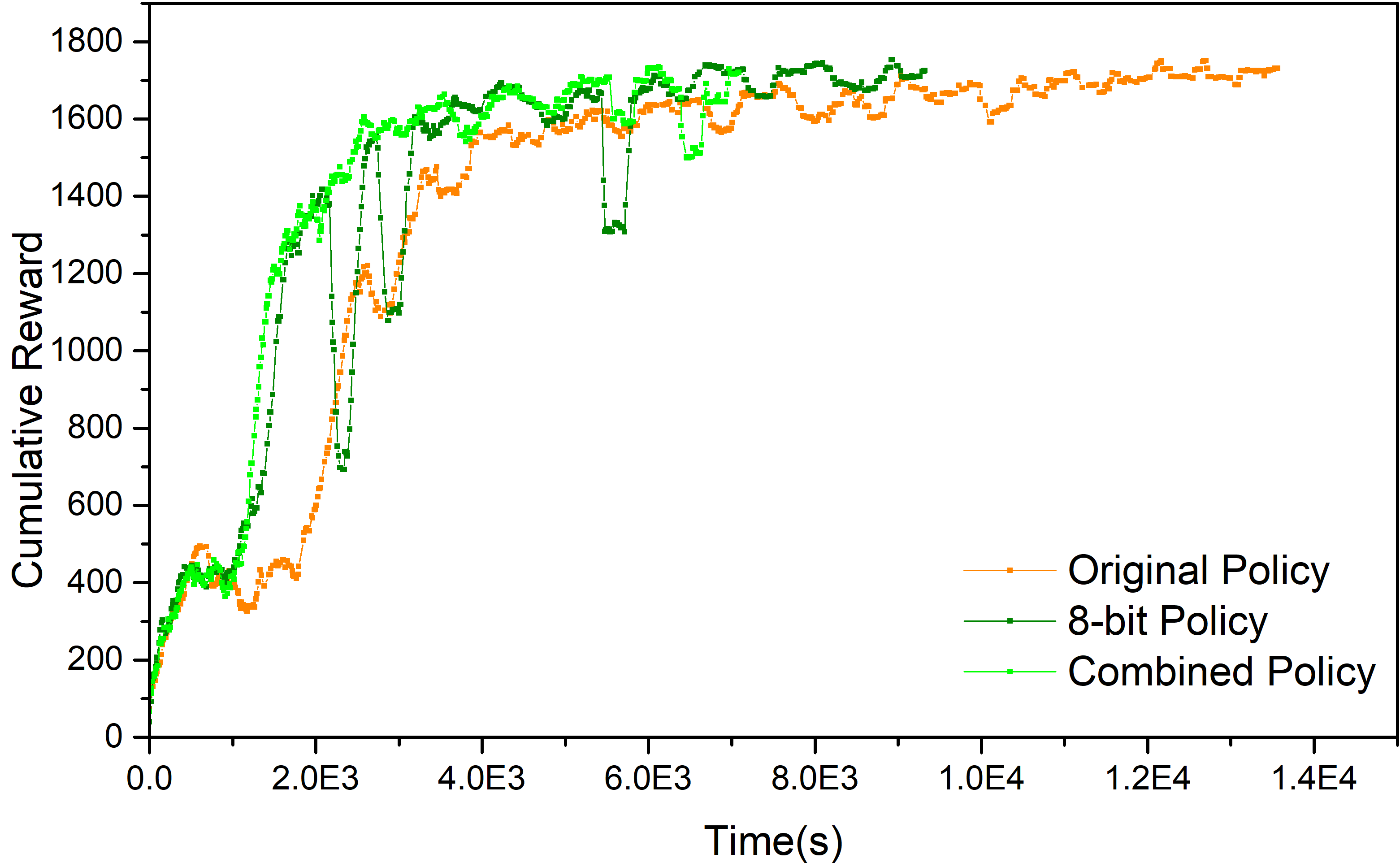}}
	\caption{The Multi-Compression Methods results with $NA=12$ and $ND=4$.}
	\label{fig:comb}
\end{figure}

\begin{figure}[H]
	\centering
	\subfigbottomskip=2pt
	\subfigcapskip=-5pt
	\subfigure[alien-step]{\label{fig:quant:1}
		\includegraphics[width=0.25\linewidth]{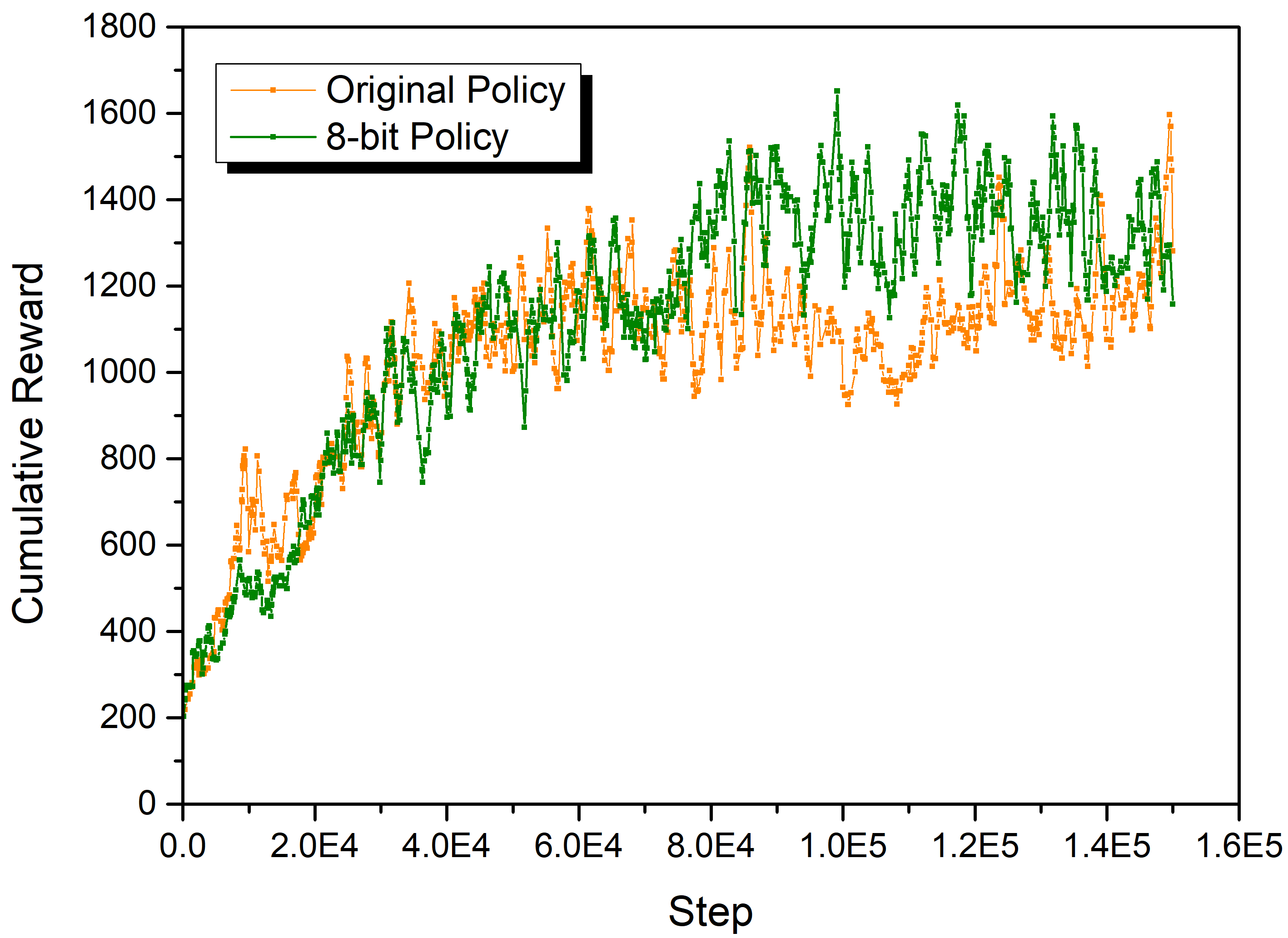}}
	\subfigure[alien-time]{\label{fig:quant:2}
		\includegraphics[width=0.25\linewidth]{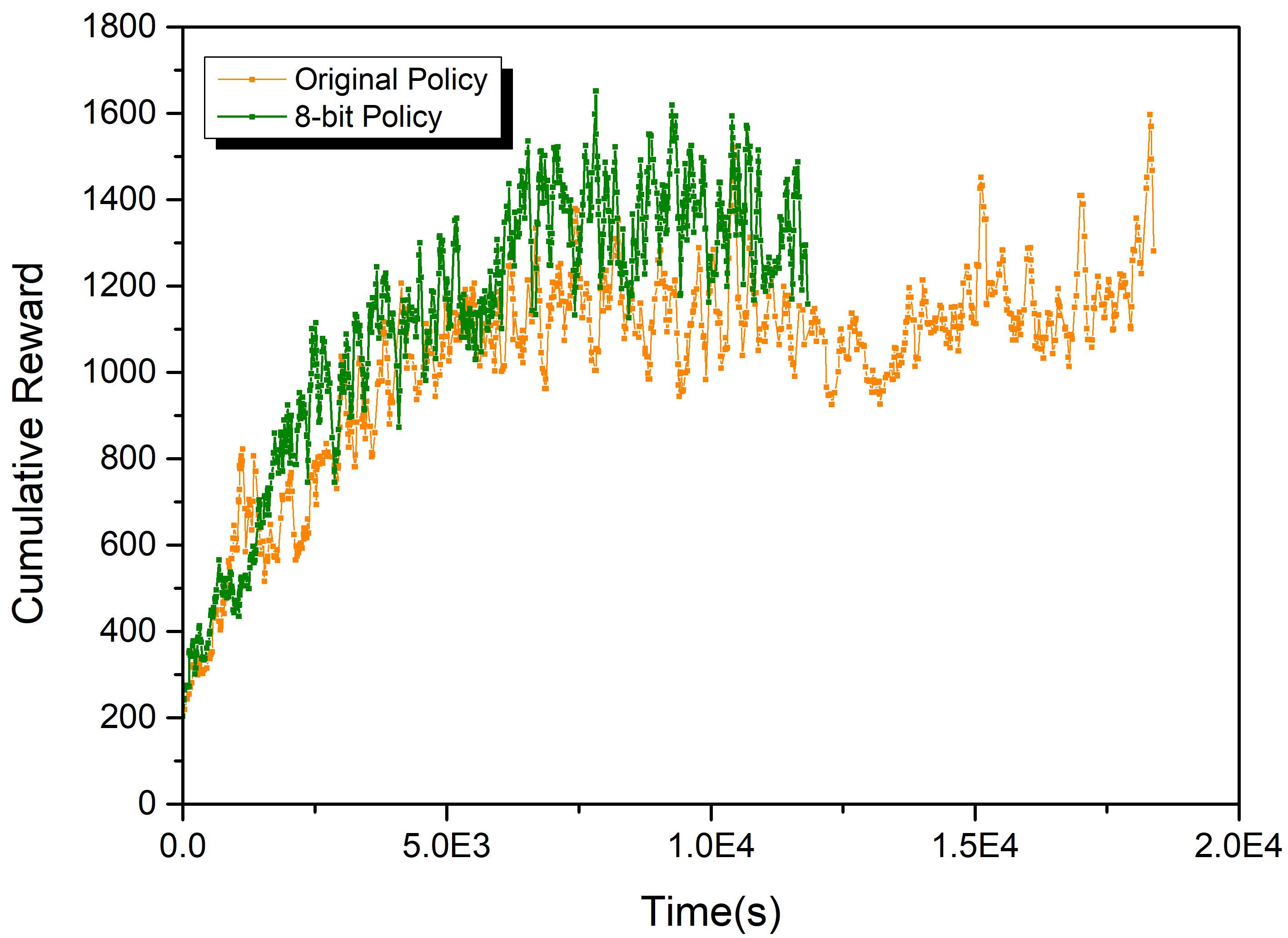}}
	\subfigure[amidar-step]{\label{fig:quant:3}
		\includegraphics[width=0.24\linewidth]{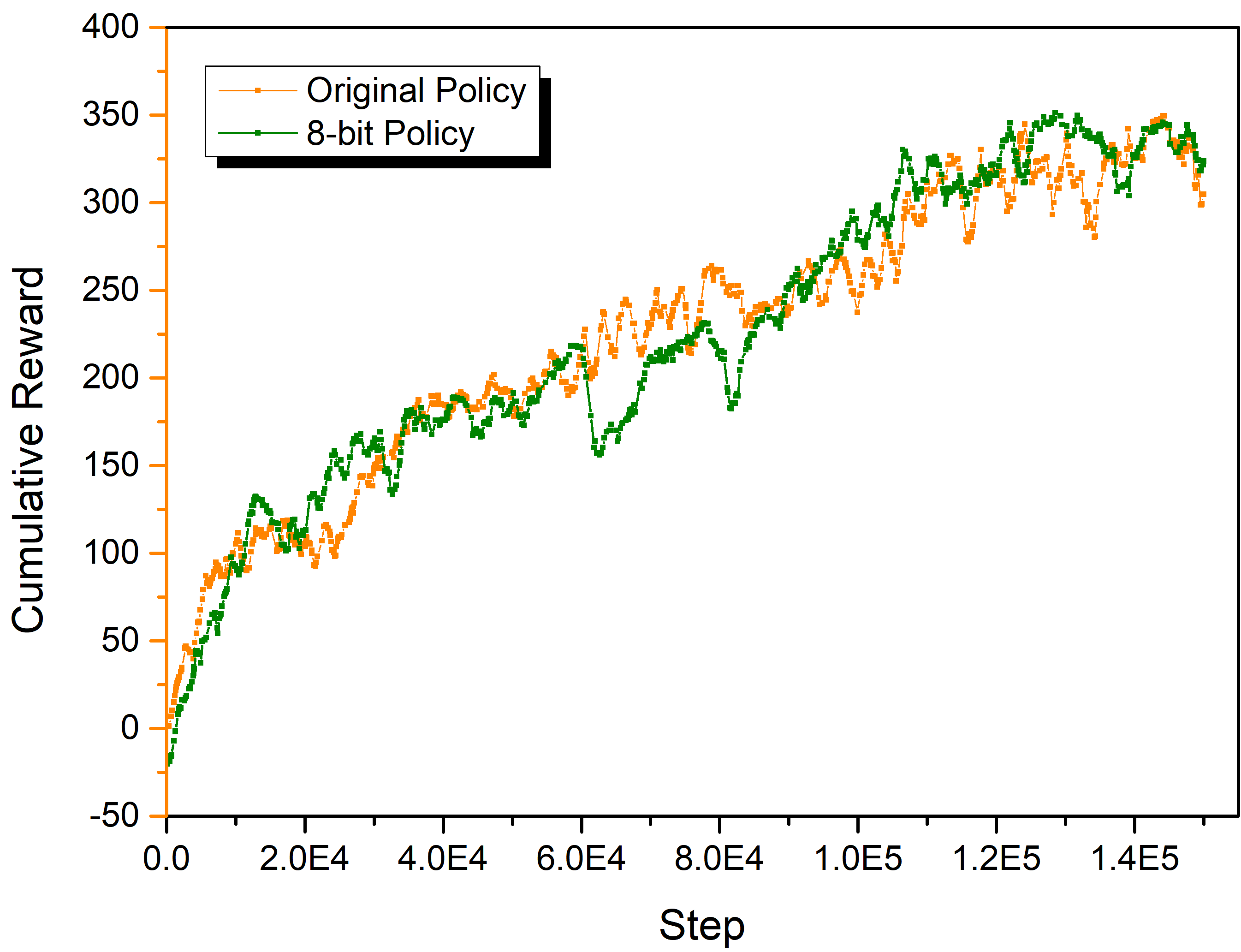}}
	\subfigure[amidar-time]{\label{fig:quant:4}
		\includegraphics[width=0.24\linewidth]{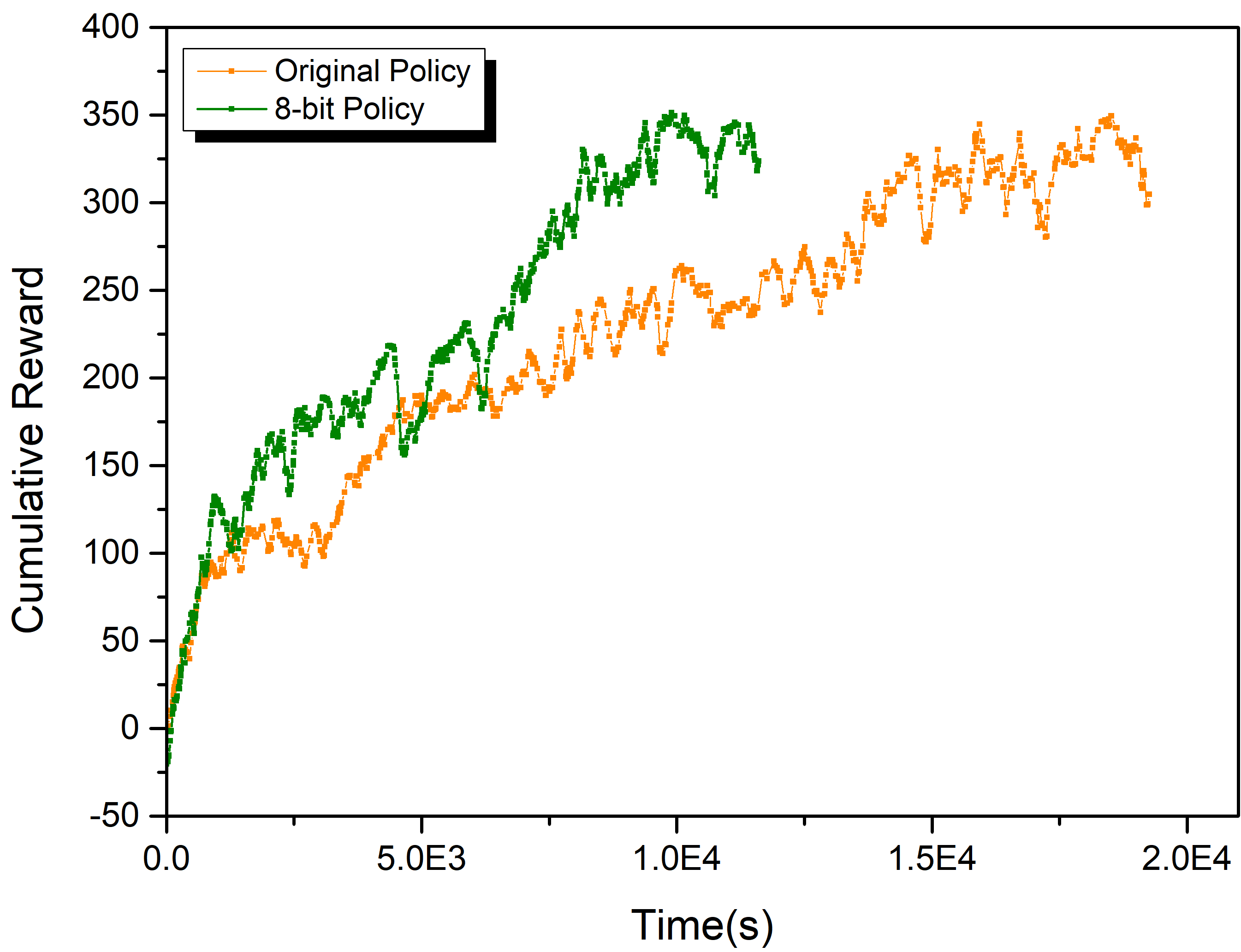}}
		
	\subfigure[bank-step]{\label{fig:quant:5}
		\includegraphics[width=0.24\linewidth]{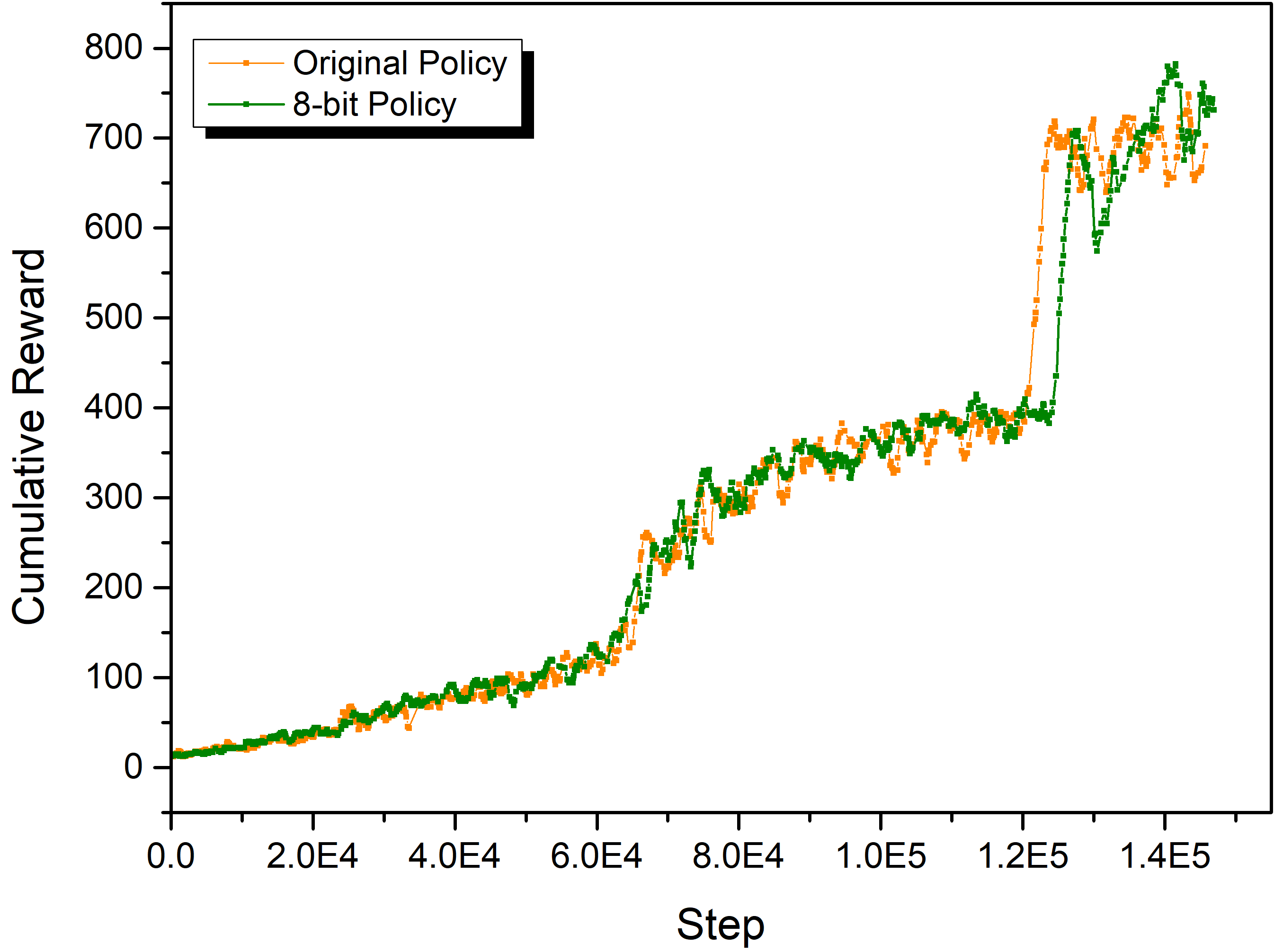}}
	\subfigure[bank-time]{\label{fig:quant:6}
		\includegraphics[width=0.25\linewidth]{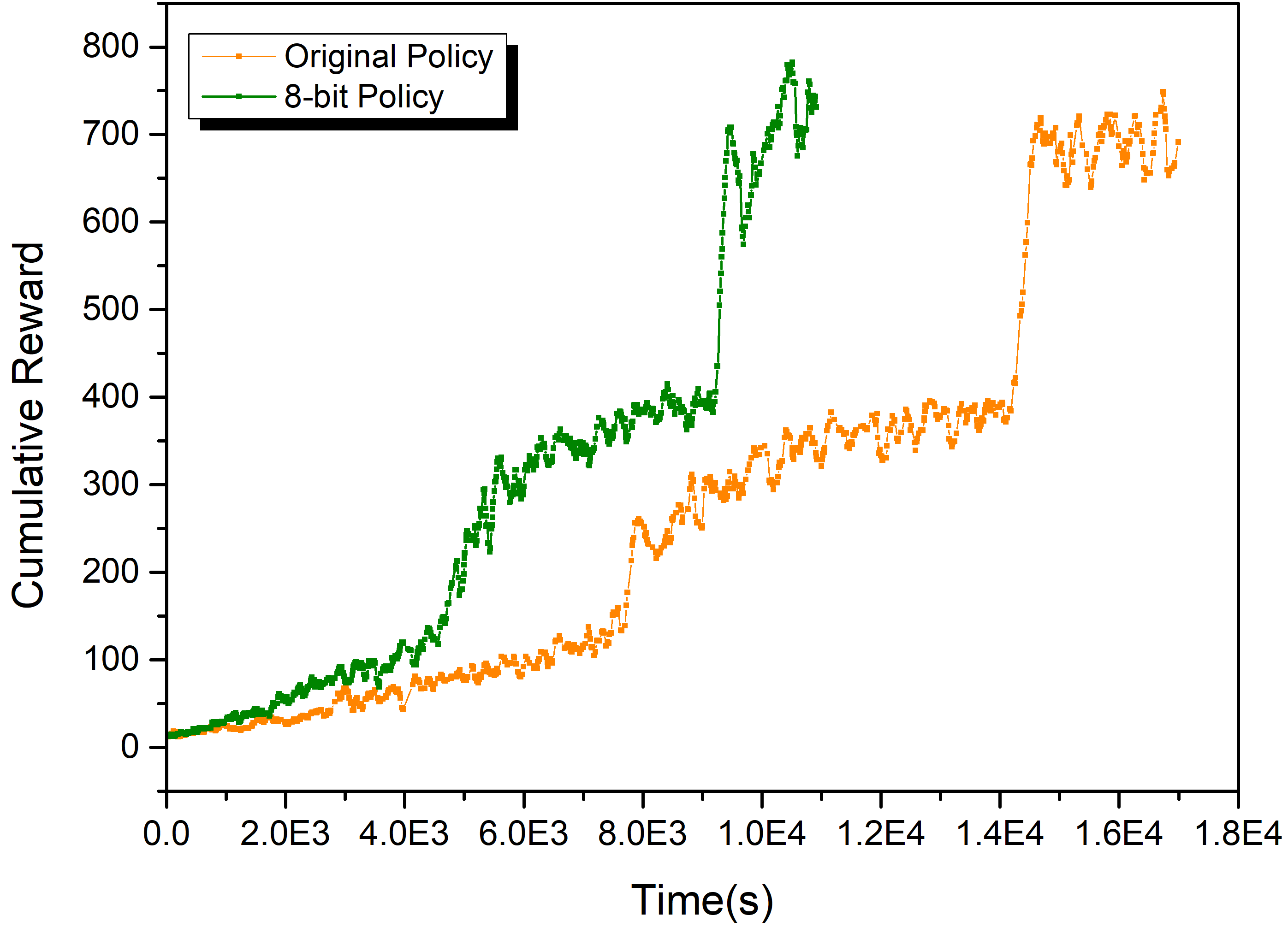}}
	\subfigure[beam-step]{\label{fig:quant:7}
		\includegraphics[width=0.245\linewidth]{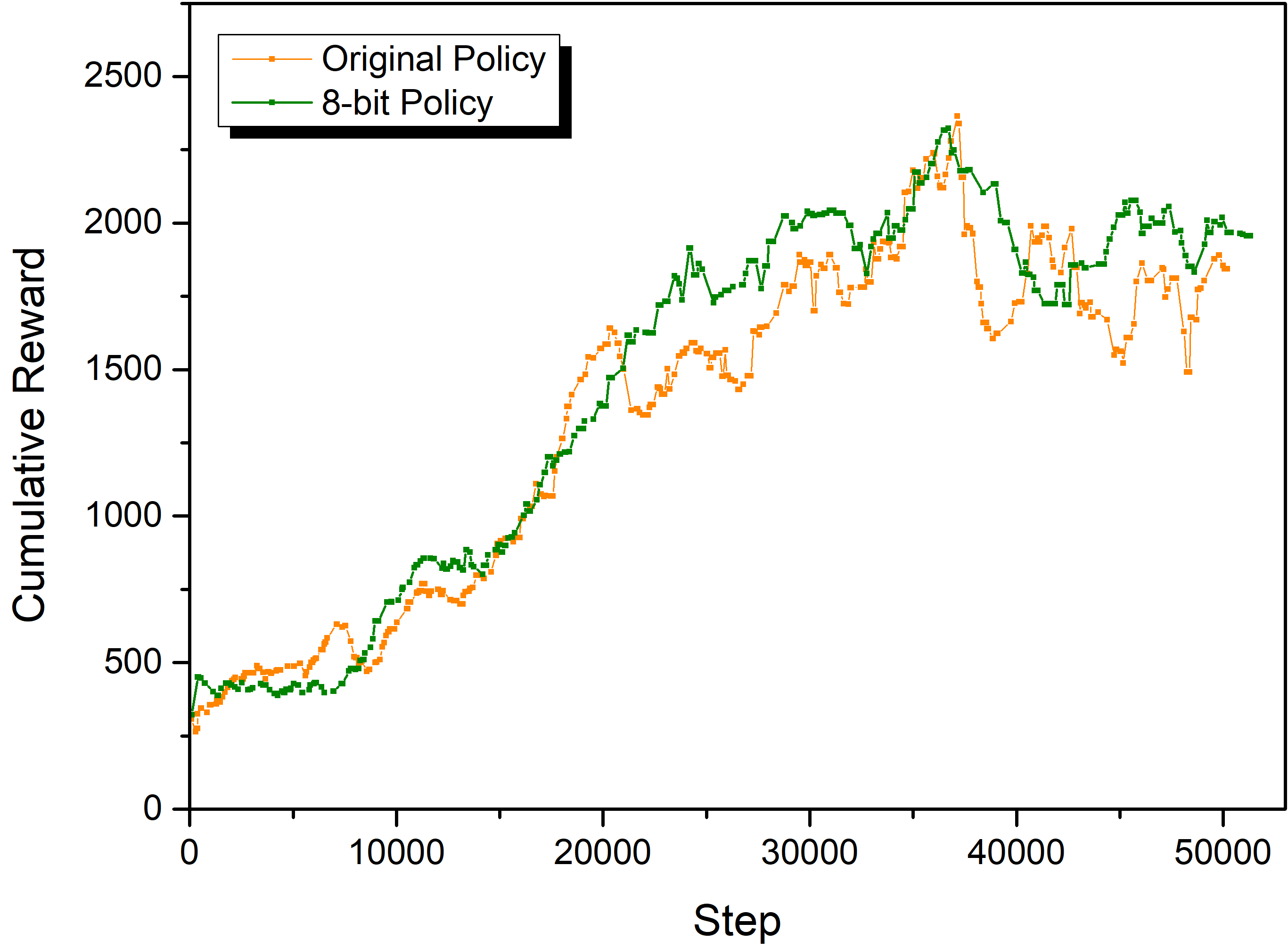}}
	\subfigure[beam-time]{\label{fig:quant:8}
		\includegraphics[width=0.245\linewidth]{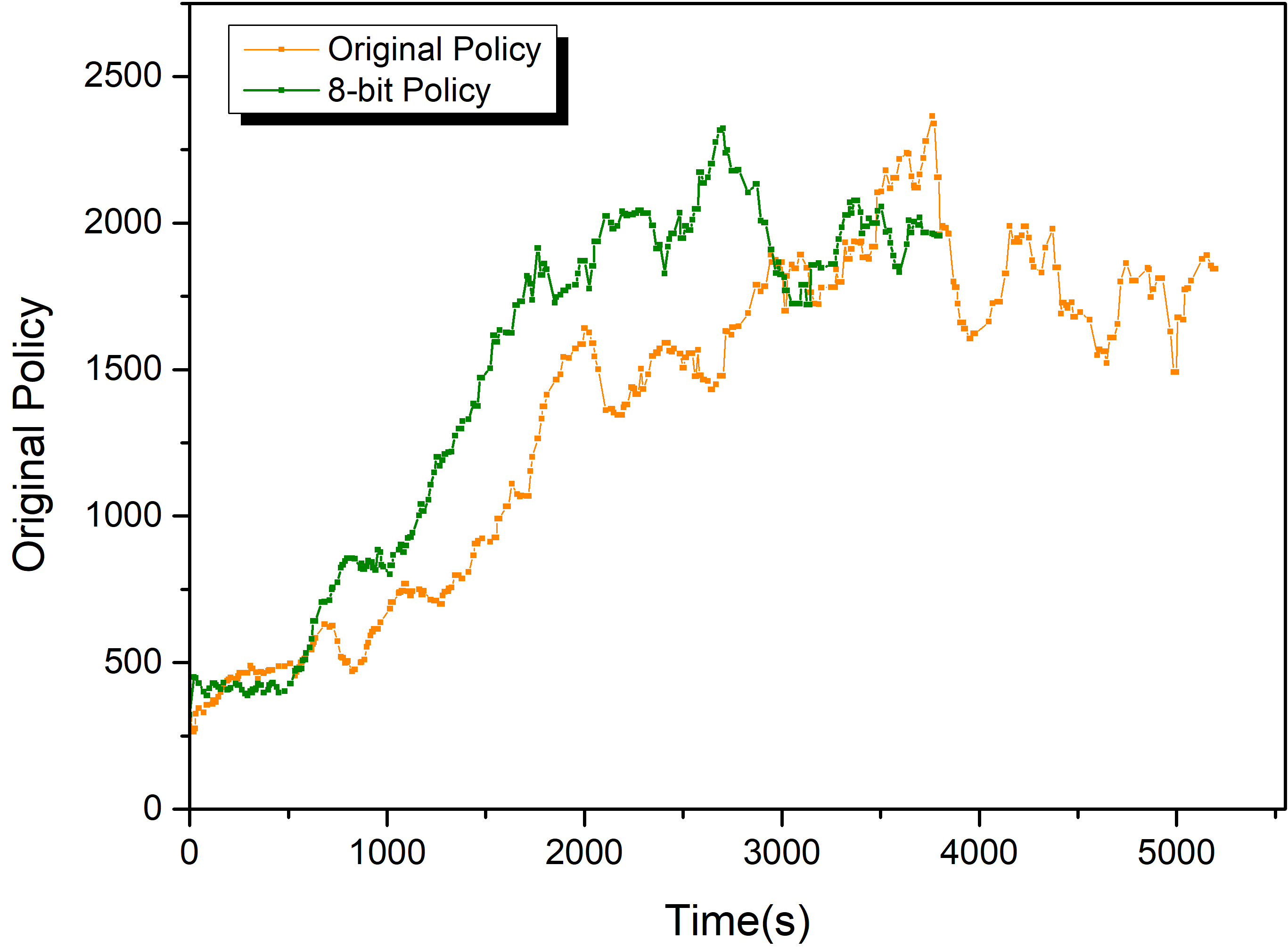}}
		
	\subfigure[break-step]{\label{fig:quant:9}
		\includegraphics[width=0.25\linewidth]{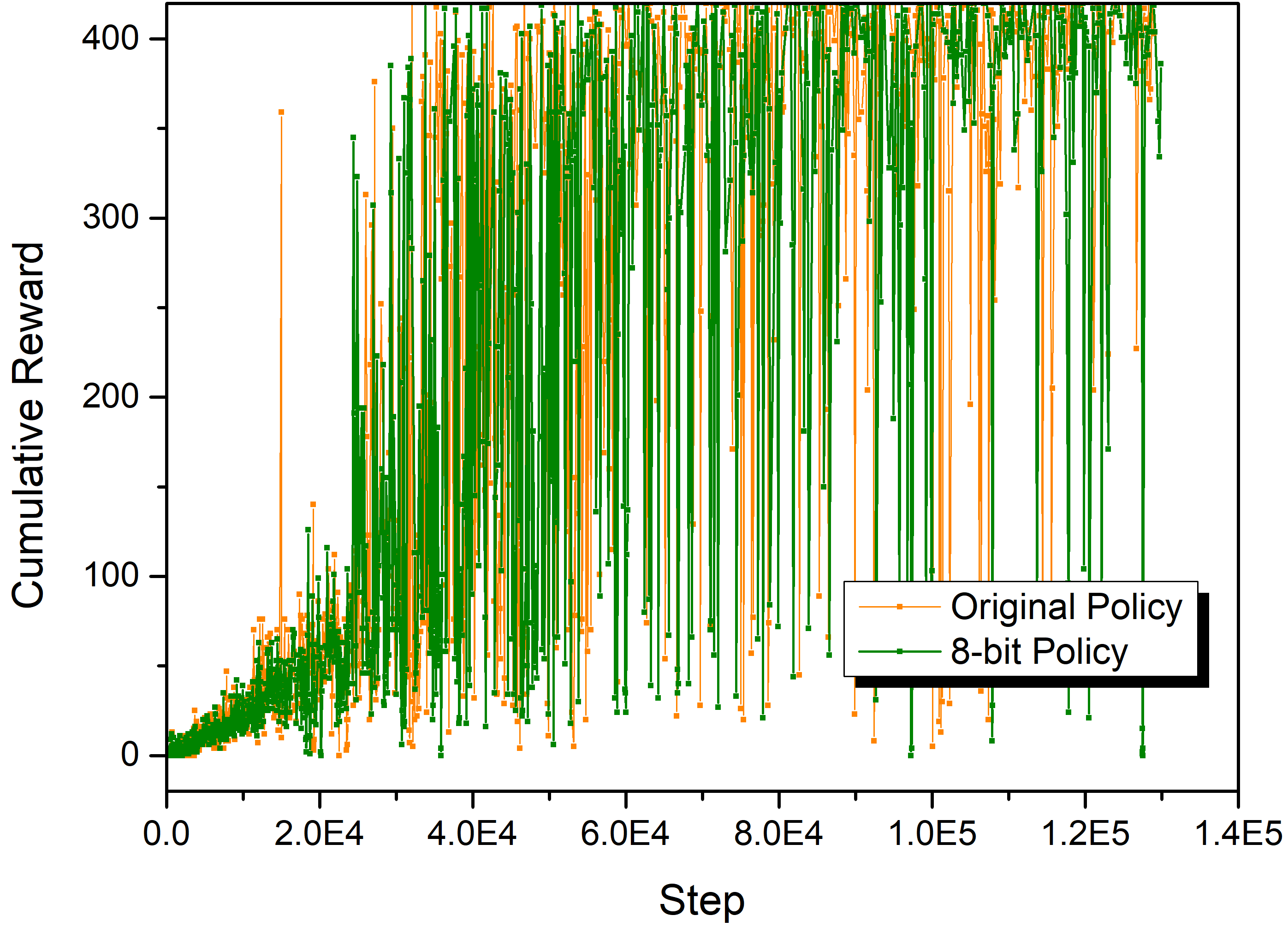}}
	\subfigure[break-time]{\label{fig:quant:10}
		\includegraphics[width=0.24\linewidth]{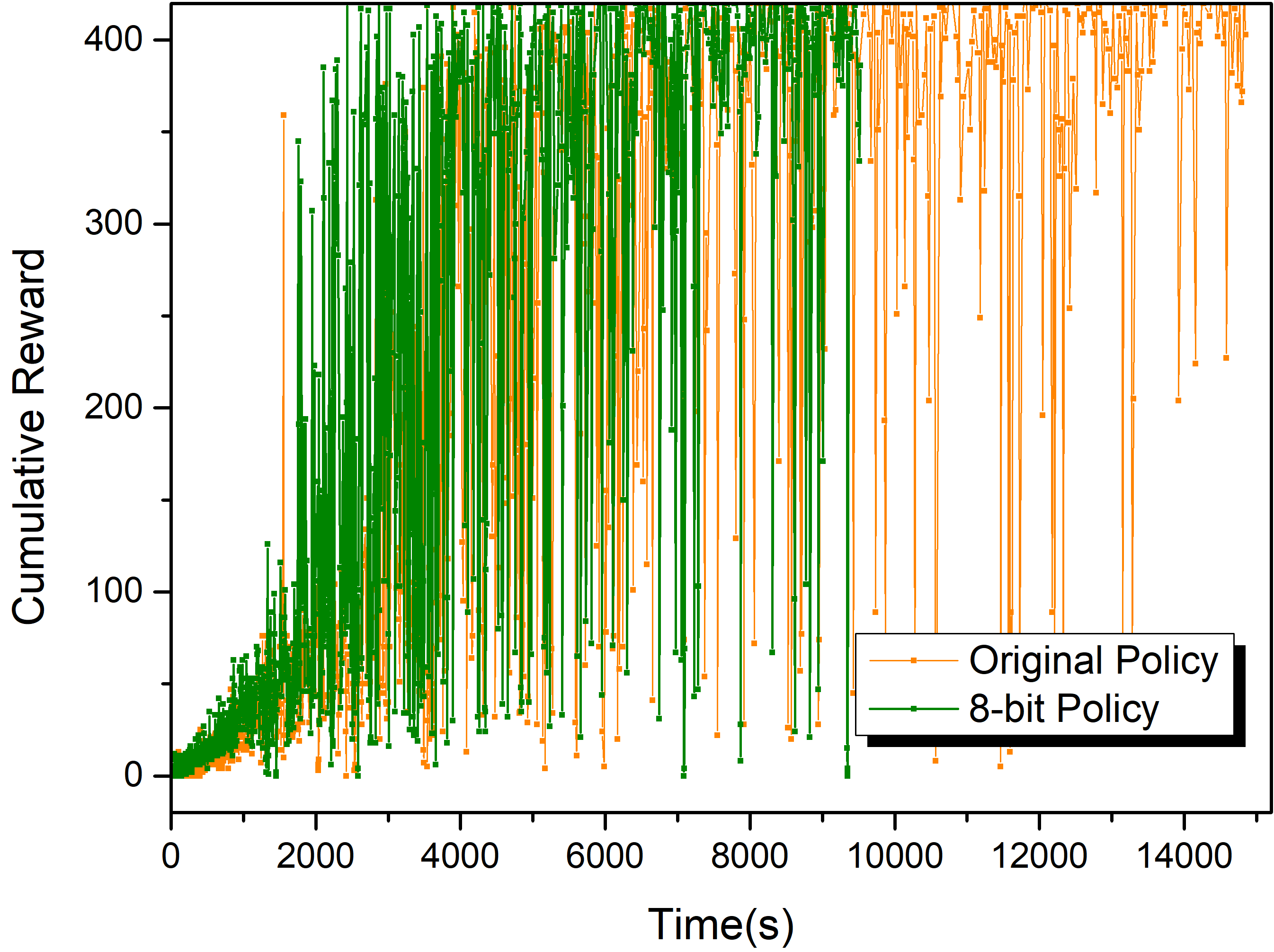}}
	\subfigure[hero-step]{\label{fig:quant:11}
		\includegraphics[width=0.25\linewidth]{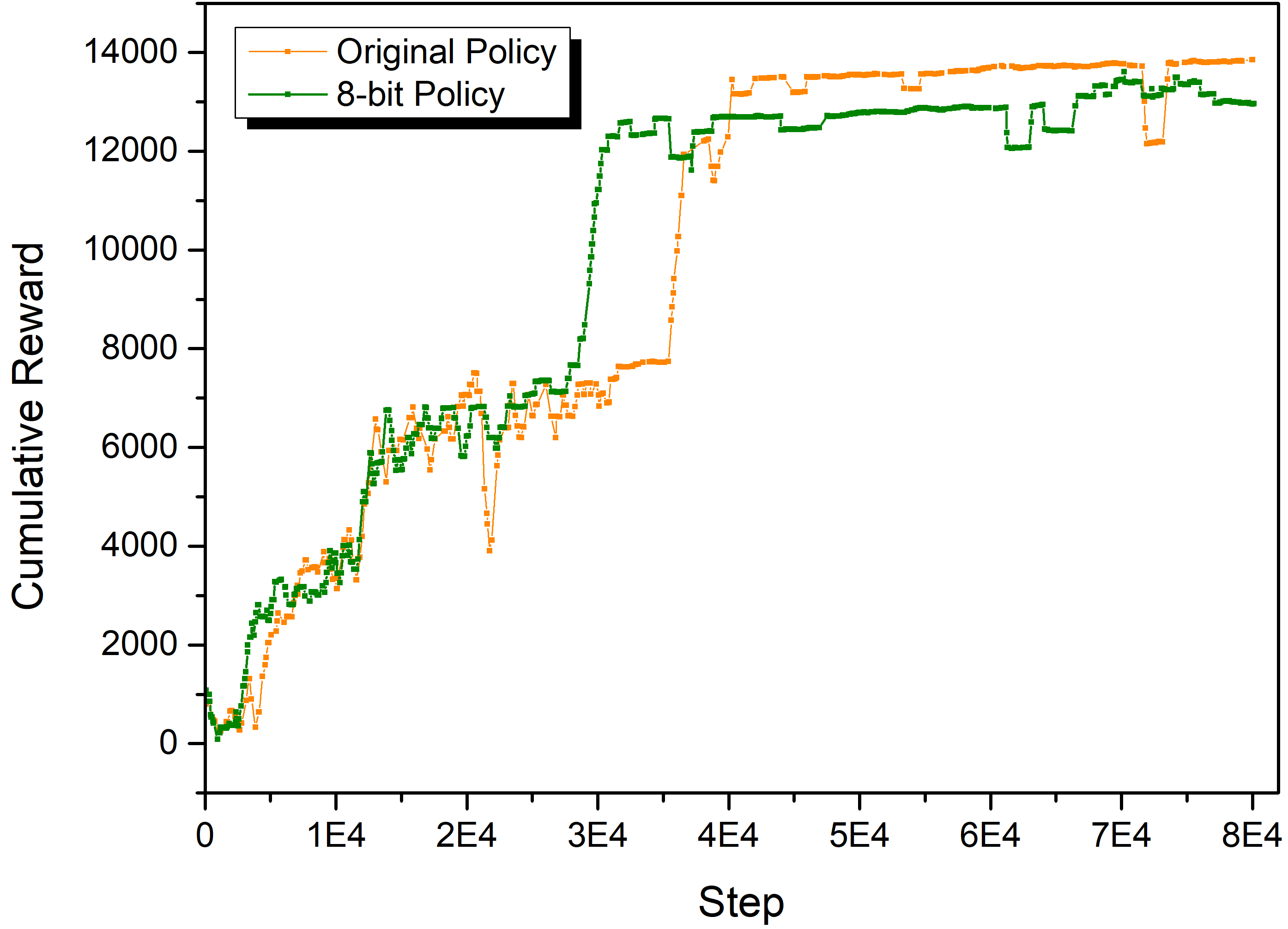}}
	\subfigure[hero-time]{\label{fig:quant:12}
		\includegraphics[width=0.245\linewidth]{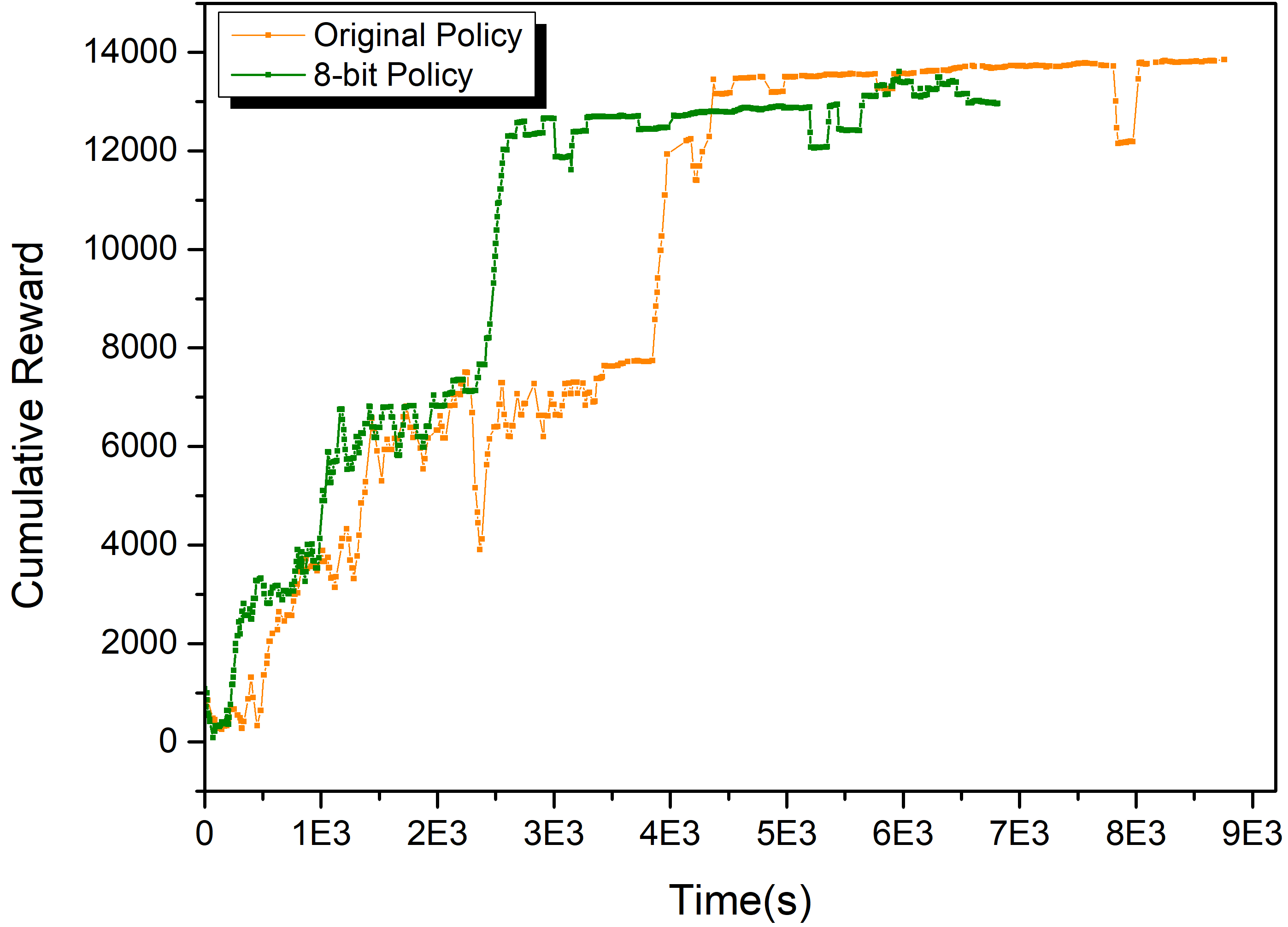}}
		
	\subfigure[name-step]{\label{fig:quant:13}
		\includegraphics[width=0.25\linewidth]{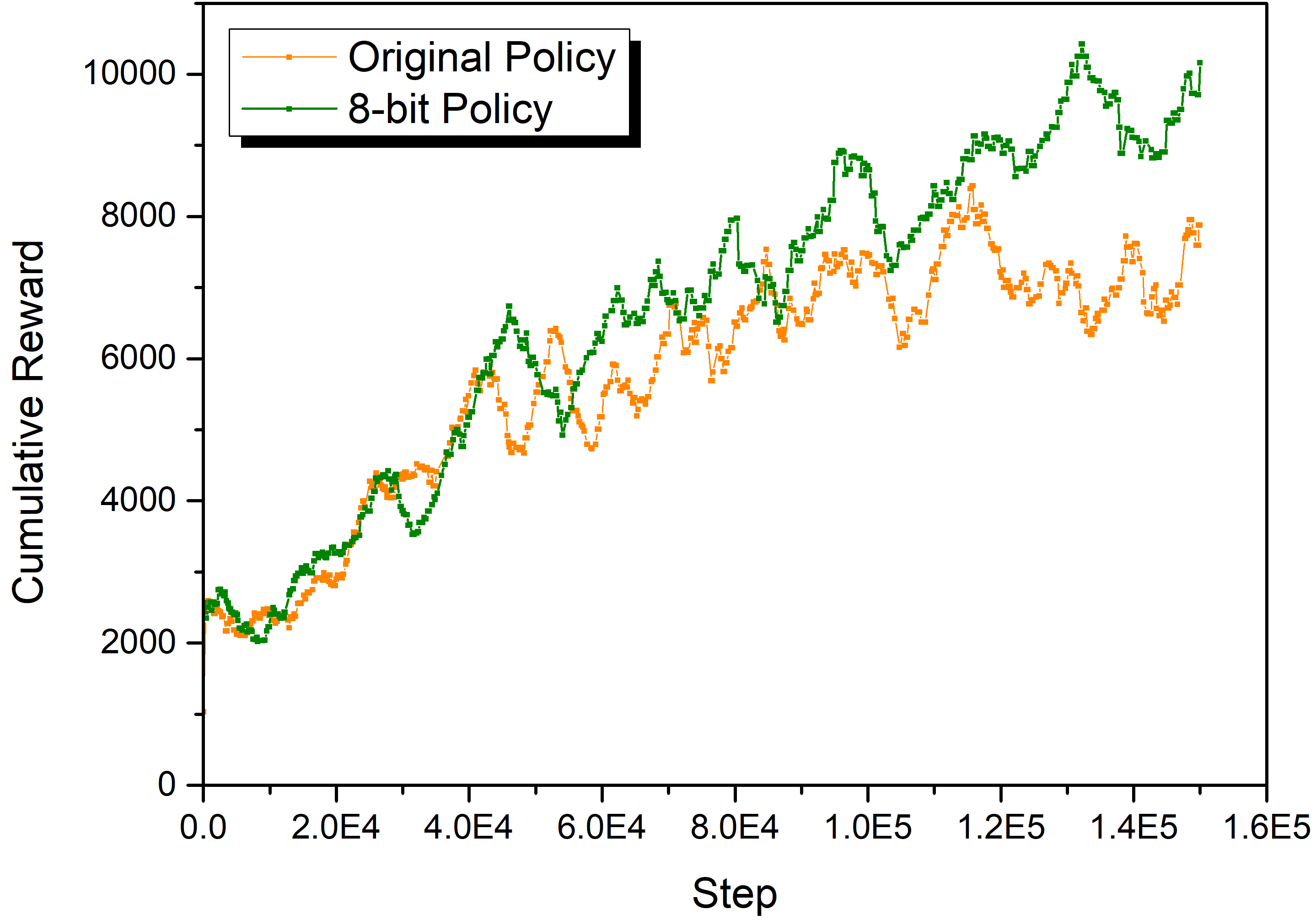}}
	\subfigure[name-time]{\label{fig:quant:14}
		\includegraphics[width=0.245\linewidth]{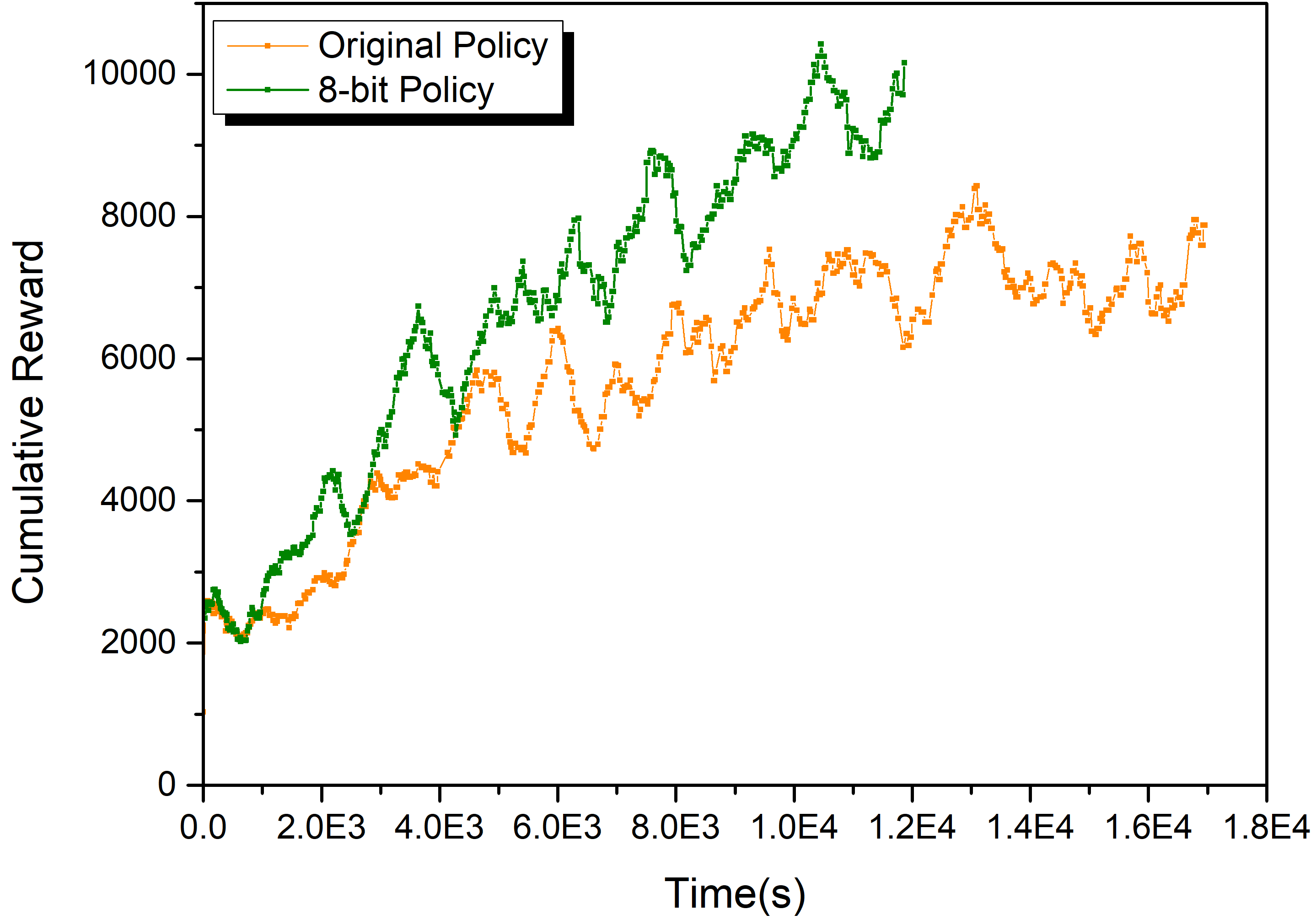}}
	\subfigure[pong-step]{\label{fig:quant:15}
		\includegraphics[width=0.245\linewidth]{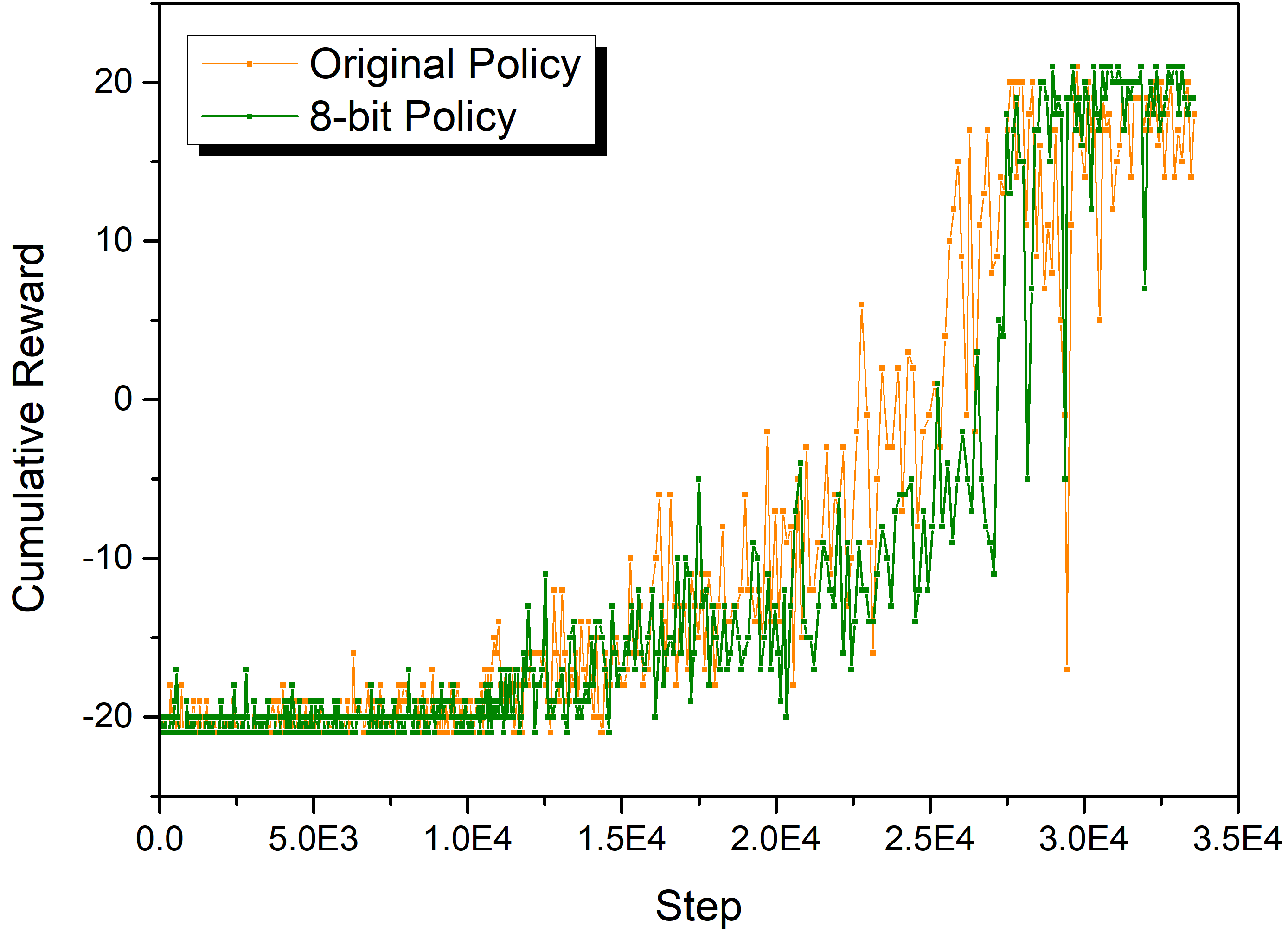}}
	\subfigure[pong-time]{\label{fig:quant:16}
		\includegraphics[width=0.245\linewidth]{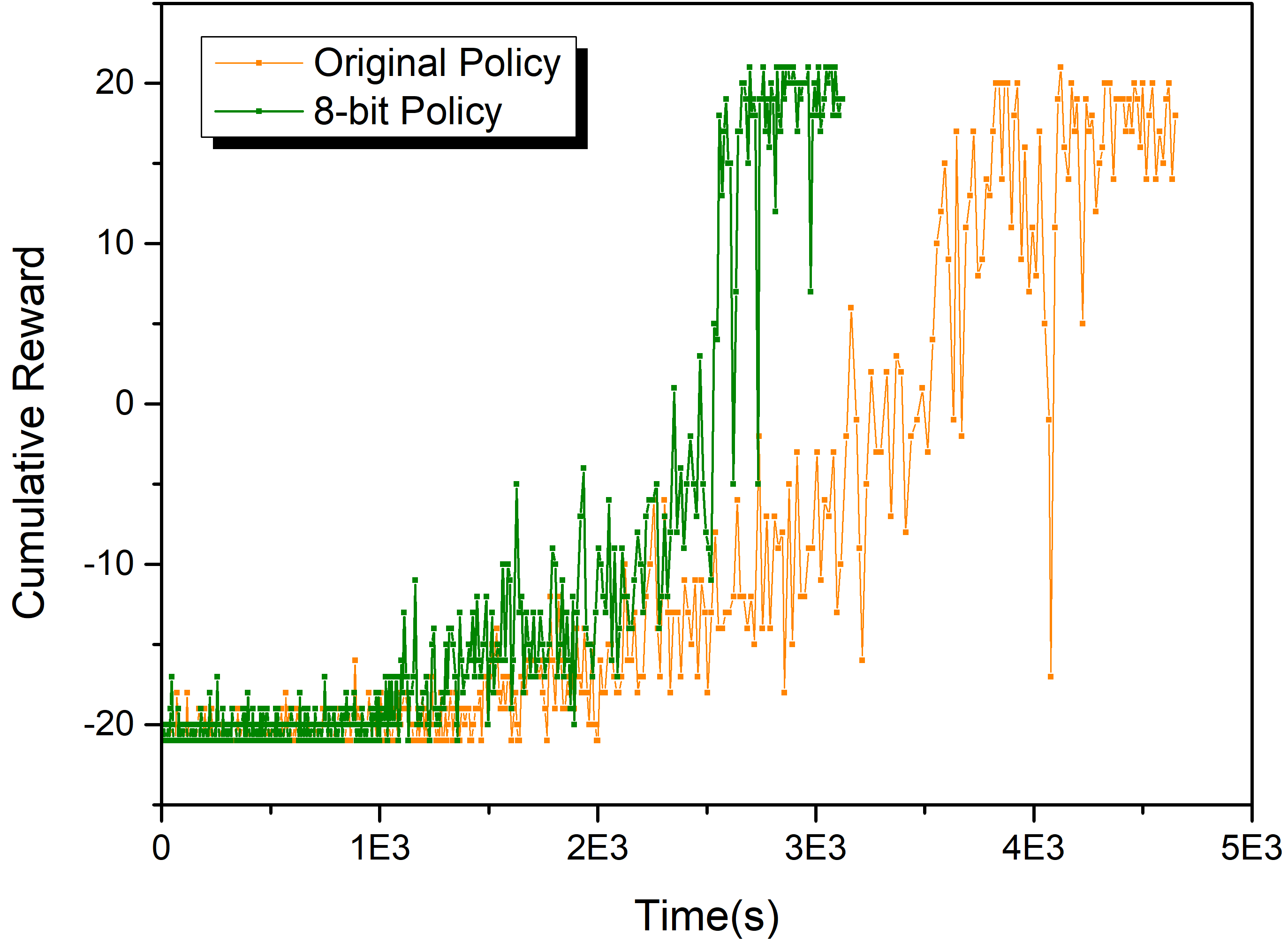}}
		
	\subfigure[river-step]{\label{fig:quant:17}
		\includegraphics[width=0.245\linewidth]{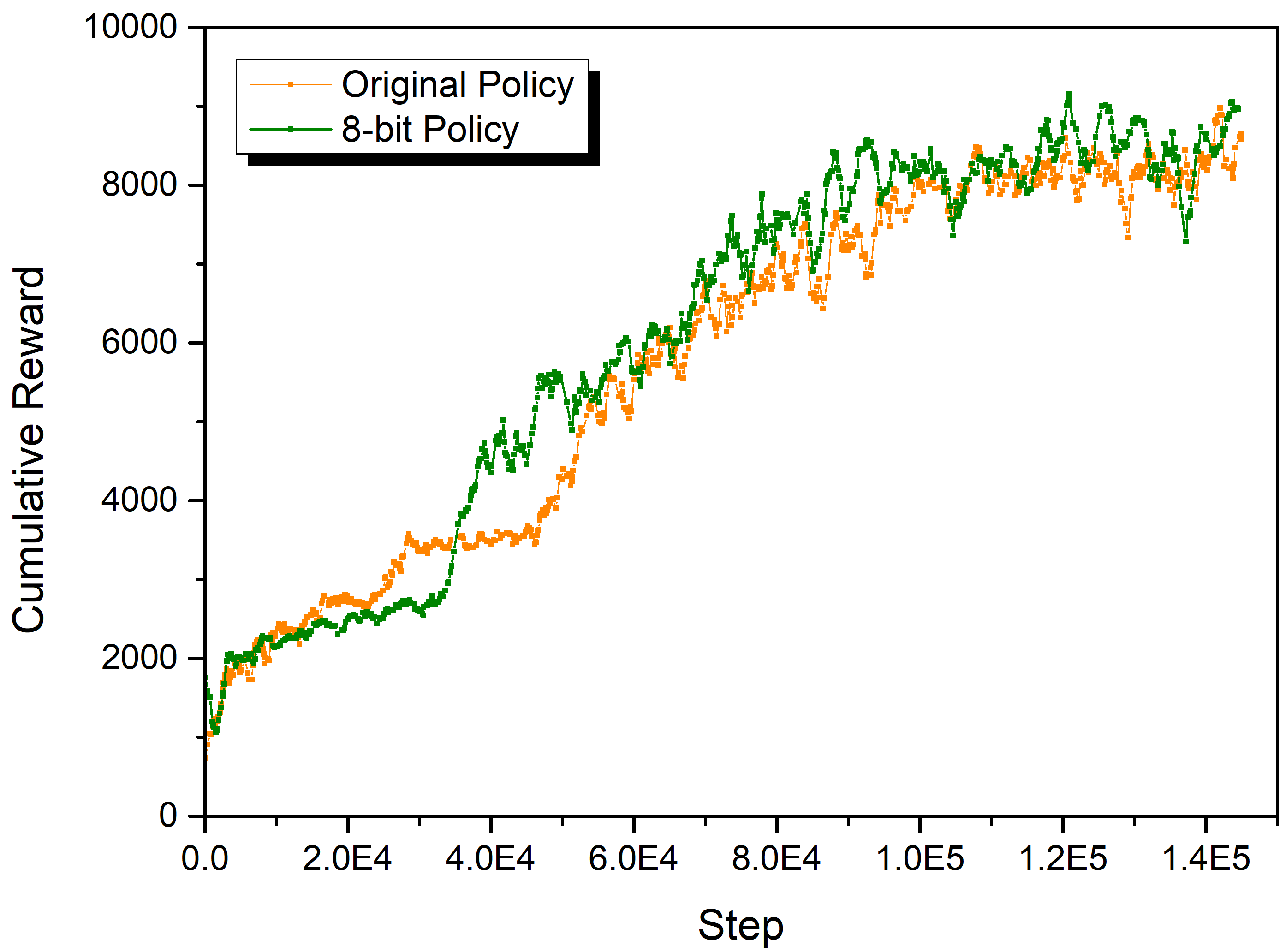}}
	\subfigure[river-time]{\label{fig:quant:18}
		\includegraphics[width=0.245\linewidth]{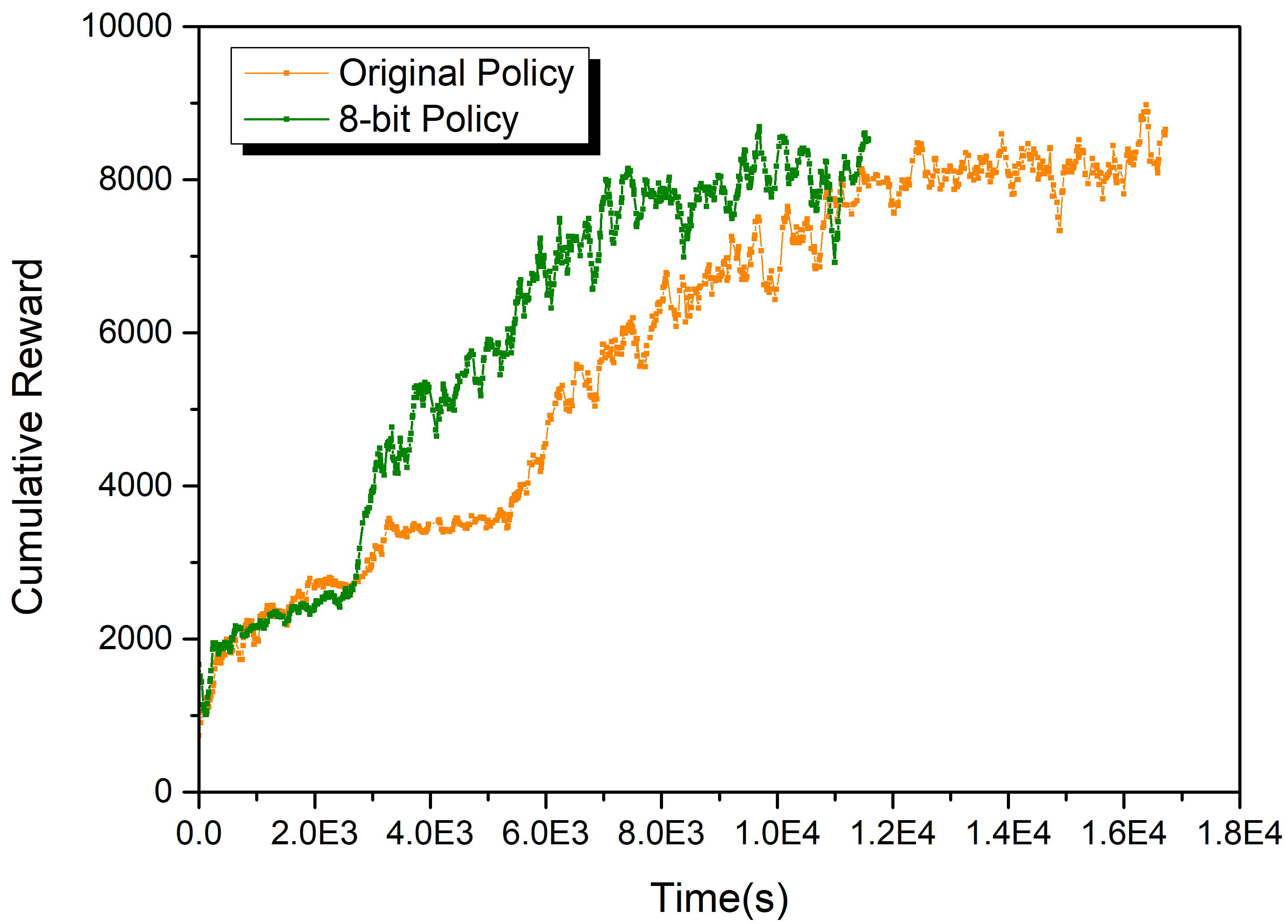}}
	\subfigure[road-step]{\label{fig:quant:19}
		\includegraphics[width=0.245\linewidth]{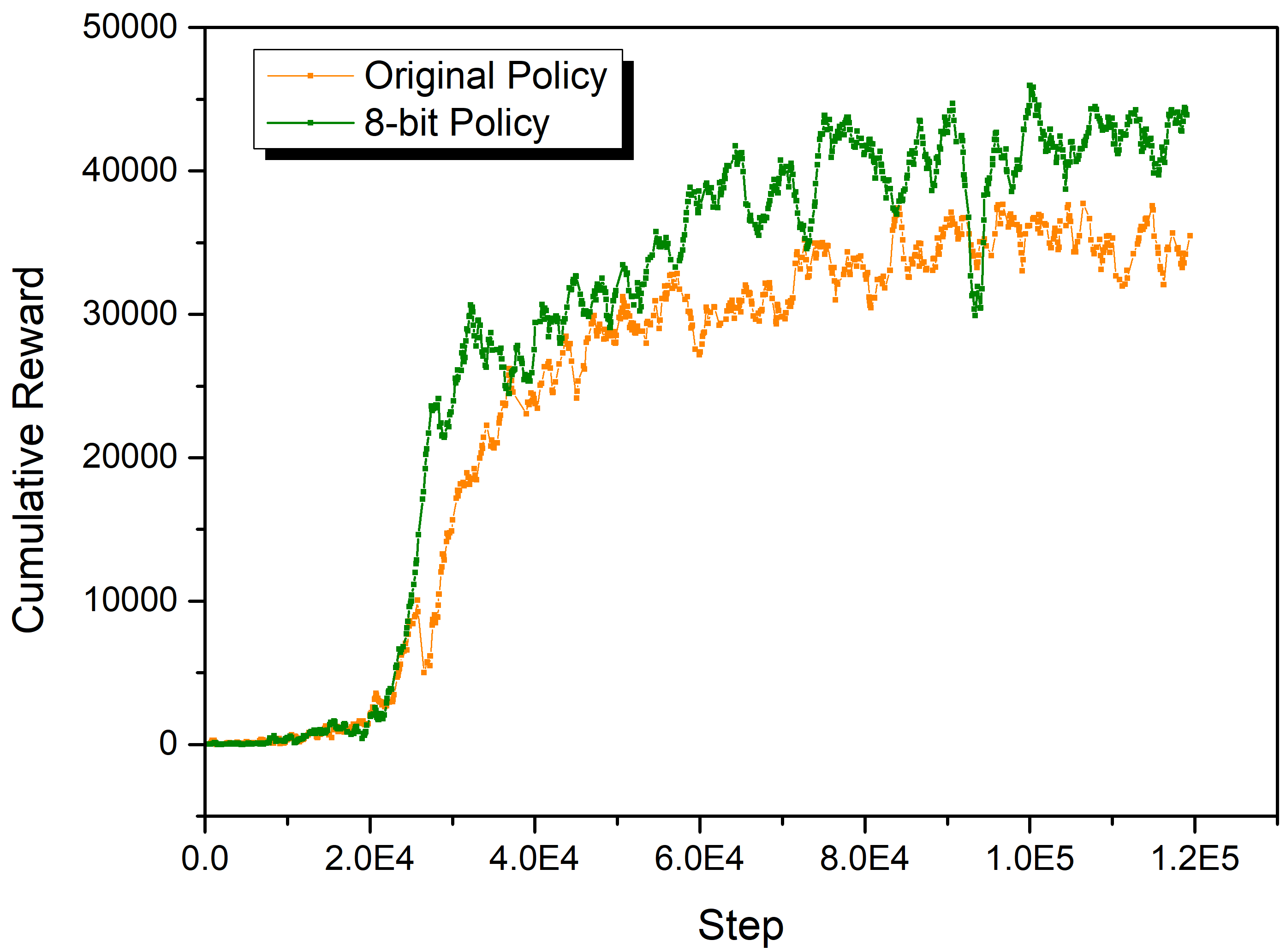}}
	\subfigure[road-time]{\label{fig:quant:20}
		\includegraphics[width=0.245\linewidth]{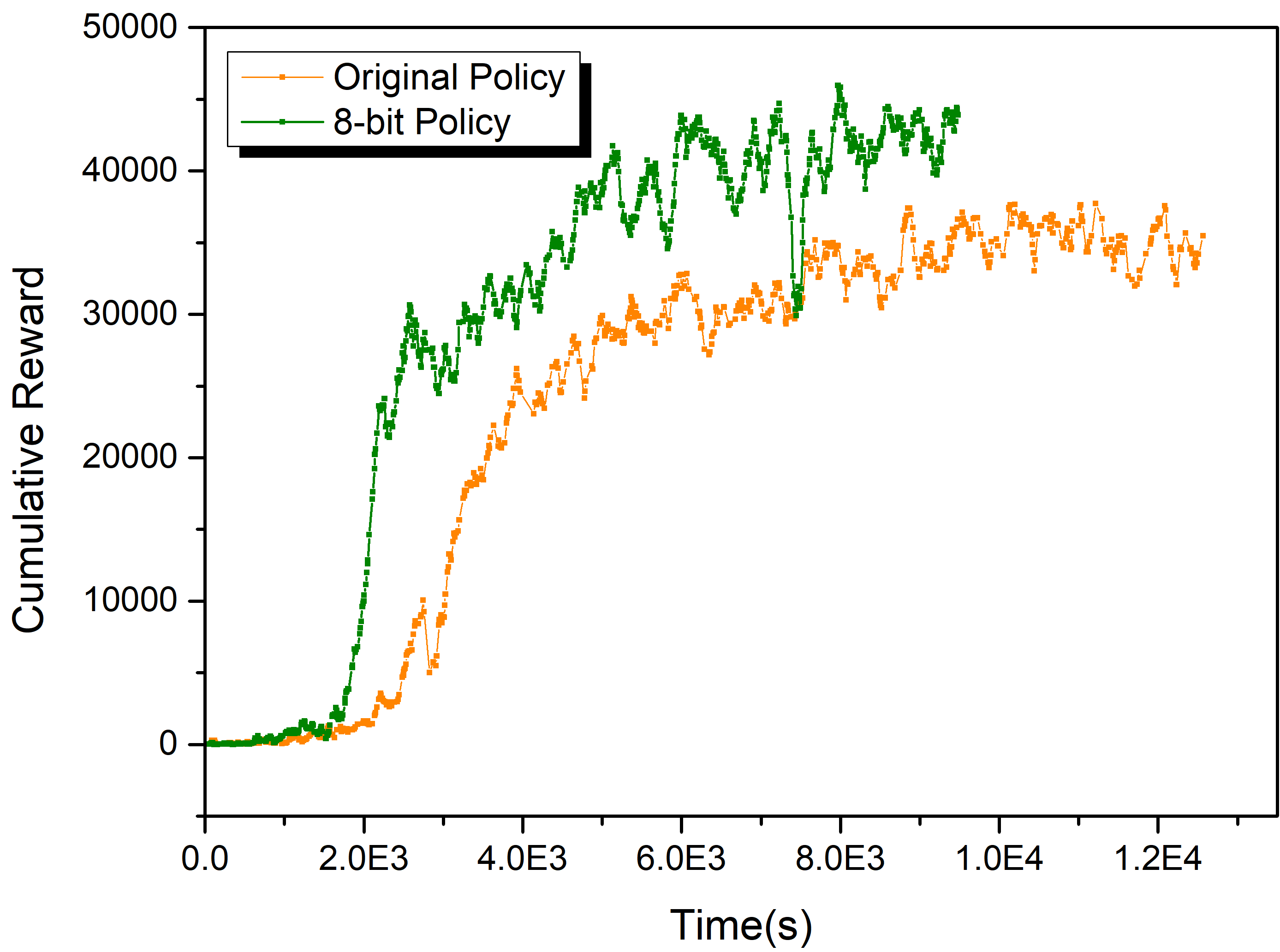}}
		
	\subfigure[sea-step]{\label{fig:quant:21}
		\includegraphics[width=0.245\linewidth]{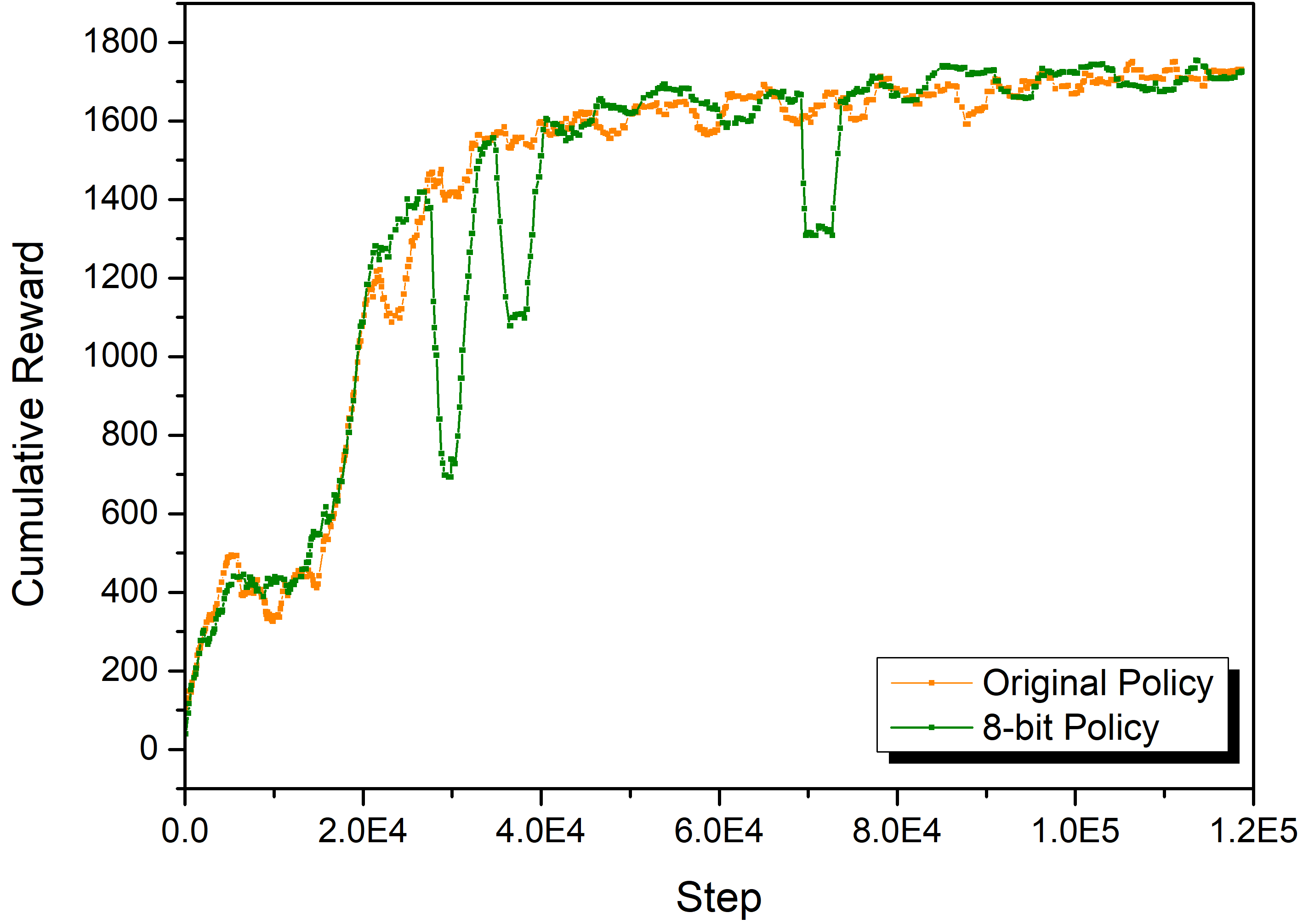}}
	\subfigure[sea-time]{\label{fig:quant:22}
		\includegraphics[width=0.245\linewidth]{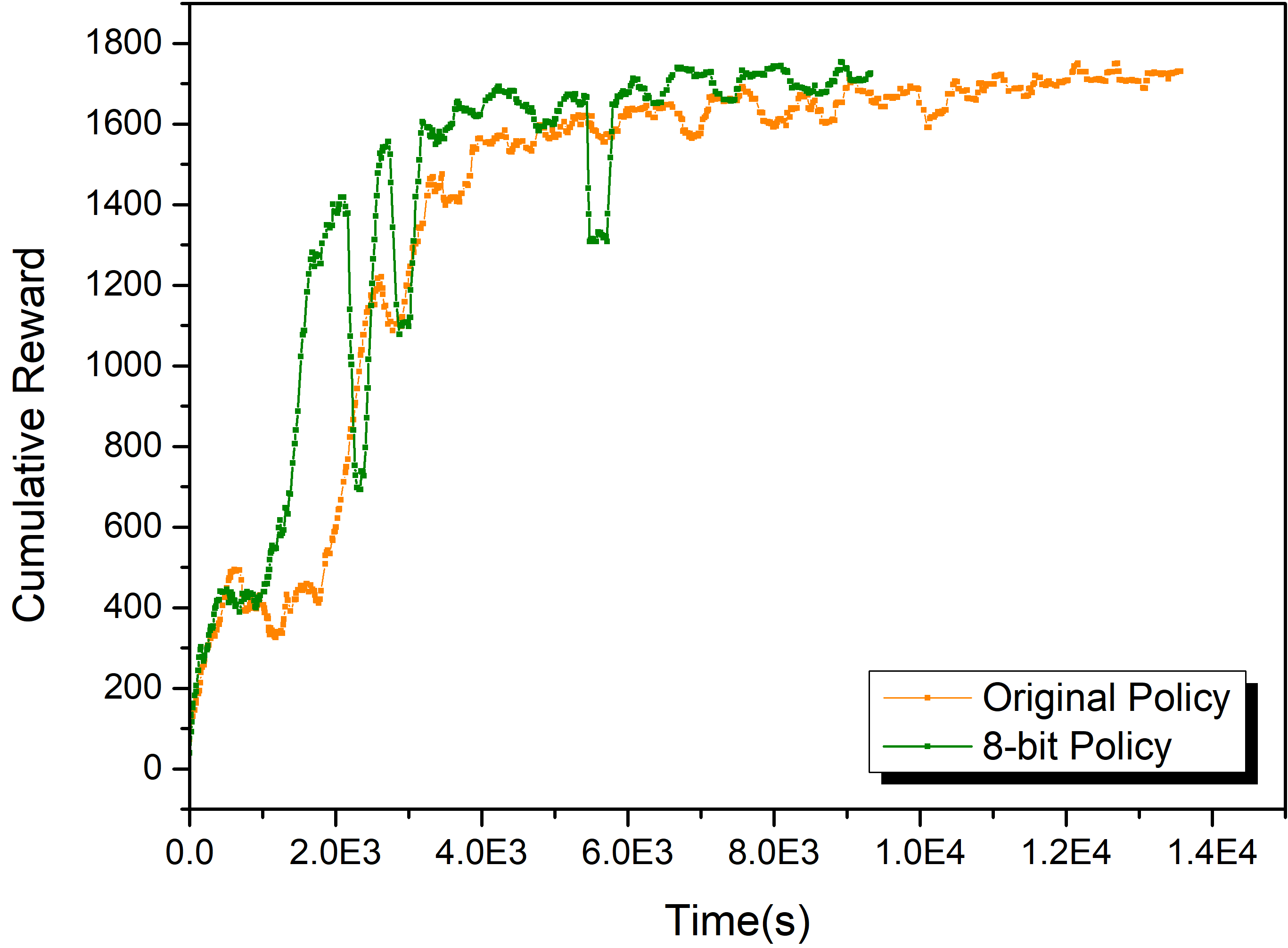}}
	\subfigure[space-step]{\label{fig:quant:23}
		\includegraphics[width=0.245\linewidth]{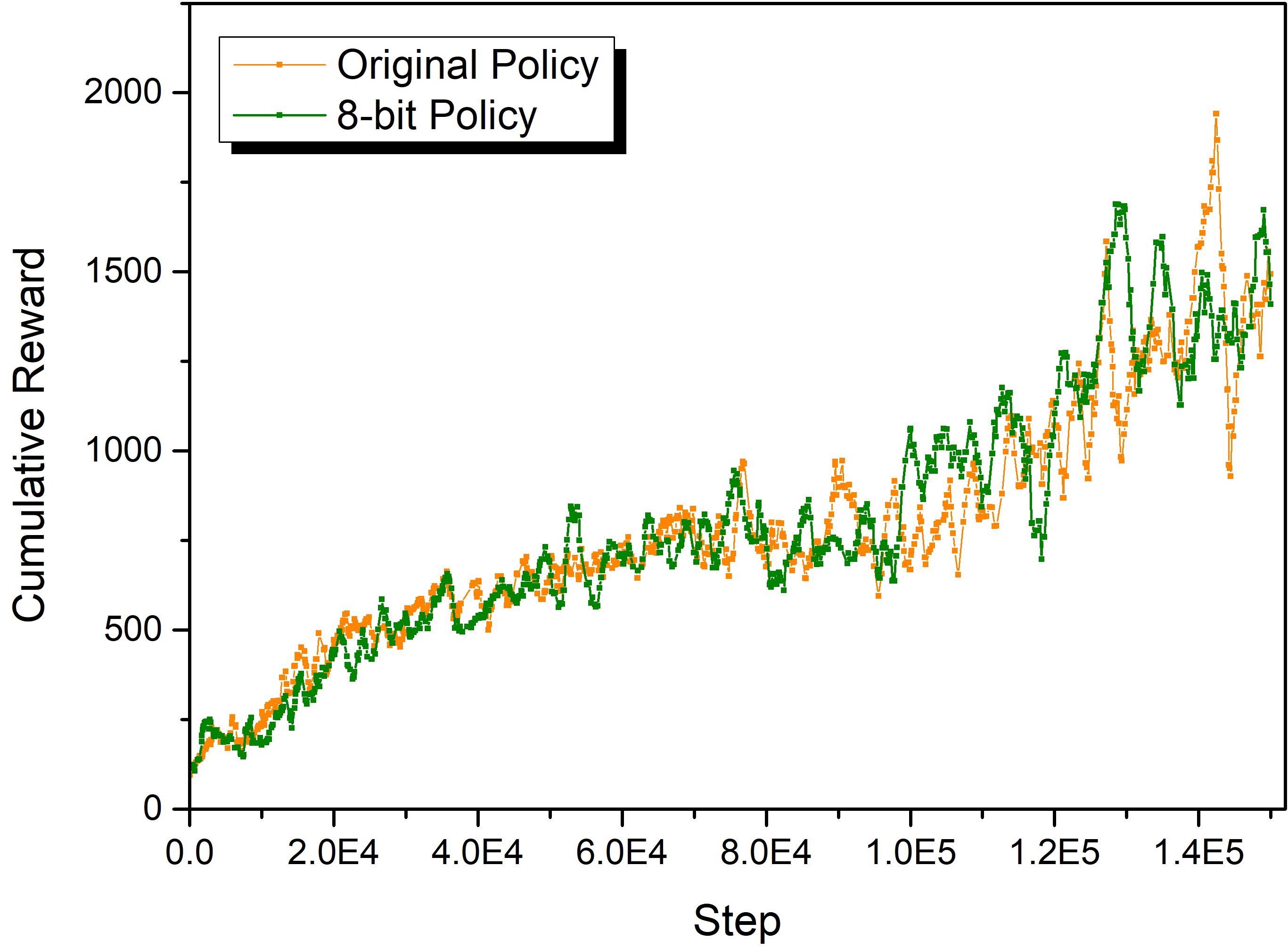}}
	\subfigure[space-time]{\label{fig:quant:24}
		\includegraphics[width=0.245\linewidth]{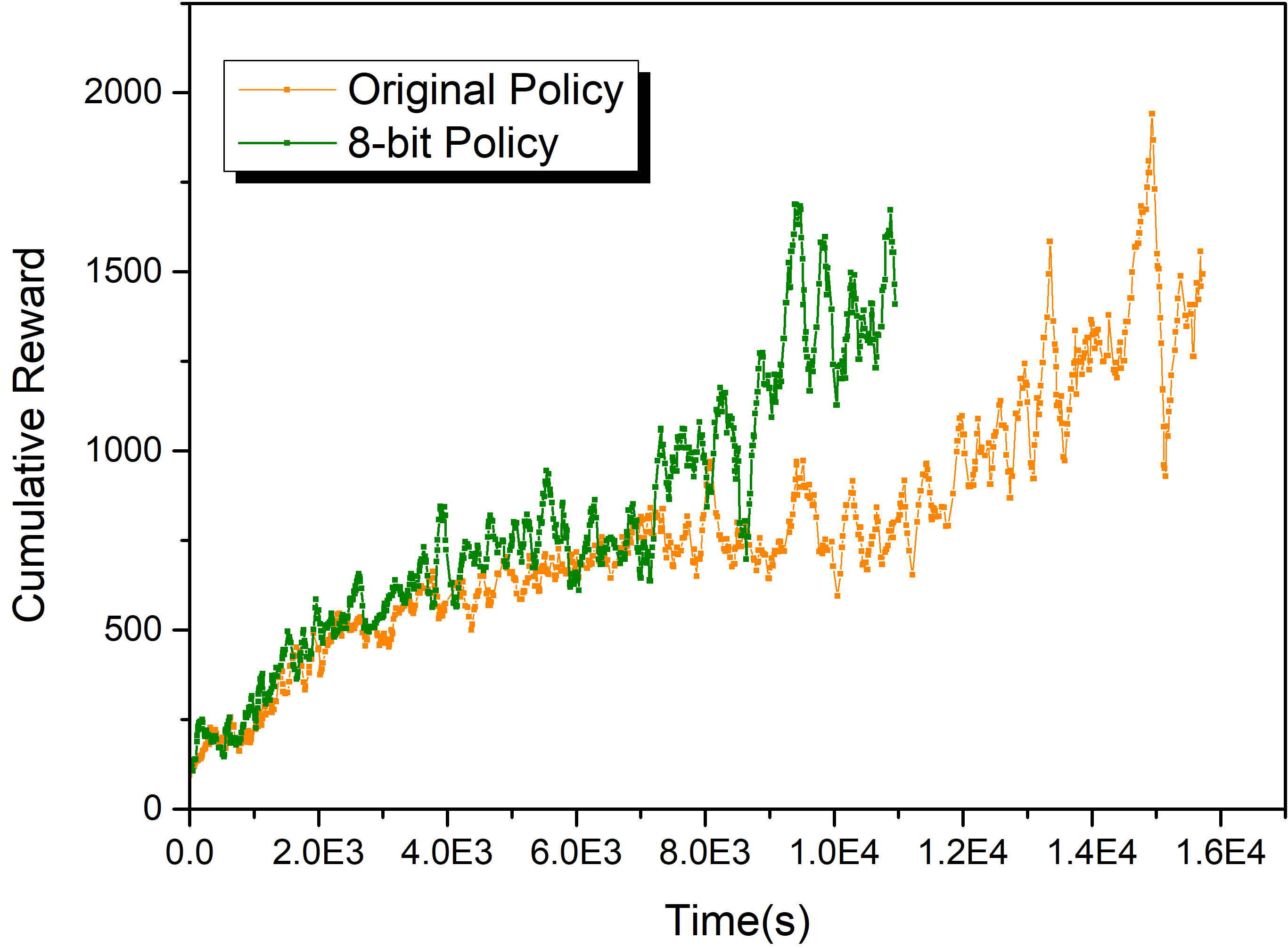}}
	\caption{The parameter quantization results with $NA=12$ and $ND=0$.}
	\label{fig:quant}
\end{figure}

\paragraph{Multi-Compression Methods Results}
As we mentioned before, \textit{AcceRL} provides users with flexible combined compression strategies to maximize the compression efficiency of the policy network. We combined knowledge distillation and 8-bit quantization methods to compress the policy network. The results are shown in Figure \ref{fig:comb}. In Figure \ref{fig:comb}, the 1-st and 3-rd columns show the policy convergence along with training step, and the 2-ed and 4-th columns show the policy convergence along with training time. We notice that the combined policy and Original (FP32) policy have the same convergence rate. But our combined policy is faster than Original policy.

Experimental results demonstrate the effectiveness and ease-of-use of our compression framework \textit{AcceRL}, through policy network compression, the training of the DRL is significantly accelerated and the corresponding policy quality is maintained. The \textit{AcceRL} not only accelerates training, but also takes up less memory with a smaller size. Under the same resources, more environments can be execute to further accelerate data collection. Furthermore, the compressed neural network reduces the communication cost when broadcasting parameters, especially in distributed training scenarios.

\section{Conclusion and Future Work}
In this paper, we proposed the policy acceleration framework \textit{AcceRL} for deep reinforcement learning based on multiple state-of-the-art compression methods and off-policy correction technologies. To the best of our knowledge, this is the first general framework that flexibly organizes multiple compression algorithms together to accelerate reinforcement learning. Compared with the traditional parallel DRL framework, the \textit{AcceRL} adds three main modules, namely \textit{Compressor}, \textit{Corrector}, and \textit{Monitor}. The \textit{Compressor} implements various compression methods and assembles them flexibly in a user-friendly manner. To tackle the off-policy problem, the \textit{Corrector} implements multiple state-of-the-art correction algorithms in a user-friendly manner. In addition, we provide users with a training \textit{Monitor} module to facilitate users to dynamically adjust strategies, which maintain the quality of the policy, and maximize training efficiency. Our framework is still under development, and we hope that everyone will join our plan to make the framework more versatile and efficient. In future work, we hope to further dig into the Advanced Indicator to make the whole framework more automatic without requiring users to specify relevant compression parameters and indicators. The algorithm is currently in the experimental stage. We hope that this framework can provide some support for the landing of DRL in real-world tasks.

\bibliographystyle{unsrt}  
\bibliography{references}  


\end{document}